\documentclass{article}

\usepackage[T1]{fontenc}
\usepackage{microtype}
\usepackage{graphicx}
\usepackage{booktabs}
\usepackage{tabularx}
\usepackage{multirow}
\usepackage{alphalph}
\usepackage{hyperref}
\usepackage{xurl}

\usepackage[preprint]{icml2026}

\usepackage{amsmath,amssymb,amsthm,mathtools}
\usepackage{algorithm}
\usepackage{algorithmic}

\newtheorem{proposition}{Proposition}

\icmltitlerunning{Beyond Reweighting in Discrete MCMC}

\graphicspath{{figures/}}

\hypersetup{
  pdftitle={Beyond Reweighting: Certified Improvements
    from New Transitions in Discrete MCMC},
  pdfsubject={Preprint}
}

\begin{document}
\twocolumn[
\icmltitle{Beyond Reweighting: Certified Improvements\texorpdfstring{\\}{ }from New Transitions in Discrete MCMC}
\begin{icmlauthorlist}
\icmlauthor{Yishan Xi}{anon}
\icmlauthor{Tinglan Wang}{anon}
\end{icmlauthorlist}
\icmlaffiliation{anon}{Independent Researcher}
\icmlcorrespondingauthor{Yishan Xi}{22210690113@m.fudan.edu.cn}
\icmlcorrespondingauthor{Tinglan Wang}{wangtinglan22@mails.ucas.ac.cn}
\icmlkeywords{Markov chain Monte Carlo, asymptotic variance, energy-based models, graph augmentation}
\vskip 0.3in
]
\printAffiliationsAndNotice{}

\begin{abstract}
Optimizing transition probabilities on a fixed graph
may leave substantial limitations on the efficiency
of reversible Markov chain Monte Carlo (MCMC).
We develop a framework for transition-support
design that considers both allowed transitions
and their stationary flows. For a fixed target
distribution, observable, and node-flow budget,
we derive a necessary and sufficient dual
condition for support augmentation to strictly
reduce the optimal value of the asymptotic variance,
together with computable primal--dual certificates
that establish improvements over every admissible
original-support reweighting. We also construct
a family in which adding a single edge reduces
integrated autocorrelation time by an
$\Omega(n)$ factor relative to optimal
original-support weighting, demonstrating
that structural limitations can impose
arbitrarily large relative losses in estimation
efficiency. To realize these gains, we investigate
energy-informed edge selection and constrained
stationary-flow optimization. Experiments on
molecular and Ising state graphs distinguish
the potential value of new transitions from
the performance of kernels constructed on them:
on identical augmented Ising supports, full
weight optimization certifies strict improvements
in 24/24 conditions, whereas degree-based
Metropolis--Hastings weighting certifies none.
Our results provide a principled approach to
characterizing, certifying, and exploiting
transition-support improvements in explicitly
represented finite state spaces.
\end{abstract}

\section{Introduction}

\begin{figure*}[!t]
\centering
\includegraphics[width=\textwidth,trim=0 0 0 90bp,clip]{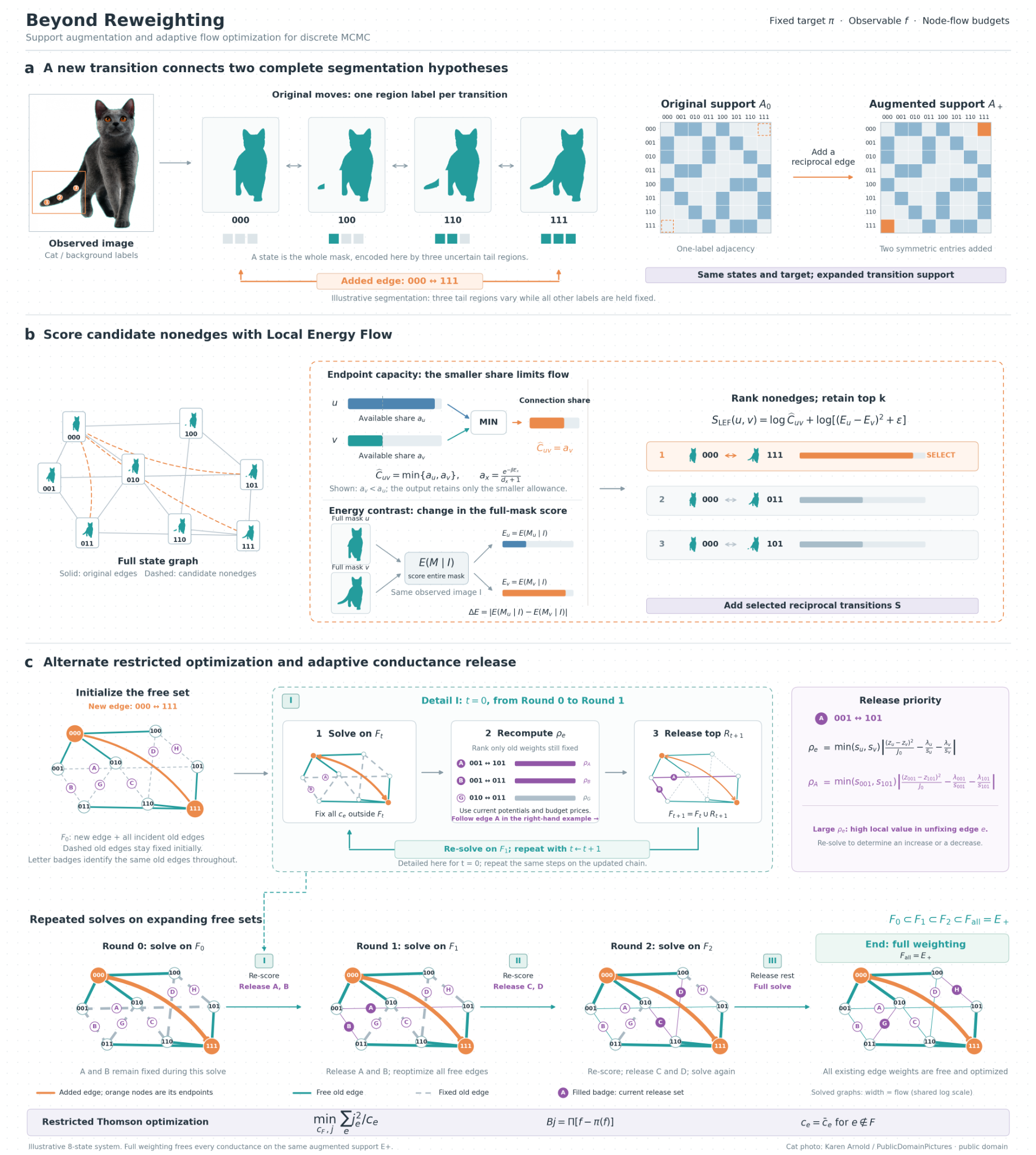}
\caption{\textbf{Transition-support design and adaptive conductance optimization.}
\textbf{(a)} In an illustrative eight-state segmentation problem, adding the
orange edge $000\leftrightarrow111$ expands the transition support under the
same target $\pi$, observable $f$, and node-flow budgets.
\textbf{(b)} Local Energy Flow (LEF) ranks candidate nonedges by combining
the smaller endpoint capacity proxy with the squared energy contrast.
\textbf{(c)} On the augmented support, restricted Thomson solves alternate
with releasing previously fixed conductances. Teal solid edges have free
old conductances, gray dashed edges remain fixed, and filled purple badges
mark the edges released at the corresponding round. Full weighting frees
every conductance on the augmented support.}
\label{fig:method-overview}
\end{figure*}

The efficiency of reversible Markov chain Monte
Carlo (MCMC) depends on which transitions are
permitted and how probabilities are assigned
to them. Most kernel optimization methods
improve transition probabilities within a
prescribed graph, but the graph itself may
limit estimation efficiency in ways that
reweighting cannot overcome. This motivates
treating the \emph{transition support} as a
design variable alongside the stationary
flows assigned to its edges.
Figure~\ref{fig:method-overview}(a) illustrates this distinction. A single
reversible move directly connects two complete segmentation masks that
otherwise require a sequence of single-region moves.

A central question is whether adding transitions
can improve estimation efficiency beyond the
best achievable weighting of the original
graph. This is stronger than outperforming
a particular baseline kernel: it identifies
benefits attributable to changing the support
rather than correcting suboptimal probabilities.
Fastest-mixing methods optimize spectral
convergence on fixed graphs
\citep{boyd2004fastest,oleskertaylor2024geometric},
while graph-jump methods introduce additional
transitions to improve movement across modes
\citep{duan2025graph}. We instead focus on
the optimal value of the asymptotic variance for a
specified observable under a fixed target distribution
and node-flow budget, directly targeting
the statistical efficiency of MCMC estimation.

We develop a theoretical and computational
framework for characterizing, certifying,
and exploiting support improvements.
Specializing resource-constrained $c$-optimal
design \citep{sagnol2011design} to reversible
stationary flows yields an exact dual
characterization: support augmentation
strictly reduces the optimal value of the
asymptotic variance if and only if every
original optimal dual solution violates
at least one added-edge constraint.
We further derive computable primal--dual
certificates that establish whether a
specific augmented kernel outperforms
every admissible original-support weighting.

The potential benefit of changing the support
can be substantial. We construct a family
of reversible chains with uniform stationary
distributions in which a single added edge
achieves an $\Omega(n)$-factor reduction
in integrated autocorrelation time (IACT)
relative to optimal original-support
reweighting. For the specified observable,
this corresponds to an equally large
asymptotic reduction in sample-mean variance
at a fixed sampling length. Thus, even
optimally allocated transition probabilities
can leave arbitrarily large relative losses
in estimation efficiency when the support
is restricted. This establishes the value
of support design beyond fixed-graph
weight optimization.

Identifying valuable transitions, however,
does not guarantee that a particular kernel
will realize their potential benefits.
An exact edge-insertion identity relates
asymptotic variance reduction to
Poisson-potential differences, effective
resistance, and added stationary flow.
We also analyze how reallocating existing
conductances can offset gains from new
connections. We therefore investigate
Local Energy Flow (LEF), an energy-informed
edge-selection heuristic that ranks candidate nonedges
(Figure~\ref{fig:method-overview}(b), Section~\ref{sec:lef}).
On the selected support, we optimize stationary flows for the specified
observable over progressively larger sets of free conductances
(Figure~\ref{fig:method-overview}(c), Section~\ref{sec:realization}).
Construction costs and finite-budget
estimation errors assess the practical
consequences of these design choices.

Our main contributions are:
\begin{enumerate}

\item \textbf{Characterizing and certifying
support improvements.}
We derive a necessary and sufficient
condition for strict support improvement
and develop independently checkable
primal--dual certificates that establish
improvements over all admissible
original-support reweightings.

\item \textbf{Quantifying the limits of
fixed-support reweighting.}
We construct a uniform-target family
in which a single added transition
achieves an $\Omega(n)$-factor IACT
improvement over the optimal original
weighting. This demonstrates that
restricted transition structures can
impose arbitrarily large relative
losses in observable-specific estimation
efficiency.

\item \textbf{Designing and realizing
improved transition kernels.}
We evaluate LEF selection and constrained
flow optimization across 3,640
cell--support pairs and independent
Ising experiments. On identical augmented
Ising supports, full weighting certifies
strict improvements in 24/24 conditions,
whereas degree-based Metropolis--Hastings
weighting certifies none. Budgeted
weighting and measured-budget experiments
further quantify the trade-off between
realized efficiency and construction cost.

\end{enumerate}

Our construction and certification methods
use explicitly enumerated finite state
spaces, normalized target probabilities,
and exact observable means. This setting
enables rigorous comparisons between
support design and weight optimization
while making their computational and
informational requirements explicit.

\section{Setting and Evaluation}
\label{sec:setting}

\paragraph{Targets and observables.}
On a finite state universe $\mathcal U$, let
$\pi(x)\propto\exp[-\beta E(x)]$ be strictly positive. For a nonconstant observable
$f$, define $h=f-\pi(f)$, $\Pi=\operatorname{diag}(\pi)$,
$q=\Pi h$, and $V=h^\top\Pi h>0$. An irreducible reversible kernel $P$ has
stationary conductances $c_{uv}=\pi_uP_{uv}=\pi_vP_{vu}$. Its symmetric Laplacian is
\begin{equation}
 A=\Pi(I-P)=\sum_{e\in\mathcal E}c_e b_eb_e^\top,
 \qquad b_{(u,v)}=e_u-e_v.
 \label{eq:laplacian}
\end{equation}
The asymptotic variance and integrated autocorrelation time (IACT) of $f$ are
\begin{equation}
 \sigma_f^2(P)=2q^\top A^\dagger q-V,
 \qquad \tau_f(P)=\sigma_f^2(P)/V.
 \label{eq:variance}
\end{equation}
Energy defines the configured log density and is available at every evaluated
state, making it a controlled target-sensitive observable. We additionally
evaluate four molecular-topology observables and Ising magnetization using the
same selected edges.

Enumeration supplies $\pi$ and $\pi(f)$ to both weighting and evaluation.
Direct weighted summation already computes this mean; the experiments measure
how support and flow constraints affect the constructed kernels' variance and
finite-time error.

Molecular energies are scalar graph-transformer outputs trained by matching
negative input-energy gradients to categorical noise--data displacements.
Appendix~\ref{app:energy-model} specifies the architecture and training objective;
the Ising energy is the explicit pairwise spin Hamiltonian.

\paragraph{Weighting hierarchy.}
All empirical support and weighting comparisons enforce a minimum self-loop
probability of $\ell=0.05$. Independent sampling serves as an unconstrained
reference with IACT one. Direct evaluations use uniform-neighbor degree-MH,
whose off-diagonal flow is
\begin{equation}
 c_{uv}=(1-\ell)\min\{\pi_u/d_u,\pi_v/d_v\}.
 \label{eq:degree-flow}
\end{equation}
Adding an edge changes the endpoint degrees and can reduce existing incident
flows. The support oracle permits \emph{all} nonnegative conductances with
$\sum_{e\ni u}c_e\le s_u$, where $s_u=\pi_u(1-\ell)$. Write $\tau^\star_{\mathcal E}$
for the IACT infimum over weights with connected positive-conductance support
in $\mathcal E$. Every upper-bound witness is connected. Between these endpoints, we optimize only
conductances incident to added-edge endpoints and fix all other flows to their
augmented degree-MH values (Section~\ref{sec:realization}). The optimization
retains full-graph current conservation and target information. This hierarchy
measures realized kernel performance, the effect of limited free conductances,
and selected-support potential under common constraints.

\paragraph{Cohorts and statistical units.}
The molecular experiments use four 65-state development components, four
structurally constructed components rescored by ten checkpoints from five seeds,
and eight new components crossed with those same checkpoints. Held-out checkpoints
were held out from LEF development and selector construction. The new components
use the same deterministic energy-blind rule, with component as the outer unit.
The external Ising study enumerates 256 states per instance and uses the natural
single-spin-flip graph. Adaptive release is developed on the original
12 coefficient instances and confirmed on eight new ones, with the rule fixed
before generating their coefficients. Table~\ref{tab:evidence-map} connects the
questions to their evidence and outer units.

\begin{table}[t]
\caption{Questions, evidence, and aggregation. Temperatures and checkpoints
remain nested within the stated outer units.}
\label{tab:evidence-map}
\centering\small
\begin{tabularx}{\columnwidth}{lX}
\toprule
Question & Evidence and outer unit \\
\midrule
Support value & Instancewise certificates; uniform-target separation family \\
Checkpoint transfer & 5 model seeds; 40 nested cells \\
Component transfer & 8 new components; 80 cells; same construction rule \\
Selection value & Common weighting; 5 molecular seeds and 12 Ising instances \\
Release attribution & 12 development Ising instances; matched free-set sizes \\
Factorial confirmation & 12 new Ising instances; selection $\times$ release; 3 repeats \\
Partial realization & 8 other fresh Ising instances; paired temperatures; 3 repeats \\
Remaining value & 12 development Ising instances; saved partial stages \\
Economic value & Same 8 fresh instances; stationary ESS cost model \\
Finite-budget error & 5 new Ising seeds; 1024--4096 states; 3 starts \\
\bottomrule
\end{tabularx}
\end{table}

Frozen protocols govern prospective tests; subsequent weighting and mechanism
analyses reuse their components. The construction scale study uses three seeds
at five sizes; the finite-budget study uses five new seeds at three paired sizes.
Intervals resample outer log effects. The two fresh Ising cohorts are distinct
and are not pooled; the appendices give protocols and effects.

\section{Certifying the Value of Added Support}
\label{sec:bridges}

Adding support weakly decreases the best achievable variance. We test whether a
four-edge selection yields a \emph{strict}, quantitatively resolved advantage over
the best base weighting. Two finite witnesses establish this comparison.

\subsection{A Checkable Lower--Upper Separation}

Write $B$ for an oriented incidence matrix. Thomson's principle expresses
$q^\top A(c)^\dagger q$ as the minimum electrical energy of currents with $Bj=q$.
Specializing resource-constrained $c$-optimal design \citep{sagnol2011design}
to edge incidences gives the convex program
\begin{equation}
 \begin{aligned}
 \min_{c,j,t}\;&\sum_e t_e\\
 \text{s.t. }&Bj=q,\quad c_e,t_e\ge0,\quad j_e^2\le c_et_e,\\
 &\sum_{e\ni u}c_e\le s_u\quad\text{for every }u.
 \end{aligned}
 \label{eq:primal}
\end{equation}
This closed program has the same infimum as connected feasible weighting
(Appendix~\ref{app:priced-support-extension}); zero conductance requires zero current.
For lower bounds, the variational identity
$q^\top A^\dagger q=\sup_z(2q^\top z-z^\top Az)$ gives a potential-based dual.

\begin{proposition}[Support-improvement certificate]
\label{prop:certificate}
Fix $\pi,f,\ell$ and original support $\mathcal E_0$. Suppose $y\ge0$ and
$y_u+y_v\ge(z_u-z_v)^2$ for all $(u,v)\in\mathcal E_0$. Then
\begin{equation}
 L_\sigma=4q^\top z-2\sum_u s_uy_u-V
 \label{eq:fixed-support-lower}
\end{equation}
lower-bounds the asymptotic variance of every admissible irreducible reversible
weighting on $\mathcal E_0$. For any feasible connected augmented conductances
$c$ and current $j$ with $Bj=q$ and $j_e=0$ whenever $c_e=0$,
\begin{equation}
 U_\sigma=2\sum_{e:c_e>0}j_e^2/c_e-V
 \label{eq:augmented-upper}
\end{equation}
upper-bounds that augmented chain's variance. Consequently,
$U_\sigma<L_\sigma$ certifies a strict topology advantage. IACT bounds are
$L_\tau=L_\sigma/V$ and $U_\tau=U_\sigma/V$.
\end{proposition}

\begin{proof}
For every feasible base conductance vector,
$z^\top A(c)z=\sum_e c_e(z_u-z_v)^2\le\sum_u s_uy_u$.
Substitution in the variational identity and Equation~\ref{eq:variance} gives
the lower bound. Any feasible current supplies an upper bound on the minimum
electrical energy, giving the upper bound by the same equation.
\end{proof}

This certificate applies classical variational and electrical-network tools
\citep{doyle1984random} to the fixed-target support intervention used here. The
comparison covers observable-optimal base weighting, thereby resolving the
objective mismatch inherent in a comparison against spectral optimization alone.

\begin{samepage}
\begin{proposition}[Exact criterion for strict support value]
\label{prop:support-criterion}
Fix the target, observable, and node budgets. For a connected support
$\mathcal E_0$ and an added edge set $S$, let $J^\star_{\mathcal E}$ denote
the minimum in Equation~\ref{eq:primal} on support $\mathcal E$. Then
$J^\star_{\mathcal E_0\cup S}=J^\star_{\mathcal E_0}$ if and only if some
optimal dual $(z,y)$ on $\mathcal E_0$ satisfies
\begin{equation}
 (z_u-z_v)^2\le y_u+y_v \quad\text{for every }(u,v)\in S.
 \label{eq:priced-support-equality}
\end{equation}
Equivalently, adding $S$ strictly reduces the best achievable asymptotic
variance if and only if every original optimal dual violates an added-edge
constraint.
\end{proposition}
\end{samepage}

Adding edges leaves the dual objective unchanged and adds the edge
constraints. By strong duality and attainment, the values coincide exactly
when the tighter feasible set contains an original optimum. A single
violating dual optimum leaves open whether another optimum extends to $S$.
Appendix~\ref{app:priced-support-extension} proves attainment, gives a
five-node example, and derives a finite-witness upper bound on possible gains.

\begin{proposition}[An unbounded support--weighting separation]
\label{prop:structural-separation}
For integer $m\ge2$, join two $m$-cliques by a length-$m$ path between designated hubs,
giving $n=3m-1$ states. Fix uniform $\pi$, a constant self-loop floor $0<\ell<1$, and
$f=+1,-1,0$ on the left clique, right clique, and internal path respectively.
Then $V=2m/n$, and the infimum over admissible base weights satisfies
\[
 \tau^\star_{\rm base}\ge\frac{m^2}{1-\ell}-1.
\]
Adding the hub--hub edge admits a connected feasible chain with
\[
 \tau_{\rm aug}\le
 \frac{8(m-1)^2+4m^2}{m(1-\ell)}-1.
\]
Consequently $\tau^\star_{\rm base}/\tau_{\rm aug}=\Omega(n)$; these bounds
strictly separate for $m\ge11$.
\end{proposition}

Every path edge carries net current $m/n$ under the base support. A direct
hub connection reroutes this current within the same node budgets, yielding
the displayed Thomson upper bound (Appendix~\ref{app:structural-extension}).
This specializes classical all-weight bottleneck reasoning
\citep{roch2005bounding,oleskertaylor2024geometric} to a fixed observable, with
$N_{\rm eff}=n$ and $V\to2/3$. On the same augmented support, every weighting
still has worst-direction relaxation time $(1-\lambda_2(P))^{-1}=\Omega(n^2)$:
the observable gain coexists with a slow corridor mode
(Appendix~\ref{app:uniform-separation}). This family isolates support value;
Sections~\ref{sec:lef}--\ref{sec:realization} evaluate how local selection and
partial weighting recover it on the empirical supports.

\paragraph{Exact arithmetic and target definition.}
We exactly normalize round-trip decimal probabilities and center the
observable under this common rational target. We repair solver outputs to
meet the exact budget, dual, and current constraints, then check the bounds
independently (Appendix~\ref{app:weighting-certificate}). A separate interval
audit of the exponential target preserves 123 LEF/capacity separations on
the 40 checkpoint and 24 Ising cells, with minimum relative margin
$5.26\times10^{-5}$ (Appendix~\ref{app:target-interval}).
Appendix~\ref{app:artifacts} includes an executable analytic witness.

\subsection{Results Across Molecular and Natural Ising Supports}

Table~\ref{tab:certificates} reports the complete certificate census. LEF-selected
supports admit strictly better observable dynamics than every original-support
weighting in 4/4 development, 38/40 checkpoint, and 80/80 new-component cells.
The Ising study extends the result to all 24 conditions on natural single-spin
supports. Capacity-selected supports certify many of the same improvements,
establishing topology value shared by both local rules.

\begin{table*}[t]
\caption{Strict topology separations against \emph{every} admissible reversible
base-support weighting. ``Degree-MH'' uses the directly reconstructed augmented
kernel; ``optimized'' uses a feasible observable-optimized upper witness on the
same selected support. All comparisons use identical targets, observables, and
self-loop floors. Counts are instancewise certificates.}
\label{tab:certificates}
\centering
\begin{tabular}{lrrrrr}
\toprule
 & & \multicolumn{2}{c}{LEF-selected support} & \multicolumn{2}{c}{Capacity-selected support}\\
\cmidrule(lr){3-4}\cmidrule(lr){5-6}
Cohort & Cells & Degree-MH & Optimized & Degree-MH & Optimized\\
\midrule
Molecular development & 4 & 3 & 4 & 3 & 4\\
Held-out checkpoints & 40 & 24 & 38 & 20 & 38\\
Same-rule new components & 80 & 72 & 80 & 68 & 78\\
Natural-support Ising & 24 & 0 & 24 & 0 & 23\\
\bottomrule
\end{tabular}

\end{table*}

The direct and optimized columns expose a substantial distinction. On Ising,
fixed-support weight optimization has enough room that the direct four-edge
degree-MH kernels yield 0/24 separations. Optimizing the augmented weights
nevertheless certifies additional topology value in 24/24 LEF conditions, with
$L_{\tau,\mathrm{base}}/U_{\tau,\mathrm{LEF}}$ ranging from 1.012 to 1.421.
On the molecular graphs, the direct LEF kernels themselves already separate in
3/4, 24/40, and 72/80 cells.

All 468 support problems have verified rational witnesses; 466 electrical-energy
bound gaps are at most $10^{-4}$, with two remaining gaps 0.00144 and 0.00565
(Appendix~\ref{app:weighting-certificate}). The following audit expands the
comparison to 3,640 unique supports, retaining complete numerical intervals.

\section{Which Selected Supports Have Value?}
\label{sec:lef}

\subsection{Edge Scores from Endpoint Information}

For a candidate nonedge $(u,v)$, let $d_x$ be the original degree of endpoint
$x$ and set $a_x=e^{-\beta E_x}/(d_x+1)$. The capacity proxy is the smaller
endpoint share, as shown in Figure~\ref{fig:method-overview}(b):
\begin{equation}
 \widehat C_{uv}=\min\{a_u,a_v\}.
 \label{eq:capacity}
\end{equation}
Capacity-only ranks this feature. LEF adds a raw observable-linked contrast:
\begin{equation}
 S_{\rm LEF}(u,v)=\log\widehat C_{uv}
      +\log\big((E_u-E_v)^2+10^{-12}\big).
 \label{eq:lef-preview}
\end{equation}
The schematic shows a top-$k$ selection. We select $k=4$ edges in one shot
with canonical tie breaking in the main experiments. The evaluated augmented
kernel uses the true final degrees,
including all shared endpoints. Given a candidate list, the scores require
cached endpoint energies, degrees, and temperature. The target normalization
cancels in the ranking. The Ising candidate generator enumerates state pairs,
and the weighting stage uses the full target and graph
(Section~\ref{sec:realization}).

Direct LEF improves development energy IACT by $9.450\times$ geometrically.
The frozen checkpoint test improves all five outer aggregates and 35/40 cells;
the separate component test improves 8/8 aggregates and 79/80 cells. On the
original 40 cells, two root-free, edge-count-matched base constructions retain
36/40 and 40/40 improvements (Appendix~\ref{app:base-graph-sensitivity}).

The original 40 targets have median effective support 1.806 of 65 states.
Base and LEF IACT medians are 2670.324 and 38.002; the median paired ratio is
16.265. Larger ratios occur on extremely slow base chains. Temperature controls
below examine more diffuse targets; the appendix gives high-precision checks.

\subsection{Common Observable-Optimal Weighting}
\label{sec:common-weighting}

To isolate selection from subsequent weighting, we apply the same
observable-optimal program to LEF, capacity, energy-only, maximum-Hamming, and
all saved random supports. Every method has four added edges and the same
target, candidate universe, and self-loop floor. The audit covers the original
40 molecular and 24 Ising cells, with 64 random portfolios per molecular cell
and 32 per Ising cell. Including the original support in each cell and
deduplicating gives 3,640 cell--support pairs. All witnesses verify
independently; 3,576 electrical-energy intervals have relative width at most
$10^{-4}$, and 64 wider intervals retain their actual bounds.

For a control $C$, Table~\ref{tab:optimized-selectors} reports
$\tau^*_C/\tau^*_{\rm LEF}$, bounded cellwise by
$[L_C/U_{\rm LEF},U_C/L_{\rm LEF}]$. The random comparator is the geometric mean
of the optimized IACTs of all portfolios within a cell. Outer aggregation uses
the original five molecular seeds or twelve Ising coefficient instances,
preserving paired temperatures. These repeated support evaluations add
within-cell comparisons to the existing cohorts.

\begin{table}[t]
\caption{Selection under common weighting. Ratios above one favor LEF;
the last column is the median paired cell ratio. Outer intervals also incorporate
numerical bounds. Random refers to the cellwise portfolio ensemble.}
\label{tab:optimized-selectors}
\centering\footnotesize
\setlength{\tabcolsep}{3pt}
\begin{tabular}{llrrr}
\toprule
Group & Control & Ratio & 95\% outer interval & Median\\
\midrule
Molecular & Random & 58.795 & [25.040, 154.300] & 19.106\\
 & Hamming & 62.691 & [26.506, 163.335] & 19.575\\
 & Energy-only & 62.922 & [26.541, 163.826] & 19.589\\
 & Capacity & 1.976 & [1.213, 4.308] & 1.005\\
\midrule
Ising & Random & 1.141 & [1.100, 1.187] & 1.125\\
 & Hamming & 1.139 & [1.092, 1.189] & 1.131\\
 & Energy-only & 1.134 & [1.091, 1.182] & 1.110\\
 & Capacity & 1.014 & [0.980, 1.044] & 1.048\\
\bottomrule
\end{tabular}

\end{table}

LEF exceeds random, distance, and energy-only on every outer aggregate in both
cohorts. Certificate frequency and effect magnitude convey distinct information:
random Ising supports certify topology gains in 656/768 portfolios, while LEF
strictly improves on 766/768 random supports after common weighting. The
molecular counts are 939/2,560 and 2,412/2,560 respectively.

Capacity is the closest comparator. The molecular geometric ratio is 1.976,
but its cell median is 1.005: the increment concentrates in a subset of cells.
There are 25 strict LEF wins, two capacity wins, eight identical supports, and
five overlapping intervals. On Ising, the overall ratio is 1.014 with outer
interval $[0.980,1.044]$. At $\beta=0.3$, all twelve cells strictly favor LEF;
at $\beta=0.8$, capacity wins in five of six ferromagnetic cells. This conditional
ordering survives common weighting and therefore reflects selected-support
differences (Appendix~\ref{app:optimized-controls}). A separate fresh-instance
comparison under matched partial weighting tests whether LEF adds value when
only a limited conductance budget can be released
(Table~\ref{tab:ising-factorial}).

\subsection{Temperature, Observable, and Support Dependence}
\label{sec:ising}

The Ising study uses six ferromagnetic and six mixed-sign $2\times4$ lattices
at $\beta=0.3,0.8$, with 512 hash-selected candidate nonedges. Direct LEF
improves base energy IACT in 23/24 conditions; at $\beta=0.3$ effective support
is 60.9--130.9. On the same edges, magnetization improves in 24/24, while
capacity beats LEF in 18/24 (Appendix~\ref{app:ising}).

On molecular components, changing temperature to effective supports
$N_{\rm eff}=6.5,19.5,45.5$ retains 40/40 direct improvements, with geometric
gains 2.095, 1.404, and 1.200 (Appendix~\ref{app:diffuse}). Four prespecified
topology observables have five favorable outer aggregates each, with
heterogeneous cell effects (Appendix~\ref{app:scope-audits}). Local energy
contrast itself has conditional value: its held-out capacity-matched edge
concordance is 0.374, and direct LEF/capacity improvement occurs in 27/40 cells
against a prespecified 30/40 safeguard. Together with the common-weighting
comparison, these results locate LEF's benefit at the composite-rule level.

\section{Realizing Value with Adaptive Weighting}
\label{sec:realization}

\subsection{A Restricted Weighting Problem}

Given selected edges $S$, start with degree-MH on $\mathcal E_+=\mathcal E_0\cup S$.
The initial free set $\mathcal F_0$ contains the added edges and all old edges
incident to their endpoints (Figure~\ref{fig:method-overview}(c)). For any
current free set $\mathcal F$, conductances outside $\mathcal F$ stay exactly
fixed at their initial values $\bar c_e$.
The residual node capacity is
$r_u=s_u-\sum_{e\ni u,\,e\notin\mathcal F}\bar c_e$.
We solve the restricted Thomson problem
\begin{equation}
 \begin{aligned}
 \min_{c_{\mathcal F},j}\quad &
 \sum_{e\in\mathcal F}\frac{j_e^2}{c_e}
 +\sum_{e\notin\mathcal F}\frac{j_e^2}{\bar c_e}\\
 \text{s.t.}\quad &Bj=q,\quad c_e\ge0,\quad
 \sum_{e\ni u,\,e\in\mathcal F}c_e\le r_u.
 \end{aligned}
 \label{eq:local-weighting}
\end{equation}
The conic implementation retains all graph currents and optimizes only
$c_{\mathcal F}$. Rational budget repair preserves fixed flows, and a positive
degree-MH mixture retains connectivity. Each output has a fixed-chain potential
lower bound and current upper bound. We retain the incumbent whenever a new
repaired proposal has a larger Thomson upper bound. This makes every nested
stage feasible while using full-graph target and potential information.

Endpoint updates improve all 128 evaluated LEF/capacity chains over degree-MH.
Molecular LEF certification rises from 24/40 to 35/40, versus 38/40 with full
weighting. Ising LEF median IACT falls from 14.568 to 9.124, but certification
remains 0/24; full weighting reaches 3.466 and 24/24. The identical selected
support thus motivates releasing additional old conductances
(Appendix~\ref{app:local-update}).

\subsection{Potential- and Dual-Guided Release}

With $g=A(c)^\dagger q$, the derivative of $J(c)=q^\top A(c)^\dagger q$ is
$\partial J/\partial c_e=-(g_u-g_v)^2$. Node budgets couple this benefit to the
cost of reallocating capacity. We use this structure to rank frozen edges.
Let $z$ be the current fixed-chain variational potential, $J_0$ the initial
degree-MH electrical-energy upper bound, and $\lambda_u$ the budget multiplier
from the accepted solve, normalized to unit node capacities and objective $J/J_0$.
For a frozen edge $e=(u,v)$, define
\begin{equation}
 \rho_e=\min(s_u,s_v)\left|
 \frac{(z_u-z_v)^2}{J_0}-\frac{\lambda_u}{s_u}-\frac{\lambda_v}{s_v}
 \right|.
 \label{eq:release-score}
\end{equation}
Figure~\ref{fig:method-overview}(c) shows the original \emph{combined} rule.
At each round, it recomputes $\rho_e$ from the latest accepted solution and
releases the highest-scoring fixed edges. The next solve determines whether
their conductances increase or decrease. The \emph{potential-only} alternative
ranks $\min(s_u,s_v)(z_u-z_v)^2/J_0$. Both rules are evaluated through exact
fixed-chain witnesses. After the endpoint solve,
nested free sets reach 10\%, 20\%, and 40\% of augmented edges, resolving
Equation~\ref{eq:local-weighting} after each release. Hash-priority release
uses matched set sizes; full weighting frees every conductance. Timing includes
all preceding construction stages.

\paragraph{Development and prospective confirmation.}
On 24 development conditions, potential-only beats combined in 23/24 at 20\%,
while combined wins 22/24 at 40\% (Appendix~\ref{app:release-ablation}).
A frozen $2\times2$ confirmation crosses LEF/capacity selection with
combined/potential-only release on 12 \emph{new} Ising instances (24 paired
temperature conditions), using matched four-edge budgets, 512 candidates and
identical endpoint starts within each selector. At 20\%, potential-only has
lower IACT for both selectors, and LEF has lower IACT under both release
policies in the four Holm-adjusted comparisons (Table~\ref{tab:ising-factorial}).
For LEF, potential-only wins 23/24 pairwise comparisons and yields 7/24
certificates against the original-support optimum. At 40\%, combined wins
23/24 on LEF and 24/24 on capacity. The reversal compares complete policies,
each with its own 10\% and 20\% release history. LEF's gain varies by temperature
and coupling, and the interaction interval spans the null
(Appendix~\ref{app:ising-four-factor}).

\begin{table}[t]
\caption{New pre-specified Ising $2\times2$ confirmation at 20\% release.
C = combined, P = potential-only, Cap = capacity; ratios $>1$ favor
the denominator (lower IACT); intervals
resample 12 coefficient instances with paired temperatures. Counts are
strict wins among 24 conditions; $p$ values are Holm-adjusted.}
\label{tab:ising-factorial}
\centering\footnotesize
\setlength{\tabcolsep}{2.5pt}
\begin{tabular}{@{}lccc@{}}
\toprule
Contrast & Ratio [95\% CI] & Wins & $p_{\rm Holm}$ \\
\midrule
LEF: C/P & 1.045 [1.031,1.057] & 23/24 & .0029 \\
Cap: C/P & 1.039 [1.027,1.052] & 23/24 & .0020 \\
C: Cap/LEF & 1.035 [1.015,1.060] & 20/24 & .0107 \\
P: Cap/LEF & 1.041 [1.023,1.060] & 21/24 & .0088 \\
\bottomrule
\end{tabular}

\end{table}

\begin{figure*}[t]
\centering
\includegraphics[width=\textwidth]{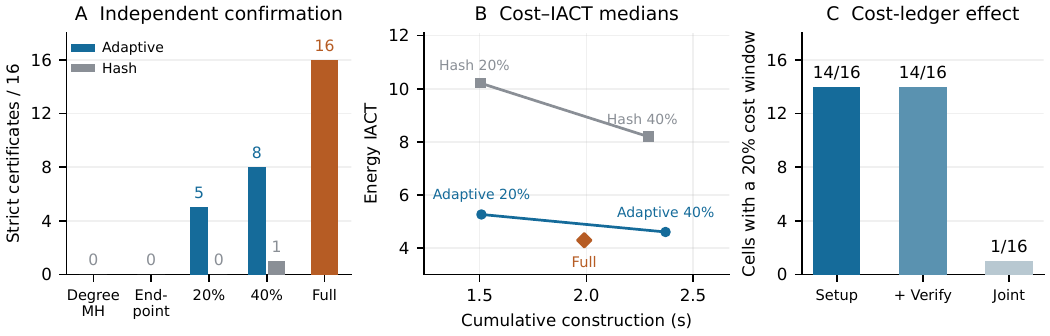}
\caption{\textbf{Independent confirmation: realization and cost.}
Earlier combined-versus-hash study on eight Ising instances at two temperatures.
Adaptive denotes combined release. A: strict topology
certificates, requiring all three construction repeats to certify.
B: cohort medians of cumulative construction time and IACT; points summarize
separate distributions. C: cells with a cheapest-cost window for adaptive 20\%
among the tested kernels, including the optimized base kernel in every ledger.
The ledgers charge sampling setup, then verification, then joint proof-and-kernel
construction. Panel C counts
windows from curves combining measured times and exact stationary IACT;
Effective sample size (ESS) is an asymptotic extrapolation.}
\label{fig:realization}
\end{figure*}

\subsection{Earlier Frozen New-Instance Confirmation}

Distinct from the 12-instance factorial test, Figure~\ref{fig:realization}
summarizes eight earlier fresh coefficient instances
(four per ensemble), each at $\beta=0.3,0.8$, under the frozen combined rule.
Each condition has three balanced-order construction repeats; the first fixes
the sampled kernel and timing uses their median. Bootstrap resamples retain
both temperatures within each coefficient instance.

Combined 20\% certifies \textbf{5/16} conditions, compared with 0/16 for
degree-MH, endpoint updates, and hash 20\%. In all five certified conditions,
the slowest adaptive construction is faster than the fastest full construction.
Combined 40\% certifies \textbf{8/16}, hash 40\% certifies 1/16, and full weighting
certifies 16/16. The 20\%, 40\%, and full medians are respectively
1.508, 2.371, and 1.990 seconds, with IACTs 5.264, 4.603, and 4.292.

Both combined checkpoints strictly improve over their hash controls in all
16 conditions and all eight paired-temperature outer aggregates. Hash/combined
IACT ratios are 1.581 (outer 95\% interval $[1.478,1.699]$) at 20\% and 1.495
($[1.379,1.626]$) at 40\%. All combined-release certificates occur at $\beta=0.8$:
5/8 and 8/8, versus 0/8 at $\beta=0.3$. This gives a partial-budget realization
point with temperature-dependent coverage. Independent checks verify all
480 kernels and 288 release steps (Appendix~\ref{app:adaptive-confirmation}).

\paragraph{Remaining IACT suboptimality.}
A stage current upper bound $U_t$ and dual-feasible $(z,y)$ on every edge of
$\mathcal E_+=\mathcal E_0\cup S$, including frozen old edges, give
$D_t=(q^\top z)^2/(s^\top y)$ when $s^\top y>0$. Then
$0\le\tau(c_t)-\tau^\star_+\le2(U_t-D_t)/V$.
Relative certificates use $U_{\tau,t}/L_{\tau,+}-1$, with
$U_{\tau,t}=2U_t/V-1$ and $L_{\tau,+}=2D_t/V-1>0$
(Appendix~\ref{app:stage-gap-extension}). On the original 24 development
conditions, at 40\% release, 9/24 are
certified within 10\% of the optimum and 5/24 within 5\%. Charging all preceding
certificate queries in a noncontemporaneous timing replay leaves 1/24 cheaper
than full weighting at 10\%, and 0/24 at 5\%
(Appendix~\ref{app:gap-audit}).

\section{When Does Weight Construction Pay Off?}
\label{sec:cost}

\paragraph{Stationary sampling and amortization.}
For target effective sample size (ESS) $M$, the stationary cost model is
\begin{equation}
 T(M)=T_{\rm setup}+M\tau_f t_{\rm step}.
 \label{eq:cost-model}
\end{equation}
Setup includes candidate construction, cumulative weighting, kernel
materialization, and sampler initialization. Ten balanced-order repeats of
64 stationary-start chains with 32,768 steps measure $t_{\rm step}$.
The curves extrapolate these timings using fixed-kernel IACT bounds.

All three ledgers include the optimized base kernel. Starting before witness
generation, the \emph{joint proof-and-kernel ledger} charges the original-support
solve, which supplies both a lower witness and a usable base kernel. Each
alternative pays for that solve once. Combined 20\% has a cheapest-cost window
in \textbf{1/16} conditions, versus 14/16 among augmented kernels alone.

With the base lower witness supplied, the \emph{sampling-setup ledger} charges
each kernel's construction and sampling; the \emph{verification ledger} adds
independent witness checks. Combined 20\% has a cheapest-cost window in 14/16
conditions under both ledgers. Median pairwise crossover ESS is
$1.09\times10^7$ against degree-MH and $3.96\times10^7$ against full weighting.
Full weighting occupies the long-run end of all 16 cost envelopes
(Appendix~\ref{app:amortization-confirmation}).

\paragraph{Measured finite-budget estimation.}
A separate frozen ring-Ising study uses five new coefficient seeds,
$\beta=0.3$, and 1024, 2048, and 4096 states. Its partial-weighting methods use
the earlier combined policy at 20\%. Median effective supports are
391, 749, and 1260. At six budgets from 1 to 360 seconds, we measure the
mean squared error of 32 chain averages per start, normalized by target
variance. All-plus, all-minus, and uniform-state starts share one kernel in
a 96-chain workload; initial transients remain in the estimator. Construction,
candidate enumeration, and sampling are timed. An additional ledger charges
fixed-chain verification. The original-support optimal lower-bound solve
belongs to the separate joint ledger above.

At 360 seconds, energy MSE ratios for base degree-MH over LEF combined 20\% are
2.224--2.740 at 1024 states (5/5 seeds favorable for each start) and
1.480--2.270 at 2048 states (4/5 or 5/5). Table~\ref{tab:finite-budget}
includes the fully weighted base support. For uniform starts, its energy MSE
ratio is 0.598 (outer 95\% interval $[0.475,0.890]$) at 1024 states, favoring
the fully weighted base. At 2048 states the ratio is 1.048
($[0.614,1.730]$), leaving the ranking unresolved. Both full-weighting methods
have lower energy error at 1024 states, whereas partial weighting has lower
magnetization error.
At 2048 states, all-minus magnetization favors base degree-MH in all five
seeds (base/combined ratio 0.730). The capacity comparison varies by start.
Enumerated-target independent sampling has lower error than LEF combined 20\% across
the tested 360-second starts and observables. The new 256-state factorial
experiment isolates fixed-chain IACT and construction cost; it does not change
these measured-budget MSE comparisons. At 4096 states, all four optimization methods provide
zero budget-available outputs: 17/20 builds time out at 900 seconds and
the remaining three exceed 360 seconds. Appendix~\ref{app:finite-budget}
reports every budget, absolute errors, uncertainty, and resource costs;
Appendix~\ref{app:scaling} gives the earlier construction-only scale study.

\begin{table}[t]
\caption{Measured 360-second mean-estimation errors, including construction.
Uniform-state starts; five-seed geometric ratios of comparator MSE to
LEF combined 20\% MSE. Values above one favor LEF combined 20\%.
Full weighting uses its returned feasible kernel.}
\label{tab:finite-budget}
\centering\footnotesize
\setlength{\tabcolsep}{3pt}
\begin{tabular}{lrrrr}
\toprule
& \multicolumn{2}{c}{Energy} & \multicolumn{2}{c}{Magnetization} \\
Comparator & 1024 & 2048 & 1024 & 2048 \\
\midrule
Base degree-MH & 2.740 & 1.480 & 1.335 & 0.970 \\
Base full & 0.598 & 1.048 & 5.694 & 13.218 \\
Capacity combined 20\% & 0.858 & 0.851 & 1.070 & 1.071 \\
LEF full & 0.695 & 0.779 & 4.804 & 7.723 \\
Independent & 0.206 & 0.136 & 0.034 & 0.022 \\
\bottomrule
\end{tabular}

\end{table}

\section{Interpreting Edge Value and Flow Reallocation}
\label{sec:information}

\subsection{Exact Value Under Conductance Insertion}

The Poisson potential $g=A^\dagger q$ solves $(I-P)g=h$; its solutions differ
by an additive constant. It accumulates long-timescale information, while LEF
uses raw endpoint energy contrast as a local heuristic.

\begin{proposition}[Reciprocal insertion gain]
\label{prop:rank-one}
Let $P$ be finite, irreducible, and reversible, with $P_{uu}\ge\ell>0$.
For $e=(u,v)$, choose
\begin{equation}
 0<\delta_e\le\min\{\pi_u(P_{uu}-\ell),\pi_v(P_{vv}-\ell)\}.
 \label{eq:insertion-budget}
\end{equation}
Move this stationary flow from the two endpoint self-loops to the reciprocal
edge, holding every old off-diagonal conductance fixed. Then
\begin{equation}
 \sigma_f^2(P)-\sigma_f^2(P_e)
 =\frac{2\delta_e[g(u)-g(v)]^2}{1+\delta_eR_e},
 \quad R_e=b_e^\top A^\dagger b_e.
 \label{eq:rank-one}
\end{equation}
\end{proposition}

Sherman--Morrison on the mean-zero subspace proves the identity
(Appendix~\ref{app:proof}). $R_e$ is resistance in the stationary-flow network;
endpoint contrasts are normalization-invariant. LEF's capacity and raw energy
contrast are local heuristics motivated by the numerator's structure.

The identity matches 232 conditional edges to relative error
$3.67\times10^{-11}$; exact greedy nearly closes global-search headroom on
the tested 16-edge, four-bridge shortlist (Appendix~\ref{app:context}).

The insertion formula describes a fixed-kernel intervention. For the optimized
support comparison, Proposition~\ref{prop:support-criterion} tests whether any
original optimal dual remains feasible on all new edges.

\subsection{Accounting for the Actual Degree-MH Intervention}

Degree-MH reconstruction also changes old flows. By
Equation~\ref{eq:degree-flow}, adding edges weakly decreases every old
conductance. Write the new-edge Laplacian as $N\succeq0$ and the old-flow loss
Laplacian as $R\succeq0$, so $A_1=A_0+N-R$. On the feasible interpolation
$A_t=A_0+t(N-R)$, let $g_t=A_t^\dagger q$. Differentiating
Equation~\ref{eq:variance} on the mean-zero subspace gives
\begin{equation}
 \begin{aligned}
 \sigma_f^2(A_0)-\sigma_f^2(A_1)
 &=\underbrace{2\int_0^1g_t^\top N g_t\,dt}_{B_{\rm new}}\\[-1mm]
 &\quad-\underbrace{2\int_0^1g_t^\top R g_t\,dt}_{L_{\rm old}}.
 \end{aligned}
 \label{eq:flow-decomposition}
\end{equation}
Both terms are nonnegative and account for the actual endpoints along this
specified conductance path.

All 80 LEF/capacity interventions on the original 40 cells close against
independent high-precision endpoint solves, with maximum error
$1.22\times10^{-72}$. Old-flow losses reverse one LEF/capacity ranking;
the other four capacity-favorable cases already prefer capacity in their
new-flow term (Appendix~\ref{app:flow-accounting}). This distinction explains
why useful support and effective degree-MH reconstruction require separate
evaluation, and motivates adjusting existing flows after selection.

\section{Related Work and Implications}

Locally balanced proposals exploit local target information
\citep{zanella2020informed}; gradient-based, any-scale, and second-order discrete
methods enlarge the design space
\citep{grathwohl2021oops,sun2023anyscale,zhou2026pdhams}. DISCS provides common
evaluations of discrete samplers~\citep{goshvadi2023discs}. We compare support
changes under common constraints, using validity-aware molecular proposals and
natural single-spin Ising supports.

FMMC optimizes spectral convergence on a fixed graph
\citep{boyd2004fastest}, with geometric barriers applying across all weights
\citep{oleskertaylor2024geometric}. Average-observable variance-optimal transition
matrices are studied by \citet{chen2012optimal}. Peskun ordering relates increased off-diagonal mass to
reduced asymptotic variance~\citep{peskun1973optimum,tierney1998note}, and the
electrical-network representation connects conductance, resistance, and
Poisson solutions~\citep{doyle1984random}. Coupled-chain methods estimate Poisson
solutions and asymptotic variances without a full operator
\citep{douc2026poisson}. Our checkable witnesses compare supports at common
target and budget; the path identity accounts for signed degree-MH flow changes.

The weighting program specializes resource-constrained $c$-optimal design
\citep{sagnol2011design} to edge incidence vectors $b_e$, contrast $q$, and
node budgets $|B|c\le s$, where $|B|$ is the unsigned incidence matrix.
Resistance allocation and pseudoinverse-based augmentation provide related
graph optimization tools \citep{ghosh2008resistance,predari2023greedy}.
Our remaining-value bounds build on duality gaps and design screening
\citep{jaggi2013revisiting,pronzato2021removing,sagnol2023screening}.

\citet{duan2025graph} combine approximate-sample graph jumps with a baseline
kernel, analyze conductance improvement, and discuss spectral weight optimization.
Our benchmark compares observable variance with every admissible original-support
weighting and measures the cost of realizing the selected support's value.

Molecular graph generators supply structured settings
\citep{vignac2023digress,kim2025ddsbm,piening2026gromov}.
Appendix~\ref{app:dynamic-refresh} reports the limits of the
trajectory-driven discovery pilot.

\section{Conclusion}

An exact dual criterion characterizes when added edges improve the best
variance for a fixed observable. Finite witnesses certify individual gains,
and a uniform-target family gives an unbounded single-edge gain despite a
persistent slow spectral mode. The Ising experiments show how release history
and construction cost affect the gain realized on enumerated targets.
\label{main:end}

\section*{Impact Statement}
This work studies the reliability and interpretation of sampling algorithms for
discrete probabilistic models. Its certificates and independent numerical checks
support more transparent comparisons of estimator efficiency. Applications to
scientific models require choosing observables and computational budgets that
reflect the downstream estimation task.

\bibliography{references}
\bibliographystyle{icml2026}
\clearpage
\appendix
\renewcommand{\thesection}{\AlphAlph{\value{section}}}
\renewcommand{\theHsection}{appendix.\arabic{section}}
\onecolumn
\section{Certificate Construction and Independent Verification}
\label{app:weighting-certificate}

\subsection{The exact finite problem}

For each stored floating-point probability or observable value $x$, the exact
input is the rational represented by its shortest round-trip decimal string,
implemented as \texttt{Fraction(str(float(x)))}. The probability vector is then
normalized exactly. The same rational target is used to center the observable,
form $q_u=\pi_u h_u$, and compute $V=\sum_u\pi_u h_u^2>0$ on every support of a
cell. The common self-loop floor is $\ell=1/20$, giving node capacities
$s_u=19\pi_u/20$. This defines the finite target certified by the witnesses.

Write $J(c)=q^\top A(c)^\dagger q$. The primal in Equation~\ref{eq:primal}
minimizes $J$ over conductances and conserved currents. Its potential dual is
\begin{equation}
 \max_{z\in\mathbb R^{|\mathcal U|},\,y\ge0}\ 2q^\top z-s^\top y,
 \qquad (z_u-z_v)^2\le y_u+y_v\quad((u,v)\in\mathcal E),
\end{equation}
where only $y$ is constrained nonnegative and $z$ is unrestricted. Equivalently,
the maximization is over $z\in\mathbb R^{|\mathcal U|}$ and
$y\in\mathbb R_+^{|\mathcal U|}$. Feasible witnesses supply $L_J\le J^*\le U_J$;
the reported IACT bounds use
\begin{equation}
 L_\tau=(2L_J-V)/V,\qquad U_\tau=(2U_J-V)/V.
 \label{eq:certificate-iact}
\end{equation}
Every zero-conductance edge carries zero current. The Thomson sum is evaluated
only over edges with strictly positive conductance.

\subsection{Numerical proposals and rational repairs}

Node and objective scaling stabilize the convex programs. Numerical solutions
are proposals for witnesses. Rational conversion and repair enforce the node
budgets, all dual inequalities, and $Bj=q$ exactly. A positive connected
degree-MH mixture retains irreducibility; current divergence is repaired on a
spanning tree. The independent checker uses Python's standard-library rational
arithmetic to reconstruct the target, check feasibility, evaluate both bounds,
and compare exact fractions. Solver status and bound tightness are separately
recorded.

The molecular schedule includes all 124 cells and their base, LEF, and capacity
supports, yielding 372 problems. Each has a fixed 120-second CPU cap; four workers
each use one BLAS thread. All 372 produce valid bounds. One LEF checkpoint
support has a wider electrical-energy gap, $0.0014351$, with IACT interval
$[398.9502809,399.5250745]$; the remaining 371 gaps are at most $10^{-4}$.
The numerical proposal for the wider interval uses the prescribed SCS fallback.

The Ising schedule includes 24 cells and four supports each, yielding 96 problems.
It uses ECOS first, with tolerances $10^{-9}$ and 300 iterations, then SCS
($10^{-7}$, 5,000 iterations) when the bound gap exceeds $10^{-4}$. The per-support
CPU cap is 180 seconds. An equilibrated Cholesky current proposal with two
refinement steps precedes a fixed 60-digit fallback; a $10^{-8}$ connected
degree-MH mixture preserves positive base flows. All 96 witnesses verify.
The LEF support of ferromagnetic instance 2026101604 at $\beta=0.8$ has gap
0.00565267 and IACT interval $[47.039605,47.312700]$; its base lower bound
53.160176 still gives strict separation. The other 95 gaps are at most $10^{-4}$.

Two-state and three-state analytically solvable cases test the bound directions
and objective scaling. Separate tests check shared endpoint capacities, current
repair, and rejection of invalid witnesses. Independent verification covers all
468 saved support problems. The molecular and Ising witness-generation runs take
221.9 and 90.7 seconds, respectively, on CPU; the saved witnesses support
verification independently of the optimization environments.

\begin{table}[ht]
\centering\small
\caption{Development support-optimal IACT intervals. Each row uses one common
rational target and energy observable. Rounding is for display; all separations
are computed from exact fractions.}
\begin{tabular}{lrrrr}
\toprule
Seed & Base optimum & Direct LEF upper & Optimized LEF & Optimized capacity\\
\midrule
0 & [9.715350,9.715361] & 31.389932 & [5.738051,5.738057] & [7.116234,7.116239]\\
1 & [548.221006,548.221611] & 123.999952 & [30.021484,30.021496] & [30.408813,30.408825]\\
2 & [835.059575,835.060517] & 463.048736 & [84.269815,84.269962] & [89.560063,89.560233]\\
3 & [420.761980,420.762137] & 72.409725 & [10.262866,10.262878] & [10.262866,10.262878]\\
\bottomrule
\end{tabular}
\end{table}

\subsection{Capacity-matched contrast diagnostics}

Within each development component, candidate nonedges are grouped by exact
equality of the LEF log-capacity feature, hash ordered, and paired without
replacement. Concordance compares squared energy contrast with squared Poisson
contrast, exact rank-one gain, and actual degree-MH gain; ties receive credit
0.5. These pairs are nested within each development component.

\begin{table}[ht]
\centering
\caption{Development capacity-matched concordance. Matching fixes the local
capacity feature; individual pairs can still have different feasible insertion
flows $\delta$.}
\begin{tabular}{lrrrr}
\toprule
Seed & Pairs & Poisson contrast & Rank-one gain & Degree-MH gain\\
\midrule
0 & 876 & 0.632 & 0.630 & 0.541\\
1 & 873 & 0.688 & 0.699 & 0.619\\
2 & 877 & 0.698 & 0.704 & 0.552\\
3 & 883 & 0.651 & 0.656 & 0.425\\
\bottomrule
\end{tabular}
\end{table}

Capacity differences are exactly zero; the median log insertion-flow difference
is also zero in every component. The held-out 2,560-pair analysis uses its own
frozen protocol and has concordance 0.374. The difference between these cohorts
motivates explicitly evaluating the conditions under which the local contrast
is informative.

\section{Structural Separation and Remaining-Value Bounds}
\label{app:structural-extension}

This appendix develops the fixed-target, fixed-observable support comparison in
three directions: an explicit diffuse-target separation family, a bound on the
value remaining after a partial weighting update, and the role of node-budget
prices in admitting a new edge. Write
$J(c)=q^\top A(c)^\dagger q$, so that
$\tau(c)=2J(c)/V-1$.

\subsection{One Bridge Can Separate Uniform-Target Supports}
\label{app:uniform-separation}

Classical bottleneck geometry limits the fastest reweighting of a fixed support
\citep{roch2005bounding,oleskertaylor2024geometric}; optimal conductance
allocation also yields size-dependent resistance effects
\citep{ghosh2008resistance}. The following construction gives explicit
fixed-observable bounds under stationary node budgets.

\paragraph{Construction and bound.}
Take two disjoint $m$-cliques, designate one hub in each, and join the hubs by a
path of $L$ edges with $L-1$ new internal vertices. Let $m\ge2$, $L\ge2$,
$n=2m+L-1$, $\pi_u=1/n$, and $0<\ell<1$. Every node has budget
$s=(1-\ell)/n$. Set $f=1$ on the first clique, $f=-1$ on the second, and
$f=0$ on the internal path vertices. Then $\pi(f)=0$, $q=f/n$, and
$V=2m/n$. Over every admissible irreducible reversible weighting of the base
support,
\begin{equation}
 \inf_{c\text{ on }\mathcal E_0}\tau(c)
 \ge \frac{Lm}{1-\ell}-1.
 \label{eq:uniform-base-lower}
\end{equation}
After adding the hub--hub edge, an explicit feasible connected weighting obeys
\begin{equation}
 \tau(c_+)\le
 \frac{8(m-1)^2+4m^2}{m(1-\ell)}-1.
 \label{eq:uniform-aug-upper}
\end{equation}

\paragraph{Proof.}
Each path edge separates all $m$ positive sources from all $m$ negative sinks.
Conservation therefore forces any feasible current to carry magnitude $m/n$
along each of these $L$ edges. The node budgets give $c_e\le s$, hence Thomson's
principle yields
$J(c)\ge L(m/n)^2/s=Lm^2/[n(1-\ell)]$.
Conversion by $\tau=2J/V-1$ proves Equation~\ref{eq:uniform-base-lower}.

For the augmented weighting, assign conductance $s/4$ to the new hub--hub edge
and to every path edge. In each clique, retain the hub star with conductance
$s/[4(m-1)]$ per star edge, and assign zero to the remaining clique edges.
Each hub uses $3s/4$, each internal path vertex uses $s/2$, and every other
vertex uses $s/[4(m-1)]$. The positive-conductance support is connected and all node
budgets hold. A feasible current sends $1/n$ from each positive non-hub vertex
to its hub, $m/n$ through the new edge, and $1/n$ from the opposite hub to each
negative non-hub vertex; the path currents are zero. Its energy is
\begin{equation}
 \sum_e\frac{j_e^2}{c_e}
 =\frac{8(m-1)^2+4m^2}{n(1-\ell)}.
 \label{eq:uniform-current-energy}
\end{equation}
This feasible current proves Equation~\ref{eq:uniform-aug-upper}.

For $L=m$, $n=3m-1$, the effective support is exactly $n$, and
$V=2m/(3m-1)\to2/3$. The ratio between the base optimum and this augmented
chain is at least
\begin{equation}
 \frac{m^2/(1-\ell)-1}
 {[8(m-1)^2+4m^2]/[m(1-\ell)]-1}
 \sim\frac{m}{12}\sim\frac{n}{36}.
 \label{eq:uniform-separation-ratio}
\end{equation}
The displayed bounds strictly separate for every integer $m\ge11$.
Thus one extra transition can provide an $\Omega(n)$ support advantage under
a uniform target and a nondegenerate observable. The construction isolates
the path bottleneck and uses the same node budgets on both supports.

\paragraph{Observable improvement and spectral relaxation.}
The same augmented support retains a slow corridor direction. For $L\ge3$,
number its internal vertices $v_1,\ldots,v_{L-1}$ and set
$\psi(v_k)=\sin(2\pi k/L)$, with $\psi=0$ on both cliques. Elementary discrete
Fourier identities give $\pi(\psi)=0$, $\operatorname{Var}_\pi(\psi)=L/(2n)$,
and the sum of squared differences along the corridor equal to
$2L\sin^2(\pi/L)$. Every edge conductance is at most $s$; all other edges,
including the shortcut, have zero $\psi$ contrast. The Rayleigh quotient
therefore gives, for every admissible augmented weighting,
\begin{equation}
 1-\lambda_2(P)\le4(1-\ell)\sin^2(\pi/L).
 \label{eq:uniform-spectral-bound}
\end{equation}
Here $\lambda_2(P)$ is the second-largest eigenvalue of the reversible kernel.
For $L=m\to\infty$ at fixed $\ell$, the relaxation time $(1-\lambda_2)^{-1}$ is $\Omega(n^2)$,
whereas the explicit chain above has clique-sign IACT $O(n)$. Thus the family
also separates improvement of the specified observable from worst-direction
spectral relaxation. Related long-path barriers under a single-edge intervention
appear in \citet{gerencser2023singleedge}, whose acceleration result uses a
non-reversible kernel.

\subsection{A Stage Certificate for Remaining Weighting Value}
\label{app:stage-gap-extension}

Fix the entire selected support $\mathcal E_+=\mathcal E_0\cup S$ and write
$J_+^\star$ for its optimal value. At a partial-weighting stage, let $c_t$ be
feasible and connected, and let $j_t$ satisfy $Bj_t=q$, with zero current on
zero-conductance edges. Define $U_t=\sum_{e:c_{t,e}>0}j_{t,e}^2/c_{t,e}$.
Choose $z$ and $y\ge0$ satisfying
\begin{equation}
 (z_u-z_v)^2\le y_u+y_v
 \quad\text{on every }(u,v)\in\mathcal E_+.
 \label{eq:stage-full-support-feasibility}
\end{equation}
Put $a=s^\top y$ and $b=q^\top z$. Scaling the feasible dual witness as
$(z,y)\mapsto(\alpha z,\alpha^2y)$ gives lower objective
$2\alpha b-\alpha^2a$. Its maximum is
\begin{equation}
 D_t=\begin{cases}b^2/a,&a>0,\\0,&a=0,\end{cases}
 \qquad
 0\le\tau(c_t)-\tau_+^\star
 \le G_t:=\frac{2}{V}(U_t-D_t).
 \label{eq:stage-gap-bound}
\end{equation}
When $a=0$, positive budgets imply $y=0$; connectivity then makes $z$ constant
and $b=0$, and we take $\alpha=0$. For $a>0$, the maximizing scale is $\alpha=b/a$.
Weak duality gives $D_t\le J_+^\star\le J(c_t)\le U_t$, proving the bound.

The electrical-energy interval has the exact conceptual decomposition
\begin{equation}
 \underbrace{U_t-D_t}_{\text{reported interval}}
 =\underbrace{U_t-J(c_t)}_{\text{current-bound error}}
 +\underbrace{J(c_t)-J_+^\star}_{\text{remaining weighting value}}
 +\underbrace{J_+^\star-D_t}_{\text{dual slack}}.
 \label{eq:stage-gap-three-terms}
\end{equation}
All three terms are nonnegative. The computed $G_t$ requires the current
stage, its feasible current and potential, node budgets, and checks over the
entire selected support. A full-weighting reference can subsequently help
resolve the three terms using its own interval. In particular, a broad stage
interval can reflect either substantial remaining value or a loose dual.

There is also a directly checkable residual identity. For the scaled witness
$(\widetilde z,\widetilde y)=(\alpha z,\alpha^2y)$, let
$d_e=b_e^\top\widetilde z$. Then
\begin{align}
 U_t-D_t={}&
 \sum_{e:c_{t,e}>0}\frac{(j_{t,e}-c_{t,e}d_e)^2}{c_{t,e}}
 +\sum_u\widetilde y_u
       \left(s_u-\sum_{e\ni u}c_{t,e}\right)\nonumber\\
 &+\sum_{e=(u,v)}c_{t,e}
       (\widetilde y_u+\widetilde y_v-d_e^2).
 \label{eq:stage-gap-residual-identity}
\end{align}
The identity follows by expansion and $Bj_t=q$; every term is nonnegative.
These bounds describe the quality of a given partial update and a possible
stopping tolerance in IACT units. Their convex-design and duality-gap background
is developed by \citet{sagnol2011design} and \citet{jaggi2013revisiting}.

\subsection{Budget Prices, Dual Degeneracy, and Candidate Bounds}
\label{app:priced-support-extension}

For a connected support $\mathcal E$ and positive budgets, the full weighting
problem has dual
\begin{equation}
 J_{\mathcal E}^\star=
 \max_{z,\,y\ge0}\left\{2q^\top z-s^\top y:
 (b_e^\top z)^2\le y_u+y_v,\ e=(u,v)\in\mathcal E\right\}.
 \label{eq:priced-support-dual}
\end{equation}
Here $z$ is unrestricted. The closed perspective objective uses $0/0=0$ and
$j^2/0=+\infty$ for $j\ne0$. Strictly positive, budget-interior conductances
and sufficiently large epigraph variables give conic strict feasibility, hence
strong duality and dual attainment. The closed primal optimum equals the
infimum over connected positive-conductance supports: mix an optimal feasible
conductance vector with any strictly positive feasible one, retaining its
current, and let the mixing weight tend to zero.

Let $\mathcal D_{\mathcal E}^\star$ denote the optimal dual set with
$\mathbf1^\top z=0$. At every optimum, scaling stationarity gives
$q^\top z=s^\top y=J_{\mathcal E}^\star$. Since $s_u>0$, this bounds every
$y_u$; the edge constraints and connectivity bound the centered $z$.
Thus $\mathcal D_{\mathcal E}^\star$ is nonempty and compact. Define the
budget-priced contrast
$v_e(z,y)=(z_u-z_v)^2-y_u-y_v$ for a candidate edge.
Proposition~\ref{prop:support-criterion} characterizes equality by the
existence of an original optimal dual satisfying $v_e(z,y)\le0$ for all
$e\in S$. For a finite nonempty candidate set $S$, compactness gives the
equivalent strict-improvement criterion
\begin{equation}
 J_{\mathcal E\cup S}^\star<J_{\mathcal E}^\star
 \iff
 \min_{(z,y)\in\mathcal D_{\mathcal E}^\star}\max_{e\in S}v_e(z,y)>0.
 \label{eq:priced-support-strict}
\end{equation}
The condition compares potential contrast with the opportunity cost of node
capacity. Its quantifier over all optimal duals handles degeneracy.

\paragraph{A five-node degeneracy example.}
On the path $1$--$2$--$3$--$4$--$5$, take unit conductances,
$s=(1,2,2,2,1)$,
$q=(1,0,-9/5,0,4/5)$, and
$z=(2,1,0,4/5,8/5)$. The primal value is $82/25$. Every
\begin{equation}
 y(t)=(t,1-t,t,16/25-t,t),\qquad 0\le t\le16/25,
 \label{eq:priced-degeneracy-example}
\end{equation}
is dual optimal with that same value. The proposed edge $(1,5)$ has
$(z_1-z_5)^2=4/25$. Its constraint is violated at $t=0$ and satisfied at
$t=2/25$. The latter witness proves that adding the edge preserves the optimum.
Thus a violation at one optimal dual is compatible with zero support gain.
For a Markov-chain realization, take $\pi=s/8$, $\ell=1/2$, and divide both
the conductances and $q$ by $16$; the same potentials and dual family apply.

\paragraph{Safe bounds from finite witnesses.}
Let $U_0$ be any original-support primal upper bound and $(z,y)$ any feasible
original dual, with $L_0=2q^\top z-s^\top y$. For candidate set $S$, choose
$\eta\ge0$ satisfying $\eta_u+\eta_v\ge[v_e(z,y)]_+$ on every $e\in S$.
Then $(z,y+\eta)$ is augmented-dual feasible, and
\begin{equation}
 0\le J_{\mathcal E}^\star-J_{\mathcal E\cup S}^\star
 \le U_0-L_0+s^\top\eta.
 \label{eq:priced-safe-gain}
\end{equation}
Minimizing $s^\top\eta$ under these linear constraints defines a tighter
repair cost $\rho(S)$. For one edge,
$\rho(\{(u,v)\})=\min(s_u,s_v)[v_e]_+$.
Summing single-edge repairs is feasible for a set; consequently, for any
selection of at most $K$ candidates, a uniform upper bound is $U_0-L_0$ plus
the $K$ largest single-edge repair costs. These are bounds on the possible
benefit of further support changes, with corresponding IACT bounds obtained
by multiplication by $2/V$. They specialize the same convex-design dual
structure to candidate bounds; design-point screening has direct precedents
\citep{pronzato2021removing,sagnol2023screening}. A small single-edge bound
controls its standalone gain; a multi-edge selection uses the corresponding
set bound to retain interactions.

\subsection{Development Audit of Remaining Optimality Certificates}
\label{app:gap-audit}

This audit evaluates Equation~\ref{eq:stage-gap-bound} using the saved 24 Ising
development conditions: 12 coefficient instances, each at $\beta=0.3,0.8$.
Each condition retains its original target, observable, selected LEF support,
conductance budgets, and frozen release rule. The audit has nine saved stages:
degree-MH, endpoint update, adaptive and hash release at 10\%, 20\%, and 40\%,
and the full-weighting feasible reference. The eight independent confirmation
instances in Appendix~\ref{app:adaptive-confirmation} retain their original role.

\paragraph{Stage information and exact validity.}
Each stage supplies its own saved fixed-chain potential $z$, feasible current,
and conductances. After centering and scaling $z$, we solve the node-price LP
\begin{equation}
 \min_{y\ge0}s^\top y
 \quad\text{subject to}\quad
 y_u+y_v\ge(z_u-z_v)^2,\qquad (u,v)\in\mathcal E_+.
 \label{eq:gap-audit-lp}
\end{equation}
All selected-support edges enter these constraints, including frozen old edges.
Floating-point prices are converted to rational values; any positive constraint
deficit is assigned to its cheaper-budget endpoint in canonical edge order.
Independent rational checks verify the target, centering, current conservation,
budgets, connected positive-conductance support, full-support dual feasibility,
optimal scale, and the residual identity in
Equation~\ref{eq:stage-gap-residual-identity}. Certification is with respect to
the stored rational target, following the original witness convention.

The protocol fixes one LP attempt per stage and repeat, solver tolerances,
the stage order, and seven descriptive precision thresholds before execution.
The complete run comprises 648 LP records (24 conditions, nine stages, three
timing repeats), all independently verified. Computation uses one CPU core and
reuses cached energies and kernels. Shared reading, hashing and problem
preparation cost a median 0.025 seconds per condition. Table~\ref{tab:gap-audit}
includes stage preparation, LP, rational repair and checking in each query.
The LP itself costs approximately 0.034 seconds at adaptive 20\% and 40\%.

\begin{table}[t]
\caption{\textbf{Remaining optimality, development conditions.} $G_t$ is an
absolute upper bound on remaining IACT suboptimality. The last columns count certificates
$U_{\tau,t}/L_{\tau,+}-1\le\varepsilon$ with $L_{\tau,+}>0$, requiring all three
repeats to pass. Medians summarize the 24 conditions. Full weighting is a
post-hoc reference; the stopping replay uses degree, endpoint, and adaptive
stages only.}
\label{tab:gap-audit}
\centering
\small
\begin{tabular}{lrrrr}
\toprule
Stage & Median $G_t$ & Query (s) & $10\%$ & $5\%$ \\
\midrule
Degree & 12.645 & 0.101 & 0/24 & 0/24 \\
Endpoint & 7.487 & 0.114 & 0/24 & 0/24 \\
10\% & 6.299 & 0.120 & 0/24 & 0/24 \\
20\% & 2.689 & 0.137 & 0/24 & 0/24 \\
40\% & 1.131 & 0.177 & 9/24 & 5/24 \\
Hash 40\% & 4.039 & 0.176 & 0/24 & 0/24 \\
Full (reference) & $6.93\!\times\!10^{-6}$ & 0.342 & 24/24 & 24/24 \\
\bottomrule
\end{tabular}

\end{table}

\paragraph{Precision and residual value.}
Adaptive 20\% reaches none of the seven precision thresholds. At 40\%, the
absolute tolerances $G_t\le0.1,0.5,1.0$ certify respectively 0, 6 and 11 of
24 conditions; relative-to-optimum tolerances of 10\% and 5\% certify 9 and 5.
All passing partial-stage conditions have $\beta=0.8$. A relative-to-current-upper
criterion, $G_t/U_{\tau,t}\le10\%$ or $5\%$, gives 10 and 5 conditions; its
denominator differs from the optimum-relative guarantee.

The largest fixed-chain IACT interval width across all records is
$3.66\times10^{-25}$. By comparison, adaptive 20\% and 40\% median remaining
bounds are 2.689 and 1.131. To interpret their conservatism, full-weighting
witnesses are loaded after all stage certificates are complete. Differences
to the full \emph{feasible chain} have medians 1.431 and 0.405 IACT units.
An independently bracketed full-support optimum gives median upper bounds
1.131 and 0.472 for the stage dual slack. These quantities separate negligible
fixed-chain numerical width from both remaining optimization and lower-bound
slack; the full feasible chain serves as a reference inside its certified
interval. Full-stage potentials give a median remaining bound
$6.93\times10^{-6}$ and meet all seven thresholds in all 24 conditions.

\begin{figure*}[t]
\centering
\includegraphics[width=\textwidth]{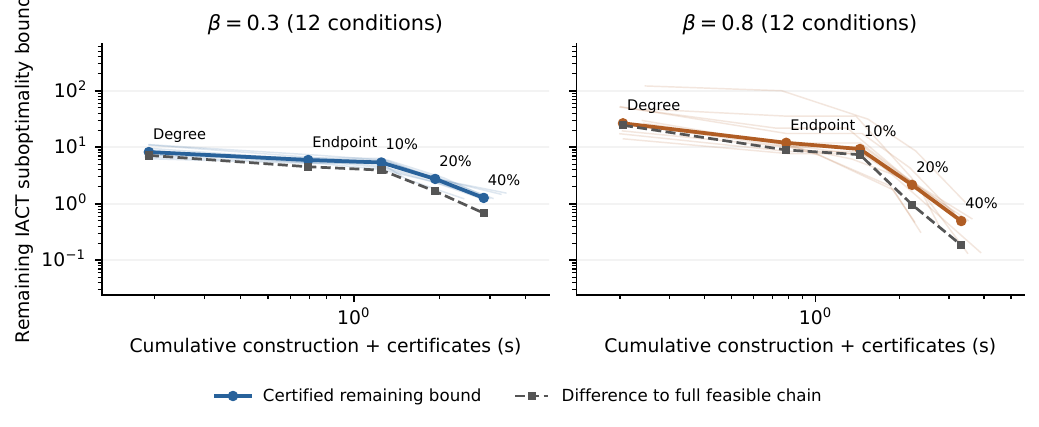}
\caption{\textbf{Cost and remaining IACT on the original development conditions.}
Faint curves show each condition; solid curves connect stage-wise medians of
certified bounds and cumulative time. Dashed curves show differences to the
post-hoc full feasible chain. Each time includes the saved construction cost,
shared input preparation, and every certificate query through that stage.
Construction and query times were measured in separate runs. Median curves
summarize different condition distributions; the full reference is used only
to interpret the stage bounds.}
\label{fig:remaining-gap}
\end{figure*}

\paragraph{Stopping replay and decision.}
Replay visits degree-MH, endpoint, 10\%, 20\%, and 40\%, charging every preceding
query before testing the fixed tolerance. Every trigger occurs at 40\%;
cumulative certificate overhead there has median 0.676 seconds. Under the
10\% optimum-relative threshold, one of the nine triggers is cheaper than
historical full weighting in all three repeats. The five 5\% triggers have
higher cumulative costs. The same single condition accounts for all cheaper
triggers under the other tolerances. Its construction-plus-query times are
2.260--2.271 seconds, compared with the historical full solve's 2.633 seconds.
This comparison combines historical construction with newly measured query
times. The development decision is to retain the error diagnostic and the
existing frozen release algorithm, completing this stopping-rule prototype
at the evaluation stage.

An accuracy rule $G_t\le\varepsilon$ has a direct mathematical guarantee.
Under the asymptotic ESS cost model, an economic stopping guarantee additionally
needs a fixed future ESS target
$M$, common step cost $t_{\rm step}$, and a valid lower bound on the next
optimization cost exceeding $M t_{\rm step}G_t$. The timing replay above
measures the empirical cost comparison with historical constructions.

\paragraph{Artifacts.}
The frozen protocol, per-stage rational witnesses, timing rows, post-hoc
intervals, stopping replay, independent verification and development decision
are in \texttt{outputs/support\_optimality\_gap\_v1/}. The standard-library
checker \texttt{experiments/verify\_support\_optimality\_gap\_v1.py} checks all
648 records independently of the proposer implementation. The structural
family and five-node example have a separate exact witness checker,
\texttt{experiments/verify\_structural\_extension\_v1.py}.

\section{Proof of Proposition~\ref{prop:rank-one}}
\label{app:proof}

Let $\Pi=\operatorname{diag}(\pi)$, $q=\Pi h$, and $A=\Pi(I-P)$.
Reversibility makes $A$ symmetric, and irreducibility gives null space
$\operatorname{span}\{\mathbf 1\}$. Because $h$ is centered,
$\mathbf 1^\top q=0$, so all following inverses can be taken on the Euclidean
mean-zero subspace. Define the Moore--Penrose solution $g=A^\dagger q$, which
satisfies $\mathbf 1^\top g=0$. If a $\pi$-centered Poisson solution is desired,
it is $g_\pi=g-(\pi^\top g)\mathbf 1$. Since
$b_e^\top\mathbf 1=0$, $b_e^\top g_\pi=b_e^\top g$, so the edge contrast is
independent of this additive normalization.

Moving stationary mass $\delta_e$ from the two endpoint self-loops to the
reciprocal edge changes the conductance Laplacian by
\begin{equation}
 A_e=A+\delta_e b_e b_e^\top.
\end{equation}
The stated upper bound on $\delta_e$ keeps both endpoint self-loop probabilities
at least $\ell$. The off-diagonal additions are
$P_e(u,v)-P(u,v)=\delta_e/\pi(u)$ and
$P_e(v,u)-P(v,u)=\delta_e/\pi(v)$; hence rows still sum to one and
$\pi(u)P_e(u,v)=\pi(v)P_e(v,u)$. Therefore $P_e$ is a valid reversible kernel
with invariant law $\pi$.

For simultaneous insertions, each endpoint additionally requires the cumulative budget
$\sum_{e\ni u}\delta_e\le\pi(u)[P(u,u)-\ell]$.

For a stationary discrete-time chain, the asymptotic variance of the centered
observable is
\begin{equation}
 \sigma_P^2(h)=2q^\top A^\dagger q-h^\top\Pi h.
\end{equation}
Since $b_e^\top\mathbf 1=0$, Sherman--Morrison applies on the mean-zero
subspace:
\begin{equation}
 A^\dagger-A_e^\dagger
 =\frac{\delta_e A^\dagger b_e b_e^\top A^\dagger}
        {1+\delta_e b_e^\top A^\dagger b_e}.
\end{equation}
The target and therefore $h^\top\Pi h$ are unchanged. Substitution gives
\begin{align}
 \sigma_P^2(h)-\sigma_{P_e}^2(h)
 &=2q^\top(A^\dagger-A_e^\dagger)q\\
 &=\frac{2\delta_e(b_e^\top A^\dagger q)^2}
         {1+\delta_e b_e^\top A^\dagger b_e}\\
 &=\frac{2\delta_e[g(u)-g(v)]^2}
         {1+\delta_e R_{\mathrm{eff}}(u,v)},
\end{align}
which proves the proposition.

\section{Unified Support-Optimal Selector Controls}
\label{app:optimized-controls}

This analysis uses the original 40 molecular checkpoint/component cells and
24 Ising cells. The outer units are five molecular training seeds and 12 Ising
coefficient instances; temperatures remain paired within an Ising instance.
Every selector adds four edges, followed by the same energy-observable weight
optimization with self-loop floor $\ell=0.05$. Molecular candidates comprise
all original nonedges; Ising uses the previously fixed 512 candidates. The
stored ensembles contain 64 random portfolios per molecular cell and 32 per
Ising cell. Energy-only and maximum-Hamming scores use canonical edge order
to break ties. All comparisons share each cell's exact rational target,
centered observable, variance, and node budgets.

For support $S$, write $[L_S,U_S]$ for verified bounds on its optimal IACT.
A control/LEF ratio lies in
$[L_{\rm control}/U_{\rm LEF},U_{\rm control}/L_{\rm LEF}]$.
Strict preference requires separated bounds. The Random comparator first takes
the geometric mean over every stored random portfolio within a cell; it is
an ensemble summary. Table~\ref{tab:updated-selector-controls} reports paired
effects. Ratios are averaged on the log scale within each outer unit, then
equally across outer units. Outer 95\% bootstrap intervals propagate the
numerical bounds: molecular intervals enumerate $5^5$ seed resamples;
six-instance Ising strata enumerate $6^6$ resamples; pooled Ising intervals use
100,000 fixed-seed resamples of coefficient instances with paired temperatures.
These analyses reuse the original outer units.

\begin{table}[ht]
\centering\small
\caption{Unified support-optimal comparisons. $R=\tau^*_{\rm control}/
\tau^*_{\rm LEF}$; values above one favor LEF. The final column gives counts
of strict LEF preference, strict control preference, identical support, and
overlapping bounds, respectively. Displayed point ratios are geometric
midpoints of numerical bounds.}
\label{tab:updated-selector-controls}
\begin{tabular}{llrrrl}
\toprule
Cohort & Control & $R$ & Outer 95\% interval & Cell median & Counts\\
\midrule
Molecular & Random ensemble & 58.795 & [25.040,154.300] & 19.106 & 38/0/0/2\\
 & Maximum-Hamming & 62.691 & [26.506,163.335] & 19.575 & 38/0/0/2\\
 & Energy-only & 62.922 & [26.541,163.826] & 19.589 & 38/0/0/2\\
 & Capacity & 1.976 & [1.213,4.308] & 1.0046 & 25/2/8/5\\
Ising & Random ensemble & 1.1413 & [1.1003,1.1867] & 1.1250 & 24/0/0/0\\
 & Maximum-Hamming & 1.1388 & [1.0918,1.1891] & 1.1306 & 23/1/0/0\\
 & Energy-only & 1.1337 & [1.0906,1.1817] & 1.1103 & 24/0/0/0\\
 & Capacity & 1.0142 & [0.9804,1.0442] & 1.0483 & 16/7/0/1\\
\bottomrule
\end{tabular}
\end{table}

Capacity is the closest comparator. In Ising, its optimized IACT relative to
LEF is 1.0843 and 1.0916 for ferromagnetic and mixed-sign instances at
$\beta=0.3$, with all 12 cells favoring LEF. At $\beta=0.8$, the ratios are
0.8934 and 1.0005; five of six ferromagnetic cells favor capacity. Molecular
capacity comparisons have a cell median of 1.0046 despite an aggregate ratio
of 1.976, locating much of the increment in a subset of cells.

Topology certification and selector quality answer complementary questions.
Random supports obtain strict topology certificates in 939/2,560 molecular
and 656/768 Ising portfolios. LEF is strictly better than 2,412/2,560 and
766/768 optimized random supports, respectively. The remaining molecular
comparisons comprise 24 random wins and 124 overlapping bounds; Ising has two
random wins. LEF/capacity/energy-only/Hamming topology counts are
38/38/0/21 out of 40 molecular cells and 24/23/24/19 out of 24 Ising cells.
Absolute optimized IACT medians for base, LEF, and capacity are
455.158, 1.342, and 2.397 in the molecular group, and 3.923, 3.466, and 3.532
in Ising.

The analysis added 3,456 optimizations and reused 184 distinct supports,
yielding 3,640 independently checked support witnesses. Electrical-energy
relative bound width is at most $10^{-4}$ for 3,576 supports; the largest
width among the remaining 64 is 0.01095335. All comparisons use their actual
bounds. New optimizations took 2,313.7 seconds with four CPU workers and one
BLAS thread per worker; endpoint energies were already cached.

\section{Endpoint-Neighborhood Weight Updates}
\label{app:local-update}

For each stored LEF or capacity four-edge set, form its augmented degree-MH
conductances $c^0$. Let $\mathcal F$ contain every augmented-support edge
incident to an endpoint of an added edge. Conductances outside $\mathcal F$
are fixed exactly to $c^0$. Optimize the Thomson objective
\begin{equation}
 \min_{c_{\mathcal F},j}\quad
 \sum_{e\in\mathcal F}\frac{j_e^2}{c_e}
 +\sum_{e\notin\mathcal F}\frac{j_e^2}{c_e^0},
 \qquad Bj=q,\quad c_e\ge0,\quad
 \sum_{e\ni u}c_e\le(1-\ell)\pi_u .
 \label{eq:local-flow-program-update}
\end{equation}
The conductance parameterization is local; conserved currents and the
observable objective use the complete finite graph. A rational repair scales
only free flows into the residual node budgets and mixes them with $10^{-8}$
of the reference flows. Remote conductances remain exactly equal to $c^0$.
The reference is retained as a feasible candidate throughout optimization.

For each returned chain, a rational potential supplies a fixed-chain
variational lower bound and a conserved current supplies a Thomson upper
bound. Thus $[L_{\rm local},U_{\rm local}]$ encloses the IACT of that specific
output. The local-family optimum is a separate quantity. Strict improvement
over the reference requires $U_{\rm local}<L_{\rm degree}$; strict topology
value requires $U_{\rm local}<L^*_{\rm base}$. Table~\ref{tab:updated-local}
reports these comparisons for the original 40+24 cells.

\begin{table}[ht]
\centering\small
\caption{Local conductance updates. The first three numeric columns are
absolute IACT medians; $D,L,F$ denote degree-MH, local update, and full
support optimization. Certificate counts use the same base-support lower
bound and the same added edges. Time is median local-update wall time.}
\label{tab:updated-local}
\begin{tabular}{llrrrrrr}
\toprule
Cohort & Selector & $\tau_D$ & $\tau_L$ & $\tau_F$
 & $U_L<L_D$ & Topology: $D/L/F$ & Time (s)\\
\midrule
Molecular & LEF & 38.002 & 3.660 & 1.342 & 40/40 & 24/35/38 of 40 & 0.170\\
 & Capacity & 45.575 & 7.622 & 2.397 & 40/40 & 20/29/38 of 40 & 0.170\\
Ising & LEF & 14.568 & 9.124 & 3.466 & 24/24 & 0/0/24 of 24 & 0.471\\
 & Capacity & 14.322 & 9.272 & 3.532 & 24/24 & 0/1/23 of 24 & 0.479\\
\bottomrule
\end{tabular}
\end{table}

Every one of the 128 returned chains strictly improves its degree-MH
reference. Paired geometric IACT improvements are 7.208 and 5.874 for
molecular LEF and capacity, and 1.569 and 1.613 for Ising. Median free-edge
fractions are 13.83\%, 13.79\%, 5.06\%, and 6.37\%, respectively. The Ising
LEF count of 0/24 describes certification of the actual returned chains;
the fully optimized versions of the same supports achieve 24/24. This
comparison isolates a remaining performance gap between the two realized
weighting procedures.

\section{Adaptive Conductance Release and Independent Confirmation}
\label{app:adaptive-confirmation}

\paragraph{A nested sequence of free variables.}
Fix the LEF four-edge support $\mathcal E_+$, its augmented degree-MH
conductances $c^0$, and node budgets $s_u=(1-\ell)\pi_u$, with $\ell=0.05$.
Start with all edges incident to an added-edge endpoint free, as in
Appendix~\ref{app:local-update}. At each subsequent stage, enlarge this set
to $\max\{|\mathcal F_0|,\lceil p|\mathcal E_+|\rceil\}$ edges for
$p=0.1,0.2,0.4$. Previously released variables remain free. Optimize the
Thomson current objective with all remaining conductances fixed exactly to
$c^0$. Thus the restriction concerns conductance degrees of freedom;
current conservation, the target, and observable information span the full
finite graph.

Let $z$ be the current fixed-chain variational potential, $J_0$ the
reference degree-MH Thomson energy, and $\lambda_u\geq0$ the budget dual
variable of the most recently accepted normalized optimization. The node
constraints in this program are divided by $s_u$, and its objective is
divided by $J_0$. Rank each still-fixed edge $e=(u,v)$ by
\begin{equation}
 \rho_e=\min(s_u,s_v)
 \left|\frac{(z_u-z_v)^2}{J_0}
       -\frac{\lambda_u}{s_u}-\frac{\lambda_v}{s_v}\right|.
 \label{eq:adaptive-release-appendix}
\end{equation}
The absolute residual prioritizes both flow recipients and potential budget
donors. Canonical edge order resolves ties. The matched control replaces
this ranking with a fixed SHA-256 ordering and frees the same number of
edges at every stage. Its four selected bridges are identical to LEF's;
only the order of releasing existing conductance constraints changes.

Both paths use the same convex program, rational budget repair, and
$10^{-8}$ mixture with the connected reference. A new candidate replaces
the incumbent only when its exact Thomson upper bound is lower. Complete
weighting uses the same primal program with every conductance free and the
same initial reference. ECOS uses absolute, relative, and feasibility
tolerances $10^{-9}$ and 300 iterations; the prescribed SCS fallback uses
$10^{-7}$ and 5,000 iterations. Recorded cumulative construction time
includes common preparation, the initial Poisson normalization, all
preceding stages, release-score computation, optimization, and repair.
In particular, the 20\% result includes the 10\% stage, and 40\% includes
both predecessors.

\paragraph{Development and fresh instances.}
Development used the original 12 Ising coefficient instances at paired
$\beta\in\{0.3,0.8\}$, plus three existing 1,024-state scaling instances.
The independent confirmation froze the complete rule before generating
eight new coefficient instances: four ferromagnetic and four mixed-sign
instances, seeds 2026102100--2026102107. Each uses the open $2\times4$
physical interaction grid, 256 spin configurations, and single-spin-flip
base support. Coupling magnitudes are integer draws from 10--30 divided
by 20; mixed-sign couplings receive independent random signs, and fields
are drawn from $-5$--5 divided by 20. A deterministic hash ordering supplies
512 candidate nonedges. Static LEF selects four once, and the same two
temperatures yield 16 cells. The coefficient instance is the outer unit.

Each cell repeats complete construction three times, alternating
adaptive/hash order and full-first/full-last order. The first repeat
fixes the kernels used for sampling and paired IACT summaries. Primary
certification requires all three repeats to satisfy
$U_{\tau,\mathrm{aug}}<L_{\tau,\mathrm{base}}$. Timing summaries first
take the median across the three repeats within a cell. The conservative
``certified and cheaper'' criterion additionally requires the largest of
the three adaptive construction times to be smaller than the smallest
of the three full-weighting times. The primary checkpoint is 20\%; 40\%
is the prespecified secondary checkpoint.

\begin{table}[ht]
\centering\small
\caption{Adaptive development and fresh confirmation on Ising. IACT and
construction time are cell medians; fresh times first aggregate the three
repeats within each cell. Strict certificates compare against all allowed
base-support weights. The final column applies the conservative three-repeat
cost criterion to the fresh cells.}
\label{tab:adaptive-cohorts}
\begin{tabular}{lrrr rrr r}
\toprule
 & \multicolumn{3}{c}{Development: 24 cells}
 & \multicolumn{3}{c}{Fresh confirmation: 16 cells} & Fresh certified\\
Method & Certified & IACT & Time (s) & Certified & IACT & Time (s) & and cheaper\\
\midrule
Degree-MH & 0/24 & 14.568 & 0.067 & 0/16 & 19.265 & 0.049 & 0/16\\
Endpoint update & 0/24 & 9.124 & 0.472 & 0/16 & 12.107 & 0.505 & 0/16\\
Adaptive 10\% & 0/24 & 8.344 & 0.910 & 0/16 & 7.809 & 0.956 & 0/16\\
Adaptive 20\% & 6/24 & 4.826 & 1.471 & 5/16 & 5.264 & 1.508 & 5/16\\
Adaptive 40\% & 12/24 & 3.849 & 2.361 & 8/16 & 4.603 & 2.371 & 0/16\\
Hash 20\% & 0/24 & 7.677 & 1.603 & 0/16 & 10.211 & 1.504 & 0/16\\
Hash 40\% & 0/24 & 5.953 & 2.511 & 1/16 & 8.194 & 2.291 & 0/16\\
Full weighting & 24/24 & 3.466 & 2.042 & 16/16 & 4.292 & 1.990 & ---\\
\bottomrule
\end{tabular}
\end{table}

\paragraph{Selection of free conductances matters.}
At both 20\% and 40\%, adaptive release beats matched hash release in all
24 development cells and all 16 fresh cells. Fresh preferences hold in
all three construction repeats. For each coefficient instance, average
the two log ratios $\log(L_{\tau,\mathrm{hash}}/
U_{\tau,\mathrm{adaptive}})$ before averaging across eight instances.
The resulting geometric ratios are 1.581 at 20\% and 1.495 at 40\%, with
95\% instance-bootstrap intervals [1.478,1.699] and [1.379,1.626]. All
eight outer ratios exceed one at both checkpoints. The fixed-seed
bootstrap uses 100,000 resamples of eight coefficient instances, retaining
paired temperatures. Its sampling unit remains the eight-instance cohort.

\begin{table}[ht]
\centering\small
\caption{Temperature-specific strict topology certificates. Fresh counts
require agreement across all three constructions.}
\label{tab:adaptive-temperature}
\begin{tabular}{llrrrrr}
\toprule
Cohort & $\beta$ & Adaptive 20\% & Adaptive 40\%
 & Hash 20\% & Hash 40\% & Full\\
\midrule
Development & 0.3 & 0/12 & 0/12 & 0/12 & 0/12 & 12/12\\
 & 0.8 & 6/12 & 12/12 & 0/12 & 0/12 & 12/12\\
Fresh & 0.3 & 0/8 & 0/8 & 0/8 & 0/8 & 8/8\\
 & 0.8 & 5/8 & 8/8 & 0/8 & 1/8 & 8/8\\
\bottomrule
\end{tabular}
\end{table}

The temperature pattern persists: partial release realizes strict topology
value at $\beta=0.8$, whereas full weighting achieves certification at both
temperatures. The five fresh 20\% certificates also satisfy the conservative
cost criterion. At 40\%, median IACT is higher and median cumulative
construction time is longer than for full weighting. This is a comparison
of cohort summaries; the individual cost curves retain each paired result.
For the three 1,024-state development instances, adaptive 20\%/40\% beats
hash release in all three cases, with geometric ratios 1.572/1.499. Their
IACT medians are 4.141/3.194, compared with 2.246 for full weighting;
the corresponding topology counts are 0/3, 0/3, and 2/3.

\paragraph{Exact objects and independent checking.}
The confirmation treats each serialized probability as an exact decimal
rational, renormalizes $\pi$ exactly, and derives $h,q,V$ from that common
target and the stored observable. Every conductance, current, and potential
witness is checked against these same quantities and $\ell=0.05$.
Fixed-chain bounds enclose each returned kernel's IACT; the base-family
lower bound ranges over its admissible weights. Independent standard-library
verification checks 480 kernels and 288 release steps, including budgets,
current divergence, frozen edges, release order, and cumulative times.
The largest relative fixed-chain IACT interval width is $6.247\times10^{-27}$.
All certificate decisions and reported fixed-chain IACTs agree across the
three repeats. This cohort's exact certification object is the stored-data
rational target; the separate real-Boltzmann interval audit covers the
earlier 64-cell comparison.

\section{Sampling Cost and ESS Amortization}
\label{app:amortization-confirmation}

\paragraph{Measured kernels and trajectories.}
All measurements use the same C++ sparse-alias sampler. A transition
includes random-number generation, state update, and observable accumulation.
For each first-repeat kernel, run ten timing repetitions of 64 interleaved
stationary-start chains, each with 32,768 measured steps after 256 warmup
steps. Method order is balanced across repetitions. The fresh cohort
therefore measures 160 kernels and 3,355,443,200 transitions. With the
known centered target mean, finite-horizon variance is estimated by
$\widehat\tau_T=T\,\operatorname{mean}(\bar h_T^2)/V$ using 640 chain
means per kernel. Its ratio to exact IACT has median 0.999 and range
0.861--1.132. The experiment uses two CPU workers with one BLAS thread
each on an Intel Xeon Platinum 8462Y+; all states and energies are cached.

\paragraph{Three cost ledgers.}
For a target effective sample count $M$, the stationary asymptotic model is
\begin{equation}
 T_a(M)=C_a+M\,\tau_a\,t_a,
 \label{eq:confirmation-cost}
\end{equation}
where $t_a$ is measured median time per transition and $\tau_a$ is the
midpoint of its verified fixed-kernel interval. The sampling intercept
$C_a$ comprises the median of three cumulative construction times plus
kernel conversion, alias-table preparation, and initialization. Common
candidate construction and every preceding optimization stage are charged
to each path. Degree-MH's offline Poisson evaluation is recorded as an
experimental measurement; the initial Poisson normalization required by
the implemented optimizer is part of its construction cost.

The \emph{sampling-setup ledger} charges construction, initialization, and
sampling of each compared kernel. Its initial external asset is the
original-support lower witness; each usable kernel is constructed and charged.
The \emph{verification ledger} additionally charges independent checks of
the original-support witness bundle and the returned kernel. The \emph{joint proof-and-kernel
ledger} starts before witness generation and charges the original-support
solve, which produces both the lower witness and a usable optimized base
kernel. An augmented alternative pays this solve in addition to its own
construction. The base-kernel alternative already includes the same solve
in its construction, so it is counted once in each alternative's total.

All three ledgers include the optimized base kernel in the full comparison.
We also report envelopes restricted to the nine weighting choices on the
same augmented support. These definitions distinguish externally supplied
proof information from jointly constructing the proof and baseline kernel.
If both products of an earlier joint solve are retained, their shared solve
cost is a sunk cost for both products; the ledgers here specify fresh kernel
construction or fresh joint construction.

\begin{table}[ht]
\centering\small
\caption{Fresh cells with a nonempty minimum-cost ESS interval for adaptive
release. Windows are computed from each cell's affine cost functions.
``With base kernel'' adds the feasible original-support optimized kernel
to the same-support comparison.}
\label{tab:confirmation-ledgers}
\begin{tabular}{lrrr}
\toprule
Cost ledger & 20\%: same support & 20\%: with base kernel & 40\%: same support\\
\midrule
Sampling setup & 14/16 & 14/16 & 1/16\\
Witness verification & 14/16 & 14/16 & 1/16\\
Joint proof-and-kernel & 14/16 & 1/16 & 1/16\\
\bottomrule
\end{tabular}
\end{table}

\begin{figure}[ht]
\centering
\includegraphics[width=\textwidth]{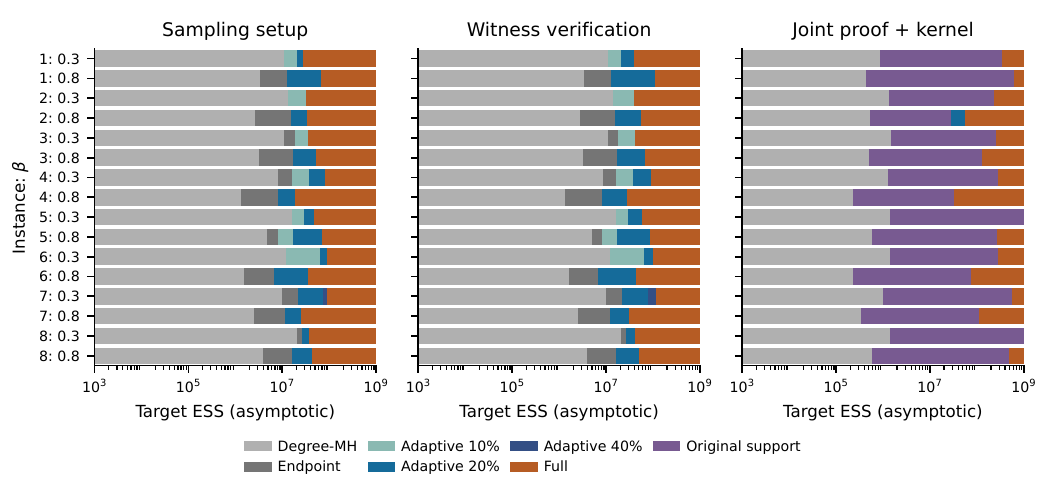}
\caption{Minimum-cost kernels for each fresh condition, including the optimized
original-support kernel. Each row is one coefficient instance and temperature;
colors mark the lower envelope of measured construction costs plus extrapolated
stationary sampling costs. The ledgers charge sampling setup, witness verification,
and joint proof-and-kernel construction. The optimized base kernel enters all
three comparisons; its solve is charged once per alternative in the joint ledger.
The plotted ESS range is $10^3$--$10^9$; counts in
Table~\ref{tab:confirmation-ledgers} use the full positive-ESS intervals.}
\label{fig:confirmation-windows}
\end{figure}

Under the sampling-setup ledger, adaptive 20\% has a minimum-cost interval in
14/16 fresh cells. Median pairwise crossover ESS is approximately
$1.09\times10^7$ against augmented degree-MH and $3.96\times10^7$ against
full weighting. Joint proof-and-kernel accounting changes the 20\% window
count to 1/16: the base solve serves both certification and sampling.
The crossover budgets are extrapolations
from measured stages and exact stationary asymptotic variance; the
observed trajectories have 32,768 steps each. Per-cell slopes, intercepts,
and crossing directions accompany the results. Figure~\ref{fig:confirmation-windows}
shows the budget intervals for each instance and temperature.

\paragraph{Earlier state-size cost study.}
A separate audit used the existing 64--1,024-state ring-Ising family at
$\beta=0.3$, with three coefficient seeds per size. It measured 75 saved
kernels and 1,572,864,000 transitions, comparing original degree-MH,
augmented degree-MH, endpoint update, original-support weighting, and full
augmented weighting. Sampling step medians range from 9.88 to 13.31 ns.
Historical construction timings and newly measured sampling costs are
combined by phase. A uniform ledger completion measured target preparation
and degree-flow construction; reconstructed conductances matched the
saved kernels exactly. Endpoint updates have minimum-cost intervals in
10/15 instances, while full weighting wins at sufficiently large budgets
in all 15. At 1,024 states, all three instances favor a short-budget base
kernel and a long-budget full-weighting kernel. The adaptive confirmation
above measures construction three times and keeps its own cost ledger.

\paragraph{Reproduction artifacts.}
All paths below are relative to the repository root. Each output directory
contains the frozen protocol, per-cell witnesses and timings, structured
summary, and independent verification report:
\begin{itemize}
\item \path{outputs/cost_amortization_v1/}: state-size sampling costs and
      \path{ledger_completion.json}.
\item \path{outputs/adaptive_weighting_v1/}: development release paths,
      exact checks, and per-instance cost curves.
\item \path{outputs/adaptive_confirmation_v1/}: fresh three-repeat records,
      \path{stage_metrics.csv}, \path{cost_curves.csv}, and
      \path{crossovers.csv}.
\end{itemize}
The corresponding independent checkers are
\path{experiments/verify_cost_amortization_v1.py},
\path{experiments/verify_adaptive_weighting_v1.py},
\path{experiments/verify_adaptive_confirmation_v1.py}, and
\path{experiments/verify_adaptive_confirmation_analysis_v1.py}.

\section{Release-Score Attribution on Development Instances}
\label{app:release-ablation}

This frozen follow-up reuses the original twelve Ising coefficient instances
at $\beta=0.3,0.8$. Every policy starts from the same LEF four-edge support
and endpoint solve, then follows nested 10\%, 20\%, and 40\% free sets with
identical solver settings, feasibility repair, and incumbent retention.
There are three complete timing repeats per cell, totaling 72 builds.
The primary comparison is 20\%; 40\% is supplementary. Each coefficient
instance is an outer unit, with both temperatures retained together.

\paragraph{Component scores.}
For each frozen edge, decompose Equation~\ref{eq:release-score} into
\[
 p_e=\min(s_u,s_v)(z_u-z_v)^2/J_0,\qquad
 d_e=\min(s_u,s_v)(\lambda_u/s_u+\lambda_v/s_v).
\]
Potential-only orders by $p_e$, dual-only by $|d_e|$, and the combined rule
by $|p_e-d_e|$, all descending with canonical tie breaking. Hash release
uses deterministic priorities. Potential-only is also the capacity-scaled
gradient-magnitude control. Every policy recomputes its scores from its own
accepted preceding stage. The comparison therefore evaluates complete nested
policies, including their different earlier releases.

\paragraph{Quality and cost.}
Table~\ref{tab:ablation-cost} uses the first repeat's rational fixed-chain
IACT upper bound. Times are medians within each cell over three repeats,
then medians across cells. Certification requires all repeats to separate
from the original-support family lower bound. The final column additionally
requires the slowest policy construction to be faster than the fastest full
construction in that cell. Cumulative cost includes the reference Poisson
initialization and every preceding solve, release, and repair.

\begin{table}[htbp]
\caption{Complete release-score development comparison. All stages retain
24 conditions; full and endpoint are shared references.}
\label{tab:ablation-cost}
\centering\small
\begin{tabular}{lrrrr}
\toprule
Stage & Median IACT & Cumulative seconds & Certified & Certified and cheaper than full \\
\midrule
endpoint & 9.124079 & 0.4729 & 0/24 & 0/24 \\
potential20 & 4.635916 & 1.4579 & 9/24 & 9/24 \\
dual20 & 8.307282 & 1.4814 & 1/24 & 1/24 \\
combined20 & 4.825849 & 1.4740 & 6/24 & 6/24 \\
hash20 & 7.677151 & 1.4938 & 0/24 & 0/24 \\
potential40 & 4.107867 & 2.2590 & 12/24 & 1/24 \\
dual40 & 6.898856 & 2.3084 & 2/24 & 0/24 \\
combined40 & 3.848505 & 2.3229 & 12/24 & 3/24 \\
hash40 & 5.952508 & 2.3313 & 0/24 & 0/24 \\
full & 3.465518 & 1.9613 & 24/24 & 0/24 \\
\bottomrule
\end{tabular}

\end{table}

Potential-only improves on the combined rule in 23/24 cells at 20\%, with
all twelve outer aggregates favorable. Nine conditions are certified and
cheaper than full, compared with six for the combined score. At 40\%, the
combined rule improves on potential-only in 22/24 cells and all twelve
outer aggregates. Both certify 12/24; their cohort median times exceed
full weighting's 1.9613 seconds, while full has the lower median IACT.
All partial-policy certificates occur at $\beta=0.8$; full certifies both
temperatures. These development results retain the original combined
rule as the object of the separate eight-instance confirmation.

\begin{table}[htbp]
\caption{Paired release comparisons. Ratios above one favor the combined
rule. Geometric aggregation first pairs temperatures within each instance;
strict wins use disjoint fixed-chain rational intervals.}
\label{tab:ablation-comparisons}
\centering\small
\begin{tabular}{rlrrrr}
\toprule
Release & Control & Control/combined & Cell median & Combined wins/losses & Leave-one-instance-out \\
\midrule
20\% & Potential & 0.965646 & 0.974851 & 1/23 & [0.962670, 0.969318] \\
20\% & Dual & 1.685595 & 1.702877 & 24/0 & [1.663885, 1.720480] \\
20\% & Hash & 1.598127 & 1.537450 & 24/0 & [1.581997, 1.619324] \\
40\% & Potential & 1.069301 & 1.047724 & 22/2 & [1.063218, 1.074180] \\
40\% & Dual & 1.801811 & 1.793052 & 24/0 & [1.751669, 1.848663] \\
40\% & Hash & 1.540321 & 1.503148 & 24/0 & [1.515505, 1.557621] \\
\bottomrule
\end{tabular}

\end{table}

In a stage-time replay using each cell's combined-20\% median construction
as budget, each control retains its latest completed checkpoint. Potential,
dual, and hash yield control/combined geometric IACT ratios 1.050809,
1.820858, and 1.725997; the combined rule wins 10/24, 24/24, and 24/24
respectively. This replay combines measured construction times and saved
fixed kernels. The independent checker verifies 1080 fixed-chain witnesses
and 864 releases, with maximum relative IACT interval width
$2.833\times10^{-26}$. Protocol, records, and aggregation checks are in
\path{outputs/adaptive_score_ablation_v1/}.

\subsection{Computational stages in the current implementation}
\label{app:implementation-complexity}

Let $n$ be the state count, $m$ the augmented edge count, $B_c$ the candidate
budget, and $F$ the free-conductance count. An Ising hypercube has
$m_0=n\log_2(n)/2$ edges. The following operation counts describe the current
implementation; rational arithmetic also depends on numerator and denominator
bit lengths.

\begin{table}[htbp]
\caption{Stage costs and retained information. Counts describe operations
and storage; conic solve time also depends on conditioning and factorization fill.}
\label{tab:implementation-complexity}
\centering\small
\begin{tabularx}{\textwidth}{lX}
\toprule
Stage & Current implementation \\
\midrule
Candidate discovery & Scans $\Theta(n^2)$ pairs; heap selection costs at most
$O(n^2\log B_c)$ with $O(B_c)$ extra candidate storage. \\
Local selection & $O(B_c)$ scores, $O(B_c\log B_c)$ sorting;
$O(n+m)$ degree preparation. \\
Poisson proposal & Dense factorization usually costs $O(n^3)$ floating-point
operations and $O(n^2)$ memory; repeated across stages and candidate kernels. \\
Restricted weighting & $m$ current variables, $F$ conductances, and $F$ cone
auxiliaries; full-graph conservation and target information remain active. \\
Release ordering & $O(m)$ scoring and $O(m\log m)$ sorting using the full
potential and node prices. \\
Witness checking & Budget and divergence sums require $O(n+m)$ rational
operations; bit lengths and current-repair tree construction add costs. \\
Kernel sampling & Sparse alias storage/build $O(n+m)$; $O(1)$ per transition. \\
\bottomrule
\end{tabularx}
\end{table}

Thus a 20\% free-conductance budget still retains all graph currents and
Poisson calculations. The measured 4096-state timeouts reflect this combined
implementation. The finite-budget study charges complete setup once before
its 96-chain workload; changing the number of chains changes each chain's
available runtime and defines a different workload.

\section{Prospective Factorial Confirmation of Selection and Release}
\label{app:ising-four-factor}

This new cohort tests whether the LEF/capacity edge-selection comparison and
the potential-only/combined release-score comparison persist independently
under matched restricted weighting. It is distinct from both the original
12-instance development study and the earlier eight-instance confirmation of
the frozen combined rule (Appendix~\ref{app:adaptive-confirmation}). Its design
was frozen before generating the new coefficient instances; all results below
are from the fixed 20\% primary and 40\% secondary analyses.

\paragraph{Targets and factorial design.}
Twelve new, independent $2\times4$ open-lattice Ising coefficient instances,
identified by seeds 2026110900--2026110911, comprise six ferromagnetic and six
mixed-sign systems. Each has 256 enumerated spin configurations, both
$\beta\in\{0.3,0.8\}$, the natural single-spin-flip original support,
$\ell=0.05$, and 512 common deterministic hash-ordered candidate nonedges.
The old coefficient protocols inspected for overlap included 14 records; none
matched these identifiers. The energy observable and the four-edge selection
budget are fixed. The two selectors are LEF
(Equation~\ref{eq:lef-preview}) and capacity-only
(Equation~\ref{eq:capacity}); each supplies its own four selected edges.
For each selector, combined and potential-only release start from exactly the
same endpoint-optimized kernel. At each stage, potential-only ranks the scaled
potential term $p_e$ and combined ranks $|p_e-d_e|$, as defined in
Appendix~\ref{app:release-ablation}. Each policy follows its own nested
10\%, 20\%, and 40\% free-conductance path before the full optimization.
The workflow retains full-graph currents and exact stationary targets,
regardless of the number of free conductances.

Each of the 24 target/temperature conditions has two selected supports, with
three complete builds per selector, totaling 144 builds. A build contains
nine kernels: degree-MH, endpoint, both release rules at three budgets each,
and fully free weighting. All three timing repeats return identical IACTs for
a given method and cell; the first supplies the fixed-chain quality result.
The primary unit for statistical inference is one of the twelve independent
coefficient instances, with its two temperatures nested. Coupling-type-stratified
instance bootstrap intervals operate on the instance-averaged paired log IACT
ratios. The four predeclared primary two-sided sign-flip comparisons enumerate
all $2^{12}=4096$ outer sign combinations and use Holm adjustment. Their
interpretation depends on instance-level independence and the symmetry
assumption for signed log contrasts. Strict cellwise ordering requires
disjoint rational fixed-chain IACT intervals, not just nominal floating-point
score differences.

\paragraph{Primary 20\% confirmation.}
In the table below, the ratio is numerator IACT divided by denominator IACT;
ratios greater than one favor the denominator. ``Wins'' counts strict pairwise
favorable cells from the 24 temperature-by-instance conditions, not
independent statistical replicates. The outer-count column uses twelve
coefficient instances. This table retains the original six-decimal effect
sizes from the completed analysis summary.

\begin{table}[htbp]
\centering\small
\caption{New $2\times2$ primary confirmation (20\% free conductances).
All four comparisons have resolved rational fixed-chain ordering in all
24 conditions. The $p$ values are corrected across these four tests.}
\label{tab:factorial-primary-full}
\begin{tabular}{lrrrrr}
\toprule
Comparison (numerator/denominator) & Geom. & 95\% CI & Median & Wins & Outer wins / $p$ \\
\midrule
LEF Combined / LEF Potential & 1.044824 & [1.031125,1.057476] & 1.047094 & 23/24 & 11/12 / .002930 \\
Cap Combined / Cap Potential & 1.038953 & [1.027005,1.052031] & 1.034746 & 23/24 & 12/12 / .001953 \\
Cap Combined / LEF Combined & 1.035469 & [1.015323,1.059798] & 1.025003 & 20/24 & 10/12 / .010742 \\
Cap Potential / LEF Potential & 1.041320 & [1.023294,1.060330] & 1.043512 & 21/24 & 10/12 / .008789 \\
\bottomrule
\end{tabular}
\end{table}

Leave-one-instance-out lower bounds for the four geometric ratios, in table
order, are 1.041405, 1.034684, 1.027971, and 1.036010. Thus the observed
positive aggregate directions are not driven by any single instance.
Nevertheless, the LEF/capacity differences are small and heterogeneous.
At $\beta=0.3$, LEF beats capacity in all 12 cells for either release rule,
with geometric capacity/LEF ratios 1.046769 (combined) and 1.058835
(potential-only). At $\beta=0.8$, the corresponding favorable counts are
8/12 and 9/12, and geometric ratios are approximately 1.024 with
stratified intervals crossing one. Ferromagnetic subgroup intervals support
LEF, while mixed-sign subgroup intervals cross one. These subgroup observations
do not establish a universal selector ranking. The factorial multiplicative
interaction ratio is 1.005651, 95\% interval [0.995789,1.015323]; there is
no demonstrated extra LEF-by-potential synergy beyond the conditional main
comparisons.

\paragraph{Secondary 40\% comparison.}
At 40\% free conductances, the release order reverses: combined/potential
geometric ratios are 0.935828 [0.924055,0.948895] on LEF support and
0.935973 [0.920358,0.952785] on capacity support, with combined better in
23/24 and 24/24 cells respectively. Both yield 12/12 favorable outer
instances. By 40\%, each policy has its own 10\% and 20\% released sets and
incumbents. These results compare complete policies and leave open the
effect of adding node prices to the last step from a shared incumbent.

\paragraph{Absolute quality, certificates, and measured setup.}
Fixed-chain IACTs are taken from the first complete build; setup and joint
costs are the median across three builds within each condition followed by
the median across the 24 conditions. The ``certified'' column requires all
three builds to strictly beat the optimized original-support lower witness.
The final column additionally requires all three setup times to be below their
corresponding full-weighting times (in these counted cases, even the slowest
partial build finishes before the fastest full build). Setup includes target
preparation, candidate enumeration and selection, reference Poisson solve,
endpoint construction, all preceding releases, optimization, and exact repair.
Joint cost additionally charges the original-support family certificate; it
is not a measured finite-budget MSE.

\begin{table}[htbp]
\centering\small
\caption{New factorial cohort: absolute IACT, complete setup cost and
joint proof-plus-kernel cost, all in seconds except IACT. ``Cert. + cheap''
uses the conservative three-repeat criterion described above.}
\label{tab:factorial-cost-full}
\begin{tabular}{lrrrrr}
\toprule
Selector / rule & IACT & Setup (s) & Joint (s) & Cert. & Cert. + cheap \\
\midrule
LEF / Combined 20\% & 4.6374 & 1.5576 & 4.7954 & 7/24 & 6/24 \\
LEF / Potential 20\% & 4.4372 & 1.4631 & 4.6586 & 7/24 & 6/24 \\
Capacity / Combined 20\% & 4.9487 & 1.5330 & 4.7552 & 7/24 & 6/24 \\
Capacity / Potential 20\% & 4.5384 & 1.5132 & 4.7368 & 7/24 & 7/24 \\
LEF / Combined 40\% & 3.6861 & 2.3711 & 5.6773 & 9/24 & 0/24 \\
LEF / Potential 40\% & 3.9225 & 2.2701 & 5.4193 & 9/24 & 2/24 \\
Capacity / Combined 40\% & 3.8921 & 2.3071 & 5.5512 & 9/24 & 2/24 \\
Capacity / Potential 40\% & 4.0772 & 2.2768 & 5.4421 & 7/24 & 0/24 \\
LEF / Full & 3.0528 & 1.9349 & 5.2273 & 24/24 & --- \\
Capacity / Full & 3.3640 & 1.8857 & 5.1980 & 20/24 & --- \\
\bottomrule
\end{tabular}
\end{table}

The LEF and capacity degree-MH medians are 14.5253 and 14.4935,
respectively; neither yields a strict support certificate (0/24). All
partial-weighting certificates occur at $\beta=0.8$. For capacity full,
four remaining conditions have overlapping certificate bounds, not verified
negative separations. At the 40\% stage, cohort-median cumulative setup time
exceeds that of full weighting; moreover, full has lower median IACT. Thus
the 40\% policy comparison describes the IACT ordering at that checkpoint.
The separate marginal medians leave per-cell cost dominance unresolved.
The eight-instance ESS cost windows, including the 1/16 joint-ledger result,
come from the earlier combined-versus-hash experiment.

\paragraph{Validation, compute environment, and evidence boundary.}
The completed analysis reports 24/24 original-support preparations, 144/144
successful constructions, 1,296 independently checked fixed-chain witnesses,
and 864 independently checked release orders. It reports no construction errors
or timeouts. Independent verification regenerated coefficients, energies,
candidate hash orderings and both selectors, and checked normalization,
centering, feasible node budgets, connectivity, current conservation,
fixed-edge flows, cumulative times, and monotone incumbents. The maximum
relative IACT interval width was $1.316\times10^{-26}$. Independent table
recomputation covered 1,296 stage rows, 192 paired rows, eight comparisons,
and 14 stage aggregates. Eleven existing algorithm/checker tests and two
statistics tests passed. The experiment used an Intel Xeon Platinum 8462Y+
CPU, two workers pinned to cores 48 and 49, single-threaded BLAS and no CUDA;
recorded compute, witness-checking and statistical-analysis times were
519.55, 142.80 and 7.54 seconds, respectively.

The strict rational certificates use exactly normalized rational versions of
stored Boltzmann probabilities. These comparisons isolate support selection,
release ordering and fixed-kernel quality on 256-state enumerated targets.
The manuscript package includes the result summary and a script that rebuilds
Table~\ref{tab:ising-factorial} from its reported aggregates
(Appendix~\ref{app:artifacts}). Recomputing the paired effects, intervals,
tests, and empirical certificates requires the original per-condition data
and witness records.

\section{Larger Finite Supports: Measured-Budget Mean Estimation}
\label{app:finite-budget}

\subsection{Frozen protocol and estimator}

This experiment tests whether the previously frozen partial-weighting rule
repays its construction cost in finite-time estimation. Five fresh coefficient
seeds, 2026103010--2026103014, generate ferromagnetic ring-Ising targets with
10, 11, and 12 spins (1024, 2048, and 4096 states) at $\beta=0.3$.
Couplings are integers sampled uniformly from 10--30 and divided by 20;
fields are integers from $-5$--5 divided by 20. Each size is paired within
seed. A separate 2048-state resource preflight fixes the final 360-second
deadline and is excluded from these estimates.

The base graph is the single-spin hypercube. Each augmented method uses the
same 512 nonedges chosen by SHA-256 priority over canonical state pairs and
selects four once, using original degrees. Streaming selection bounds the
stored candidate pool while inspecting all pairs. Coverage is 0.09872\%,
0.02456\%, and 0.00612\% of all nonedges at the three sizes. Targets are
exactly renormalized stored-probability rationals; both observables are centered
under that target. Median effective supports $1/\sum_x\pi_x^2$ are 391.390,
749.392, and 1260.170, calculated over all five targets at each size.
The energy mean is computed before optimization to construct $q$; the same
exact mean supplies the error reference. Thus the experiment isolates the
behavior and cost of constrained transition kernels with full target information.

Seven methods are evaluated: base degree-MH, base full weighting, LEF
degree-MH, LEF adaptive20, capacity adaptive20, LEF full weighting, and
enumerated-target independent sampling. In the archived labels used in this
appendix, adaptive20 denotes combined release at 20\%. Adaptive20 pays for the complete
endpoint $\to$10\%$\to$20\% path. The floor $\ell=0.05$, connected mixture
$10^{-8}$, release rule, solver settings, and rational repairs are inherited.
Full weighting uses the same objective with all conductances free and returns
a feasible kernel. Weighting optimizes energy; normalized magnetization is
evaluated on the same trajectories. Independent sampling pays for target
construction and an alias table and permits arbitrary state-to-state draws.

Each method is constructed and sampled once per instance. One kernel serves
96 chains: 32 all-plus, 32 all-minus, and 32 independently uniform-state starts.
For each start group, retain every observation after the first transition and
evaluate at batch budgets $b\in\{1,10,30,60,180,360\}$ seconds:
\[
 \widehat{\operatorname{NMSE}}_f(b)=
 \frac{1}{32V_f}\sum_{r=1}^{32}
   \bigl(\bar f_{r,b}-\pi(f)\bigr)^2.
\]
The 32 trajectories are nested error-estimation replicates; the five coefficient
seeds are the outer units. The budget includes one construction shared by the
96-chain workload, deducted in full before sampling. The estimator above
averages individual-chain squared errors; a pooled-chain mean would define
a different estimator. Each curve reports measured batch cost and finite-time
error, complementing the stationary ESS extrapolation in Equation~\ref{eq:cost-model}.

The primary ledger charges target/candidate construction, all preceding
weight solves, kernel materialization, input serialization, runtime parsing,
alias tables, numerical validation, initialization, and sampling. A second
ledger adds independent fixed-chain witness verification and, for degree-MH,
offline Poisson/IACT evaluation. Both use prefixes of the same run. Generating
the original-support optimal lower witness belongs to the separate joint
proof-and-kernel experiment in Appendix~\ref{app:amortization-confirmation}.
Process launch/final delivery and research-artifact generation are recorded
separately. The clock is checked every 16 sweeps; across 705 sampled snapshots,
the median deadline overshoot is $1.54\times10^{-5}$ seconds and the maximum
is $5.97\times10^{-5}$ seconds.

\subsection{Availability and construction resources}

Each isolated method receives one CPU core, one BLAS thread, a 900-second
build limit, and a 16-GiB address-space limit. At most two tasks run
concurrently, with balanced method order. The 105 scheduled tasks yield
88 completed builds and 17 timeouts. Of the completed builds, 85 supply
samples at some deadline. The other three are 4096-state adaptive builds
with setup times 845.654, 652.380 (LEF), and 440.745 seconds (capacity).
At 4096 states, both full-weighting methods time out in 5/5 instances;
LEF adaptive20 times out in 3/5 and capacity adaptive20 in 4/5. Every
optimized method is unavailable at every sampling deadline.

\begin{figure}[htbp]
\centering
\includegraphics[width=\textwidth]{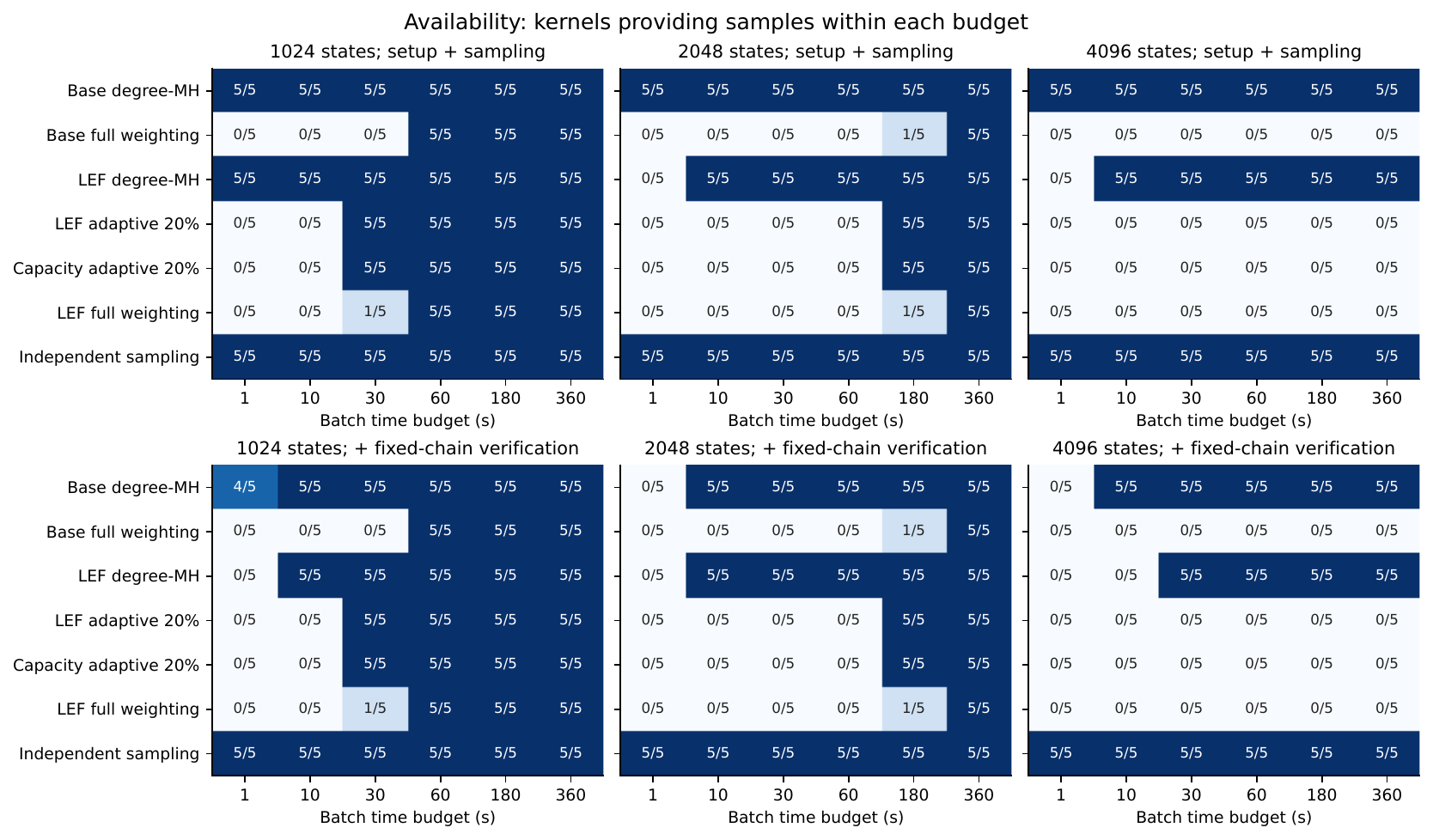}
\caption{\textbf{Budget availability.} Every entry retains all five coefficient
instances. Rows distinguish the construction-plus-sampling ledger from the
ledger adding fixed-chain verification. At 4096 states, degree-MH and independent
sampling remain available, while every weight-optimization method exceeds
the sampling deadlines.}
\label{fig:budget-availability}
\end{figure}

Figure~\ref{fig:budget-availability} retains these outcomes in the denominator.
For example, at 2048 states and 180 seconds, both adaptive methods are
available in 5/5 instances, whereas each full-weighting method is available
in 1/5. An unavailable run has an undefined MSE. Partial-subset estimates
are marked separately from full five-instance comparisons.

\begin{table}[htbp]
\caption{Construction and resource measurements. Setup is the completed-task
median in seconds; the two counts are completed builds and builds providing
samples within any deadline. RSS is the median of the 0.5-second monitored
process-tree peaks across all five attempts, including timeouts. Short-lived
peaks may fall between polls; per-process high-water marks are also stored.
Timeout memory reflects the stages actually reached.}
\label{tab:budget-resources}
\centering\small
\begin{tabular}{llrrr}
\toprule
States & Method & Completed; budget-available & Setup seconds & Peak RSS (MiB) \\
\midrule
1024 & Base degree-MH & 5/5; 5/5 & 0.111 & 115.3 \\
1024 & Base full & 5/5; 5/5 & 43.551 & 150.7 \\
1024 & LEF degree-MH & 5/5; 5/5 & 0.459 & 102.1 \\
1024 & LEF adaptive20 & 5/5; 5/5 & 16.069 & 149.1 \\
1024 & Capacity adaptive20 & 5/5; 5/5 & 19.009 & 145.1 \\
1024 & LEF full & 5/5; 5/5 & 32.127 & 150.7 \\
1024 & Independent & 5/5; 5/5 & 0.028 & 79.0 \\
2048 & Base degree-MH & 5/5; 5/5 & 0.234 & 212.5 \\
2048 & Base full & 5/5; 5/5 & 272.788 & 311.3 \\
2048 & LEF degree-MH & 5/5; 5/5 & 1.647 & 212.6 \\
2048 & LEF adaptive20 & 5/5; 5/5 & 123.579 & 284.5 \\
2048 & Capacity adaptive20 & 5/5; 5/5 & 134.649 & 284.9 \\
2048 & LEF full & 5/5; 5/5 & 204.887 & 311.5 \\
2048 & Independent & 5/5; 5/5 & 0.057 & 83.0 \\
4096 & Base degree-MH & 5/5; 5/5 & 0.537 & 559.7 \\
4096 & Base full & 0/5; 0/5 & -- & 559.6 \\
4096 & LEF degree-MH & 5/5; 5/5 & 6.180 & 558.8 \\
4096 & LEF adaptive20 & 2/5; 0/5 & 749.017 & 678.9 \\
4096 & Capacity adaptive20 & 1/5; 0/5 & 440.745 & 656.0 \\
4096 & LEF full & 0/5; 0/5 & -- & 559.8 \\
4096 & Independent & 5/5; 5/5 & 0.114 & 88.0 \\
\bottomrule
\end{tabular}

\end{table}

Completed 1024/2048 builds spend most time on weight solves. Their timing
decomposition and per-process memory are retained alongside Table~\ref{tab:budget-resources}.
The dense Poisson initialization and exhaustive candidate-pair scan remain
part of the measured implementation. The 4096-state failures meet the frozen
protocol's stopping condition for further scale expansion.

\subsection{Finite-budget errors, observable dependence, and uncertainty}

\begin{figure}[htbp]
\centering
\includegraphics[width=.94\textwidth]{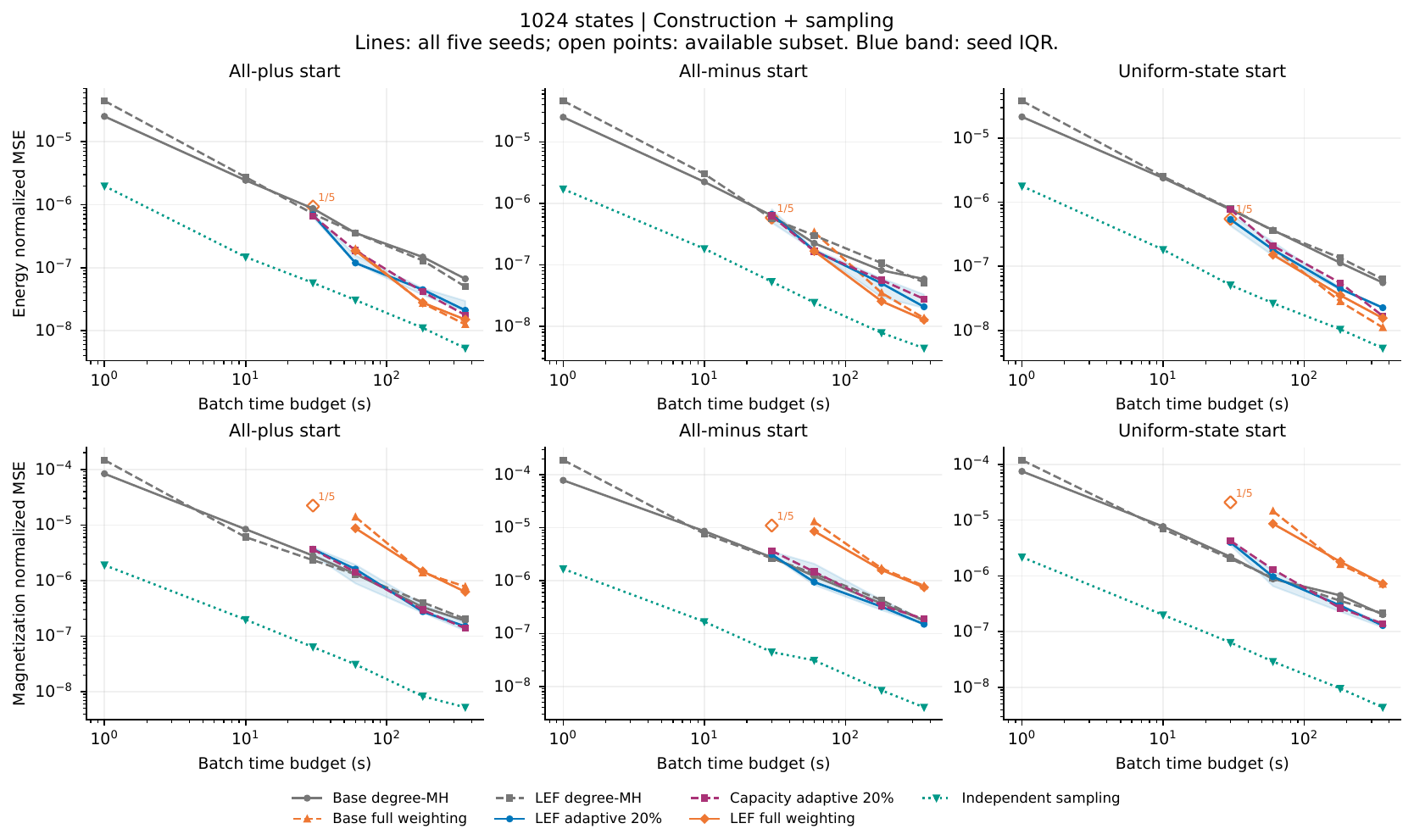}\par
\includegraphics[width=.94\textwidth]{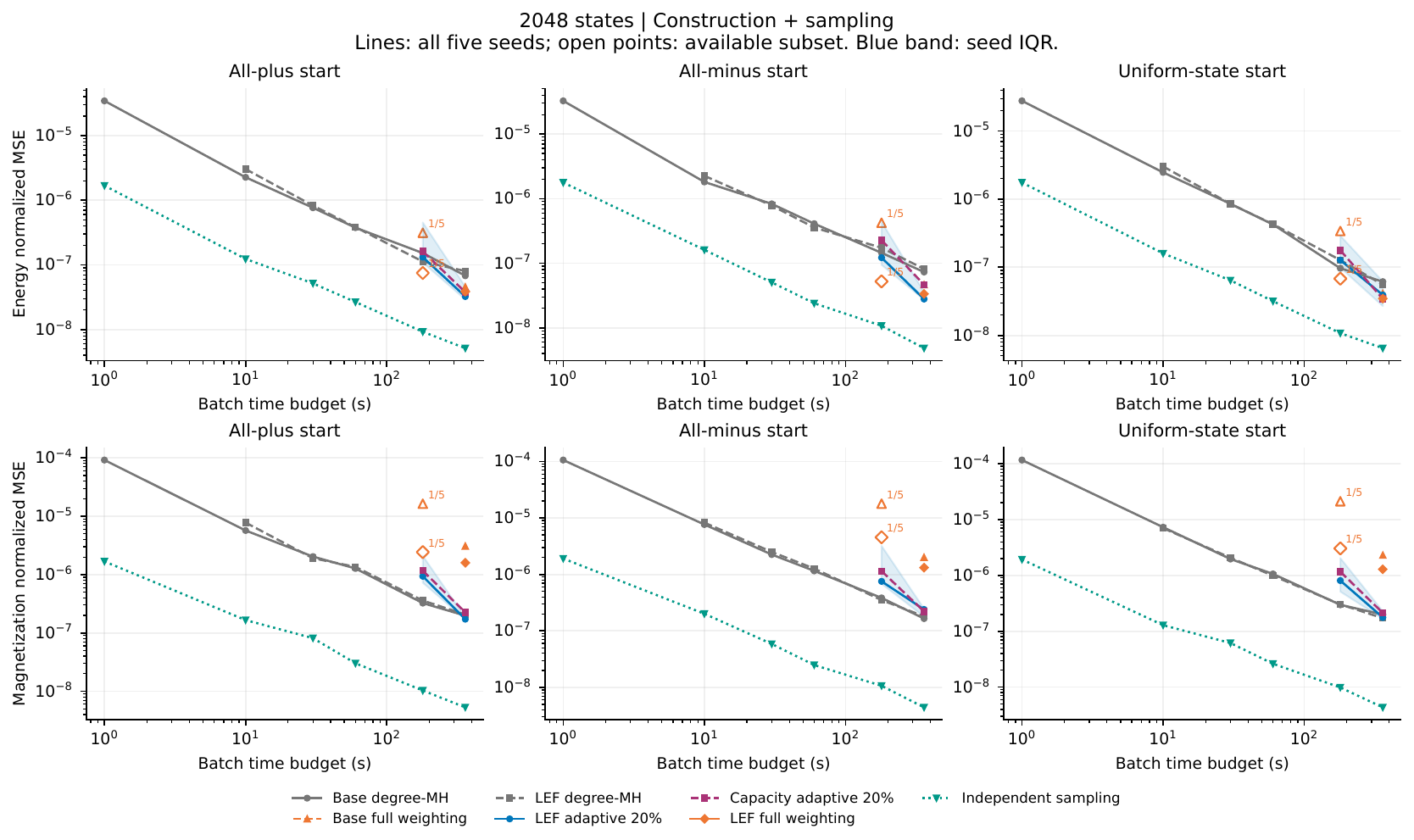}
\caption{\textbf{Finite-budget errors: 1024 states (top), 2048 (bottom).}
All starts and both observables use the same measured trajectories.
At 360 seconds, the 1024-state full-weighting energy advantage reverses for
magnetization. At 2048 states and 180 seconds, full-weighting hollow points
summarize one available seed; all five are available at 360 seconds.
The all-minus magnetization comparison favors base degree-MH over adaptive20
in all five 2048-state instances.}
\label{fig:budget-curves}
\end{figure}

Figure~\ref{fig:budget-curves} shows all six deadlines,
three starts, and two observables. Each full curve uses the median over five
outer seeds. The blue band is the interquartile range across seeds. Hollow
points indicate available subsets, with their counts displayed.

Tables~\ref{tab:budget-energy}--\ref{tab:budget-magnetization} give all starts
at 360 seconds. Ratios divide a comparator's MSE by LEF adaptive20 MSE on
the same seed, then geometrically aggregate the five ratios. Parentheses
give the number of seeds favoring adaptive20. The capacity comparison changes
with start and size. Full energy weighting also changes the ranking for
magnetization. These are finite-budget performance comparisons of the complete
selection-and-weighting procedures.

\begin{table}[htbp]
\caption{Energy MSE ratios at 360 seconds, including construction.
Every comparison contains five paired seeds. Larger than one favors
LEF adaptive20; parentheses report favorable seeds.}
\label{tab:budget-energy}
\centering\footnotesize\setlength{\tabcolsep}{3pt}
\begin{tabular}{llrrrrrr}
\toprule
States & Start & Base degree & LEF degree & Base full & LEF full & Capacity20 & Independent \\
\midrule
1024 & all-plus & 2.437 (5/5) & 2.302 (5/5) & 0.599 (1/5) & 0.525 (0/5) & 0.737 (1/5) & 0.213 (0/5) \\
1024 & all-minus & 2.224 (5/5) & 2.393 (5/5) & 0.586 (0/5) & 0.518 (0/5) & 1.171 (3/5) & 0.208 (0/5) \\
1024 & uniform-state & 2.740 (5/5) & 2.789 (5/5) & 0.598 (1/5) & 0.695 (0/5) & 0.858 (2/5) & 0.206 (0/5) \\
2048 & all-plus & 2.182 (5/5) & 2.016 (5/5) & 1.278 (4/5) & 0.993 (3/5) & 1.123 (4/5) & 0.135 (0/5) \\
2048 & all-minus & 2.270 (5/5) & 2.572 (5/5) & 1.373 (3/5) & 1.019 (3/5) & 1.311 (3/5) & 0.140 (0/5) \\
2048 & uniform-state & 1.480 (4/5) & 1.469 (3/5) & 1.048 (3/5) & 0.779 (2/5) & 0.851 (2/5) & 0.136 (0/5) \\
\bottomrule
\end{tabular}

\end{table}

\begin{table}[htbp]
\caption{Magnetization MSE ratios under the same budget and trajectories.
Notation matches Table~\ref{tab:budget-energy}.}
\label{tab:budget-magnetization}
\centering\footnotesize\setlength{\tabcolsep}{3pt}
\begin{tabular}{llrrrrrr}
\toprule
States & Start & Base degree & LEF degree & Base full & LEF full & Capacity20 & Independent \\
\midrule
1024 & all-plus & 1.175 (3/5) & 1.141 (3/5) & 4.837 (5/5) & 3.919 (5/5) & 0.822 (1/5) & 0.031 (0/5) \\
1024 & all-minus & 1.066 (3/5) & 1.037 (2/5) & 4.741 (5/5) & 4.189 (5/5) & 1.040 (3/5) & 0.024 (0/5) \\
1024 & uniform-state & 1.335 (5/5) & 1.487 (5/5) & 5.694 (5/5) & 4.804 (5/5) & 1.070 (3/5) & 0.034 (0/5) \\
2048 & all-plus & 1.126 (3/5) & 1.132 (3/5) & 14.663 (5/5) & 8.088 (5/5) & 1.134 (3/5) & 0.024 (0/5) \\
2048 & all-minus & 0.730 (0/5) & 0.755 (1/5) & 10.116 (5/5) & 7.165 (5/5) & 0.936 (2/5) & 0.021 (0/5) \\
2048 & uniform-state & 0.970 (2/5) & 0.979 (2/5) & 13.218 (5/5) & 7.723 (5/5) & 1.071 (3/5) & 0.022 (0/5) \\
\bottomrule
\end{tabular}

\end{table}

\begin{table}[htbp]
\caption{Absolute normalized energy MSE at 360 seconds, uniform-state start:
median over five seeds, construction-plus-sampling ledger. A dash means
zero budget-available outputs. These medians and the paired geometric ratios
summarize different quantities.}
\label{tab:budget-absolute}
\centering\small
\begin{tabular}{lrrr}
\toprule
Method & 1024 & 2048 & 4096 \\
\midrule
Base degree-MH & 5.570e-08 & 6.149e-08 & 7.698e-08 \\
Base full & 1.113e-08 & 4.121e-08 & -- \\
LEF degree-MH & 6.278e-08 & 5.544e-08 & 8.627e-08 \\
LEF adaptive20 & 2.264e-08 & 3.919e-08 & -- \\
Capacity adaptive20 & 1.678e-08 & 3.447e-08 & -- \\
LEF full & 1.556e-08 & 3.477e-08 & -- \\
Independent & 5.287e-09 & 6.485e-09 & 4.861e-09 \\
\bottomrule
\end{tabular}

\end{table}

\begin{table}[!htbp]
\caption{Small-sample sensitivity for 360-second energy ratios with uniform-state
starts. Pointwise percentile intervals exhaust all $5^5=3125$ ordered outer-seed
bootstrap resamples; the last column gives the range after excluding each seed
in turn. Shared starts, budgets, observables, and ledgers remain nested.}
\label{tab:budget-uncertainty}
\centering\small
\begin{tabular}{llrrrr}
\toprule
States & Comparator & Ratio & Favorable & Outer 95\% interval & Leave-one-seed-out \\
\midrule
1024 & Base degree-MH & 2.740 & 5/5 & [2.460, 3.090] & [2.611, 2.838] \\
1024 & Base full & 0.598 & 1/5 & [0.475, 0.890] & [0.492, 0.636] \\
1024 & LEF full & 0.695 & 0/5 & [0.598, 0.827] & [0.643, 0.738] \\
1024 & Capacity adaptive20 & 0.858 & 2/5 & [0.597, 1.233] & [0.739, 0.966] \\
1024 & Independent & 0.206 & 0/5 & [0.154, 0.258] & [0.191, 0.237] \\
2048 & Base degree-MH & 1.480 & 4/5 & [1.033, 2.119] & [1.328, 1.712] \\
2048 & Base full & 1.048 & 3/5 & [0.614, 1.730] & [0.893, 1.315] \\
2048 & LEF full & 0.779 & 2/5 & [0.451, 1.224] & [0.661, 1.017] \\
2048 & Capacity adaptive20 & 0.851 & 2/5 & [0.592, 1.204] & [0.736, 0.989] \\
2048 & Independent & 0.136 & 0/5 & [0.091, 0.211] & [0.113, 0.155] \\
\bottomrule
\end{tabular}

\end{table}

With fixed-chain verification charged, the uniform-start base-degree/adaptive20
energy ratios at 360 seconds are 2.658 and 1.444 for 1024 and 2048 states,
respectively; corresponding outer intervals are $[2.374,2.983]$ and
$[1.037,2.010]$. Complete second-ledger curves, per-seed effects, Monte Carlo
standard errors, and bias/scatter decompositions are stored in the analysis
artifact. The decomposition uses the identity
$\widehat{\mathrm{NMSE}}=\bar e^2/V+(32V)^{-1}\sum_r(e_r-\bar e)^2$,
where $e_r$ is a chain's mean error.

\paragraph{Verification and evidence role.}
The independent standard-library checker verifies 73 rational fixed-MCMC-chain
witnesses and 15 independent-sampling records. A separate aggregation checker
recomputes all 4,230 sampled MSE records and 2,160 paired summaries; the full
7,560-row ledger also includes 2,106 setup-over-budget and 1,224 timeout rows.
Five coefficient seeds remain the outer sample size. All frozen outcomes
enter the descriptive estimates and availability counts. These measurements
extend the evidence to finite-time estimation on larger enumerated targets;
the original support-separation certificates retain their separate comparison
against all admissible base weights.

The frozen inputs are in \path{outputs/large_state_budget_v1/}; analysis and
verification receipts are in \path{outputs/large_state_budget_analysis_v1/}.
\path{per_seed_mse.csv} records every planned condition;
\path{paired_comparisons.csv} and \path{paired_seed_ratios.csv} retain
paired effects and availability; \path{cost_by_task.csv} records all attempts.
Figure and table generators check source hashes and the independent analysis
receipt before producing manuscript assets.

\section{State-Size, Construction-Cost, and IACT Tradeoffs}
\label{app:scaling}

The scaling family contains ferromagnetic ring Ising energies with 6--10
spins, $\beta=0.3$, and three fixed coefficient seeds per size
(2026101700--2026101702). Couplings are integer draws from 10--30 divided by
20, and fields from $-5$--5 divided by 20. Base support is the single-spin
hypercube. Canonically hash-ordering all nonedges supplies 512 candidates;
unchanged LEF selects four once. Candidate construction includes the cost of
enumerating all state pairs. Candidate coverage decreases from 28.07\% at
64 states to 0.0987\% at 1,024 states. Effective support ranges from
33.9--43.5 at the smallest size to 407.1--469.9 at the largest.

\begin{table}[ht]
\centering\small
\caption{Scaling medians across three instances per size. IACT columns are
medians of feasible upper bounds. Local/full timing is measured in the same
run; peak memory covers the complete process, including full optimization
and verification.}
\label{tab:updated-scaling}
\begin{tabular}{rrrrrrrrr}
\toprule
States & Candidate (s) & Local (s) & Full (s) & Peak MiB
 & $\tau_D$ & $\tau_L$ & $\tau_F$ & Full certified\\
\midrule
64 & 0.0024 & 0.082 & 0.501 & 130.4 & 4.180 & 3.338 & 1.018 & 3/3\\
128 & 0.0323 & 0.205 & 1.229 & 159.3 & 5.566 & 4.264 & 1.295 & 3/3\\
256 & 0.0572 & 0.460 & 3.042 & 224.7 & 7.236 & 5.798 & 1.624 & 3/3\\
512 & 0.2036 & 1.245 & 8.983 & 391.7 & 8.143 & 7.044 & 1.775 & 3/3\\
1024 & 0.9604 & 4.246 & 25.701 & 823.9 & 9.823 & 8.786 & 2.246 & 2/3\\
\bottomrule
\end{tabular}
\end{table}

LEF scoring takes 0.00078--0.00196 seconds across the five size medians;
final exact verification takes 0.040--7.171 seconds. Median paired full/local
construction-time ratios range from 5.98 to 7.02, while local/full geometric
IACT ratios range from 2.94 to 3.89. These are construction-cost and stationary
IACT measurements for cached finite kernels. All 15 local outputs strictly
improve degree-MH; full weighting certifies topology value in 14/15 cases.
For the remaining 1,024-state instance, base-optimal bounds
$[2.160224,2.297374]$ overlap augmented bounds $[2.236747,2.236757]$.
Local outputs yield 0/15 topology certificates. The run uses two CPU workers
with one BLAS thread each, and a 600-second per-instance limit.

\section{Interval Certificates for the Real Boltzmann Target}
\label{app:target-interval}

This audit covers the original 40 molecular and 24 Ising cells, each with
LEF and capacity supports. Regard stored decimal energies and $\beta$ as
exact inputs and define
\begin{equation}
 \pi'_u=\frac{\exp[-\beta(E_u-E_{\min})]}
 {\sum_v\exp[-\beta(E_v-E_{\min})]} .
\end{equation}
Rational range reduction to $|x|\le1/8$, a degree-64 Taylor sum with a
geometric upper tail, and outward rounding to 80 decimal places enclose each
exponential and normalized probability. Let $[p_u^-,p_u^+]$ enclose $\pi'_u$.
All certificate operations use integer and rational interval arithmetic.

Let $h$ be the reference-centered observable. For each normalized target in
the enclosure, recenter it by $\mu'=\sum_u\pi'_u h_u$, giving
$q'_u=\pi'_u(h_u-\mu')$ and variance
$V'=\sum_u\pi'_u h_u^2-(\mu')^2$. Original base-support witnesses $z,y$ remain
feasible because their edge inequalities are independent of target masses.
Interval evaluation therefore supplies
\begin{equation}
 L_J\le 2(q')^\top z-(1-\ell)(\pi')^\top y
 \le \inf_{c\text{ on base}}(q')^\top A(c)^\dagger q'.
\end{equation}
For the augmented upper bound, scale the stored feasible conductances by
$\alpha=\min\{1,\min_u p_u^-/\pi_u\}$, so that
$c'_e=\alpha c_e$ respects every new node budget. Repair the stored conserved
current from source $q$ to $q'$ by subtree residuals on a positive-conductance
spanning tree. Exact normalization ensures closure at the root. Interval
evaluation of $\sum_{e:c'_e>0}(j'_e)^2/c'_e$ gives $U_J$; zero-conductance
edges carry zero current. With $0<V_-\le V'\le V_+$ and $L_J>0$,
\begin{equation}
 L_\tau=2L_J/V_+-1,\qquad U_\tau=2U_J/V_--1.
 \label{eq:interval-target-iact}
\end{equation}
Strict $U_\tau<L_\tau$ certifies a feasible target-adjusted chain on the
same augmented support against every allowed base weighting. This establishes
the comparison for the real exponential law defined by the stored energies
and temperature.

\begin{table}[ht]
\centering\small
\caption{Real-target interval certification. Relative margin is
$(L_\tau-U_\tau)/L_\tau$; minima range over certified cells.}
\label{tab:updated-target-interval}
\begin{tabular}{llrrrr}
\toprule
Cohort & Selector & Rational target & Real target
 & Minimum IACT margin & Minimum relative margin\\
\midrule
Molecular & LEF & 38/40 & 38/40 & 0.00181536 & 0.1990\%\\
 & Capacity & 38/40 & 38/40 & 0.00111776 & 0.1225\%\\
Ising & LEF & 24/24 & 24/24 & 0.10490894 & 1.1717\%\\
 & Capacity & 23/24 & 23/24 & 0.00009853 & 0.005264\%\\
\bottomrule
\end{tabular}
\end{table}

All 123 originally strict comparisons remain strict. Maximum relative target
deviation bounds are $1.623\times10^{-14}$ for molecular cells and
$3.433\times10^{-15}$ for Ising cells. Independent standard-library checking
verifies all 207 records: 128 local outputs, 15 scaling instances, and 64
target-interval records containing both selectors. Twenty tests cover analytic
small graphs, fixed flows, budget repair, interval exponential enclosures, and
certificate inequalities. An initial SCS compilation incompatibility was
resolved by the equivalent conic-epigraph setting \texttt{use\_quad\_obj=False};
all local and scaling results use that implementation, with the original
records preserved. Machine-readable summaries and verification entry points
are \path{outputs/optimized_selector_controls_v1/summary.json},
\path{outputs/reviewer_gap_audits_v2/summary.json},
\path{experiments/verify_optimized_selector_controls_v1.py}, and
\path{experiments/verify_reviewer_gap_audits.py}.

\section{Molecular Energy Checkpoints}
\label{app:energy-model}

The held-out and new-component evaluations use QM9 graph-transformer energy
checkpoints trained for 60,000 updates. Their saved configuration has 12 layers,
eight attention heads, node/edge/global hidden widths 256/64/128, and SiLU
activations. Inputs contain categorical atom and bond features, molecular
features, and relative random-walk features with 20 steps. Time embedding is
fixed to zero. If $X_i^{(L)},D_{ij}^{(L)},y^{(L)}$ are the final hidden features,
the scalar potential is
\begin{equation}
 E_\theta(x)=\operatorname{MLP}_{448\to256\to1}
 \left(\left[\sum_iX_i^{(L)},\ \sum_{i\ne j}D_{ij}^{(L)},\ y^{(L)}\right]\right),
\end{equation}
with masks excluding padding and diagonal edges. The energy head acts directly
on the hidden representation of the same graph transformer.

Training couples categorical noise and data graphs, optionally aligns the data
graph through QAP, and draws a categorical interpolant $x_t$ at uniform $t$.
For node and upper-triangle edge one-hot coordinates $a\in\{X,D\}$, the target
is $a_1-a_0$. The loss averages the unit-threshold Huber discrepancy between
that displacement and $-\operatorname{clip}(\nabla_aE_\theta(x_t),-10,10)$,
with unit weights on node and edge losses. The saved runs use AdamW, learning
rate $2.5\times10^{-5}$, batch size 256, and parameter-gradient clipping 10.
The contrastive-loss weight is zero. Thus the flow-matching training objective
learns the scalar potential through its input gradients; the sampling study
then evaluates the configured Boltzmann law of that frozen potential. Seeds
21--25 each contribute QAP-off/on checkpoints. All methods within a cell use
the same cached scalar energy vector.

\section{Held-Out Component and Kernel Specification}
\label{app:component-spec}

The held-out checkpoint and prospective new-component experiments use the
following deterministic construction. The four consumed development components
in the development analyses come from frozen
physical-CIM candidate banks. The held-out
construction is stated here independently of artifact paths.
\begin{enumerate}
\item Scan the seed-0 QM9 test split in ascending source-index order after removing
hydrogens. A root is eligible when it has nine heavy atoms, is connected, obeys
the atom-valence upper bounds, and admits at least 64 structurally valid edits.
\item Enumerate every canonical one- or two-coordinate change among upper-triangle
edge-type coordinates. Retain candidates that remain connected and satisfy the
same valence bounds, encode each edit as a 180-bit replacement vector, and
canonically deduplicate.
\item Select 64 nonroot states by farthest-first traversal under edit-bit Hamming
distance. The first lexicographic candidate initializes the traversal; all ties
are lexicographic, making this a purely structural selection step.
\item Add the zero-edit root, connect it to all 64 states, and for every nonroot
state add edges to its four closest nonroot peers in Hamming distance, again with
lexicographic ties. Symmetrizing the union gives the base proposal graph.
\item Exactly rescore all 65 states and restrict the checkpoint Boltzmann law to
this universe. The frozen inverse temperatures are 14.3322378 for QAP-off and
11.9435315 for QAP-on checkpoints. The base kernel is uniform-neighbor MH with
laziness $\ell=0.05$:
\begin{equation}
 P(u,v)=(1-\ell)\min\left\{\frac{1}{d(u)},
 \frac{\pi(v)}{\pi(u)d(v)}\right\},\qquad (u,v)\in\mathcal E,
\end{equation}
with the remaining row mass assigned to $P(u,u)$.
\item Score every base-graph nonedge once using Equation~\ref{eq:lef-preview} and
the original degrees, select four with canonical tie breaking, and rebuild the
degree-MH kernel on the augmented graph. Selection uses one-shot scoring.
\end{enumerate}

The original family uses QM9 source indices 3969, 21540, 21546, and 21561. The
prospective family uses the eight indices listed in
Appendix~\ref{app:new-components}. The latter changes component identity while
holding the construction and kernel fixed.

Three edge interventions are intentionally distinct:
\begin{center}
\scriptsize
\begin{tabularx}{\linewidth}{p{0.23\linewidth}p{0.30\linewidth}X}
\toprule
Analysis & Intervention & What changes besides the new edge \\
\midrule
LEF evaluation & Rebuild degree-MH after four one-shot selections & All transition
probabilities incident to augmented endpoints may change \\
Four-edge oracle & Re-solve SLEM-minimizing FMMC on augmented support & Potentially
all conductances on the connected graph \\
Proposition diagnostic & Move stationary mass from endpoint self-loops &
Pre-existing off-diagonal conductances remain fixed \\
Held-out crossover & Apply both degree-MH and fixed-old-flow interventions &
Selected portfolios stay fixed; only the kernel intervention changes \\
\bottomrule
\end{tabularx}
\end{center}
Each single self-loop-to-edge insertion is a rank-one Laplacian update; a four-edge portfolio is a sum of four such updates.

\section{Prospective New-Component Transfer}
\label{app:new-components}

This experiment was frozen before any new-component checkpoint energy was
evaluated. Starting from the ascending QM9 test split, a deterministic
structure-only builder excludes the four earlier component roots and accepts the
first eight eligible roots. Eligibility, local edit enumeration, canonical
deduplication, the 65-state component size, LEF formula, four-edge budget, target
construction, and exact finite-state evaluation are unchanged. The resulting
source indices are 21583, 21593, 21597, 21603, 21611, 21634, 21645, and 21649;
their eligible structure-only candidate pools contain 267, 374, 354, 172, 440,
1020, 505, and 293 states, respectively. A no-science preflight verified
deterministic rebuilding, disjointness from the earlier family, all checkpoint
contracts, and all $8\times65$ feature batches without instantiating or
forwarding a model.

Each component is crossed with the same ten checkpoints (five training seeds and
two QAP modes), producing 80 nested exact finite-state cells and 5,200 unique
checkpoint-energy evaluations. Component is the outer statistical unit. The
prespecified criteria require at least 7/8 favorable component aggregates, a
100,000-replicate deterministic component-bootstrap lower ratio above one, at
least 60/80 favorable cells, gap noninferiority in at least 7/8 components, and
all finite-state certificates. Table~\ref{tab:new-components} reports the
component-level result; the minimum column exposes the single unfavorable cell
rather than hiding it in an aggregate.

\begin{table}[ht]
\caption{Prospective transfer to eight new energy-blind components. Ratios are
base energy IACT divided by LEF energy IACT over the ten nested checkpoints.}
\label{tab:new-components}
\centering
\scriptsize
\setlength{\tabcolsep}{4.0pt}
\begin{tabular}{rrrrr}
\toprule
component & QM9 index & candidates & minimum cell ratio & component geometric ratio \\
\midrule
1 & 21583 & 267  & 1.435 & 39.205 \\
2 & 21593 & 374  & 2.896 & 309.173 \\
3 & 21597 & 354  & 22.239 & $3.410\times10^{5}$ \\
4 & 21603 & 172  & 1.787 & $2.169\times10^{5}$ \\
5 & 21611 & 440  & $7.340\times10^{3}$ & $1.732\times10^{6}$ \\
6 & 21634 & 1020 & 2.971 & 288.869 \\
7 & 21645 & 505  & 0.998 & 92.917 \\
8 & 21649 & 293  & $5.758\times10^{3}$ & $3.728\times10^{6}$ \\
\bottomrule
\end{tabular}
\end{table}

All frozen criteria pass: 8/8 component aggregates and 79/80 cells favor LEF,
8/8 component aggregates are gap-noninferior, and the component-bootstrap lower
ratio is 463.011. Under an independent fair-sign null, 8/8 favorable units
have one-sided tail $1/256=0.00390625$. Component selection is deterministic;
the reported 8/8 is descriptive for this prospectively fixed cohort, and the
tail probability is conditional on that sign model. All stochasticity, row sums, stationarity, detailed
balance, and original double-precision IACT decomposition certificates pass; the
maximum realized finite-kernel residual is $1.21\times10^{-15}$.

The ratios in Table~\ref{tab:new-components} are heavy-tailed. Nested-cell ratios
range from 0.998 to $7.16\times10^9$ because some
base operators are nearly disconnected for learned energy. Baseline IACT has
median $2.12\times10^5$ and maximum $2.05\times10^{12}$, while LEF IACT has
median 22.52. The targets also remain concentrated: median maximum state mass is
0.840, median inverse-participation effective support is 1.383, and 35/80 cells
place more than 0.9 mass on one state. We therefore headline directional counts
and report the $1.06\times10^4$ component-geometric ratio as a heavy-tail summary.

A separately frozen arbitrary-precision audit reconstructs every base and LEF
kernel from serialized decimal targets and energies. Its 80- versus 120-digit
Poisson solves agree within $1.57\times10^{-70}$ and their largest scaled residual
is $1.15\times10^{-113}$; all 80 stored improvement directions are retained. It
nevertheless fails its formal numerical-support criterion because one extreme
base kernel has stored-spectral versus 120-digit relative error
$1.4866\times10^{-4}>10^{-4}$. Removing the one, three, or five cells with largest
stored baseline IACT leaves 78/79, 76/77, and 74/75 favorable cells. Thus the
directional conclusion is high-precision stable, but the exact magnitude of every
extreme ratio falls outside the prespecified certification tolerance. V1 of this audit is
invalidated because its comparison helper used the wrong outer arithmetic context;
V2 changes only that implementation context.

\begin{table}[ht]
\caption{Arbitrary-precision audit for the nine prespecified diagnostic cells
(the eight largest stored base IACTs plus the single unfavorable cell). Full
base/LEF results for all 80 cells are released in the machine-readable summary.
$\lambda_{\min}^+$ is the 80-digit smallest positive eigenvalue of the symmetric
normalized conductance Laplacian
$I-\Pi^{1/2}P\Pi^{-1/2}$ for the reconstructed base kernel.}
\label{tab:extreme-iact-precision}
\centering
\scriptsize
\setlength{\tabcolsep}{2.8pt}
\begin{tabular}{lrrrrr}
\toprule
cell & stored $\tau_{\rm base}$ & 120-digit $\tau_{\rm base}$ & rel. error & 120-digit base/LEF & $\lambda_{\min}^+$ \\
\midrule
22-off-c4 & $1.2074\times10^{10}$ & $1.2074\times10^{10}$ & $1.09\times10^{-7}$ & $8.9221\times10^8$ & $1.54\times10^{-10}$ \\
22-off-c5 & $3.0718\times10^{10}$ & $3.0719\times10^{10}$ & $1.71\times10^{-5}$ & $2.0422\times10^9$ & $1.92\times10^{-11}$ \\
22-on-c4  & $1.7859\times10^{10}$ & $1.7859\times10^{10}$ & $6.17\times10^{-6}$ & $1.4587\times10^7$ & $9.32\times10^{-11}$ \\
22-on-c7  & $2.9746\times10^1$  & $2.9746\times10^1$  & $1.14\times10^{-13}$ & $9.9785\times10^{-1}$ & $1.14\times10^{-3}$ \\
23-off-c3 & $2.6406\times10^{10}$ & $2.6407\times10^{10}$ & $1.64\times10^{-5}$ & $1.6810\times10^9$ & $7.43\times10^{-11}$ \\
23-off-c4 & $2.0544\times10^{12}$ & $2.0547\times10^{12}$ & $1.49\times10^{-4}$ & $8.9024\times10^7$ & $7.82\times10^{-13}$ \\
23-off-c5 & $1.3457\times10^{11}$ & $1.3458\times10^{11}$ & $2.07\times10^{-5}$ & $7.1614\times10^9$ & $1.09\times10^{-11}$ \\
23-on-c5  & $3.0505\times10^{11}$ & $3.0506\times10^{11}$ & $2.60\times10^{-5}$ & $2.6638\times10^7$ & $6.08\times10^{-12}$ \\
25-off-c4 & $1.2128\times10^{11}$ & $1.2128\times10^{11}$ & $9.39\times10^{-6}$ & $1.6284\times10^4$ & $1.42\times10^{-11}$ \\
\bottomrule
\end{tabular}
\end{table}

This experiment establishes prospective finite-state transfer beyond the original
fixed component family under the same deterministic QM9 construction rule. The
immutable protocol, input artifact, preflight, and summary hashes are recorded in
the accompanying outcome document and release manifest.

\section{Base-Graph Construction Sensitivity}
\label{app:base-graph-sensitivity}

The held-out and new-component tests above use one deterministic
root-star-plus-peer-4NN builder. A frozen artifact-only sensitivity reuses the
original 40 held-out state/target cells and preserves each original edge count,
but replaces the base graph with two deterministic, target-blind, root-free
alternatives. The Hamming-local construction forms a Hamming-distance Kruskal
tree and adds shortest remaining pairs; the hash-based matched graph starts from a canonical
cycle and adds SHA-256-ranked pairs. Both then score all nonedges and select
LEF-4 or capacity-4 once from the unaugmented graph.
Here a favorable primary cell means
$\tau_E(P_{\rm base})/\tau_E(P_{\rm LEF+4})>1$. LEF-versus-capacity results are
reported separately below.

\begin{table}[ht]
\caption{Base-graph sensitivity on the same 40 held-out cells. Intervals exactly
enumerate all ordered size-five outer-seed resamples.}
\label{tab:base-graph-sensitivity}
\centering
\scriptsize
\setlength{\tabcolsep}{3.4pt}
\begin{tabular}{lrrrr}
\toprule
base graph & seeds & cells & outer interval & geometric base/LEF \\
\midrule
original star + peer-4NN & 5/5 & 35/40 & [17.230, 139.959] & 45.023 \\
Hamming-local matched & 5/5 & 36/40 & [12.243, 128.575] & 38.959 \\
hash-based matched & 5/5 & 40/40 & [70.195, 499575.981] & 5921.782 \\
\bottomrule
\end{tabular}
\end{table}

Both prespecified alternatives pass. The large ratios reflect extremely slow
finite base kernels. Capacity-over-LEF favors LEF in 28/40 Hamming-local and 26/40
hash-based cells, preserving a heterogeneous capacity comparison. The original
root has median target mass $4.41\times10^{-7}$ (range
$2.58\times10^{-12}$ to $5.72\times10^{-4}$), and deleting it leaves the nonroot
graph connected in all 40 cells. Weighted stationary flow therefore remains the
relevant bottleneck diagnostic. The audit covers these two fixed constructions.

\section{Additional Observable and Target-Family Scope Audits}
\label{app:scope-audits}

The observable-specific audit uses a Poisson residual tolerance of $10^{-6}$; its maximum realized residual is $6.16\times10^{-8}$. On the same 40 held-out cells and with
the same LEF edges, base/LEF outer IACT ratios are 12.516 for edge count, 16.771
for total bond order, 383.391 for wedge count, and 2.312 for triangle count. All
four frozen criteria pass. These effects are heterogeneous: the triangle-count
nested-cell median is 1.008 and wedge-count ratios are heavy-tailed.

A retrospective descriptive dependence audit reconstructs the same four
observables and learned energy over the same 65 states, then computes within-cell
Pearson correlations under each frozen target. Table~\ref{tab:observable-correlation}
reports the median absolute correlation across the 40 cells. Five of the six
topology-observable pairs are below 0.1, while edge and wedge count retain a
moderate target-weighted relation (0.553). They are more redundant under uniform
state weighting: their correlation is 0.966--0.977 across the four components.
The four columns therefore span distinct but partially correlated structural
signals, with stronger dependence in some concentrated cells.

\begin{table}[ht]
\caption{Median absolute target-weighted Pearson correlation on the same 40
restricted cells. This is descriptive and adds no outer replicates.}
\label{tab:observable-correlation}
\centering
\scriptsize
\setlength{\tabcolsep}{4.0pt}
\begin{tabular}{lrrrrr}
\toprule
 & energy & edges & bond order & wedges & triangles \\
\midrule
energy & 1.000 & 0.071 & 0.369 & 0.142 & 0.006 \\
edges & 0.071 & 1.000 & 0.010 & 0.553 & 0.050 \\
bond order & 0.369 & 0.010 & 1.000 & 0.096 & 0.001 \\
wedges & 0.142 & 0.553 & 0.096 & 1.000 & 0.012 \\
triangles & 0.006 & 0.050 & 0.001 & 0.012 & 1.000 \\
\bottomrule
\end{tabular}
\end{table}

\begin{figure}[ht]
\centering
\includegraphics[width=0.86\linewidth]{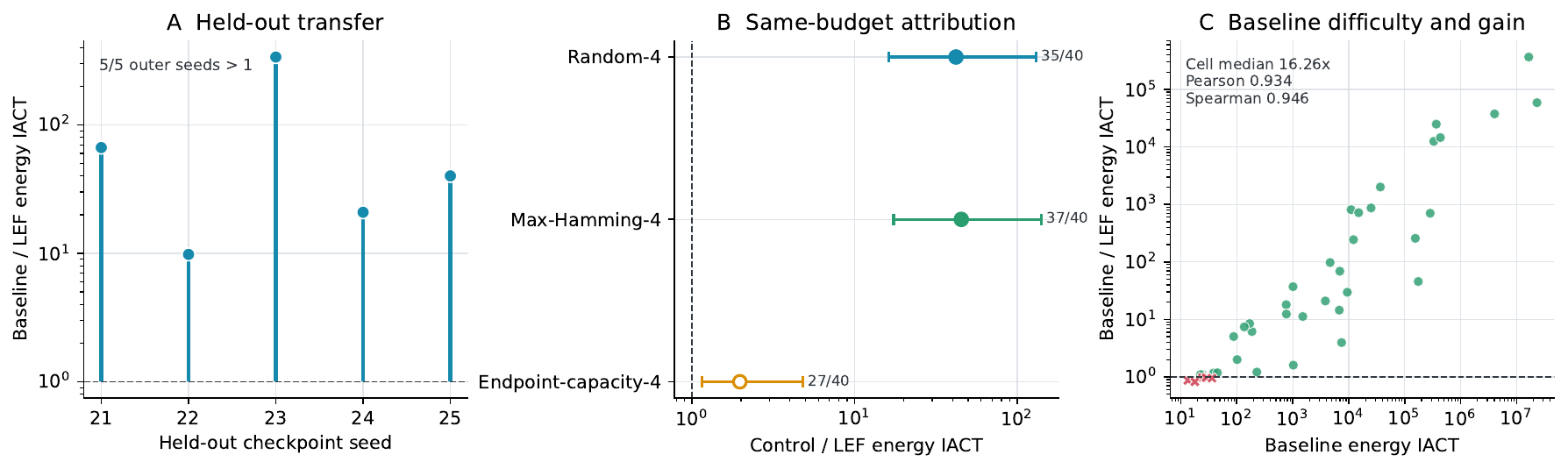}
\caption{Held-out transfer and attribution. All five seed aggregates improve;
LEF beats random, maximum-Hamming, and energy-only, while capacity remains
heterogeneous across nested cells. Panel C is descriptive because baseline IACT
appears in both axes.}
\label{fig:heldout}
\end{figure}

The controlled synthetic family holds one 16-state graph topology fixed across
16 seeds and orthogonally crosses three levels each of within-mode temperature,
exact mode mass, bottleneck conductance, and observable slow-mode alignment. All
1,296 cells are retained. At the fully uniform target and lowest cross-mode
conductance (the most severe bottleneck), the exact observable-specific edge
improves IACT by 21.999 times. Greater alignment is favorable in 426/432 matched
cells, lower cross-mode conductance is favorable in 144/144, and a fixed cross
edge beats a fixed within-mode edge in all 16 diffuse
controls. Effective support spans 1.449--16.000. These results isolate the
finite-state bridge mechanism across this controlled target family.

\section{Molecular Temperature Sensitivity}
\label{app:diffuse}

The original 40 checkpoint/component cells retain their states, energies,
base supports, and candidate nonedges. In each cell, choose $\beta$ from the
energy vector alone to reach $N_{\rm eff}/65\in\{0.1,0.3,0.7\}$, then apply
the unchanged LEF and capacity formulas. The corresponding inverse-temperature
ranges are 1.336--7.920, 0.817--3.603, and 0.422--1.215. Exact conditional greedy
evaluates all remaining candidates after each full degree-MH reconstruction.
Candidate sets have 1,783--1,785 nonedges; every method receives four insertions.

\begin{table}[ht]
\centering
\caption{Absolute energy IACT medians and paired effects in the diffuse-target
sensitivity analysis. Intervals exhaustively resample the five original outer
seed effects ($5^5$ resamples).}
\begin{tabular}{rrrrrrr}
\toprule
$N_{\rm eff}$ & Base & Capacity & LEF & Greedy & Base/LEF [95\%] & Capacity/LEF [95\%]\\
\midrule
6.5 & 46.753 & 31.682 & 25.406 & 19.492 & 2.095 [1.796,2.609] & 1.301 [1.210,1.457]\\
19.5 & 15.665 & 13.054 & 10.808 & 9.096 & 1.404 [1.348,1.476] & 1.189 [1.150,1.249]\\
45.5 & 6.289 & 5.751 & 5.135 & 4.322 & 1.200 [1.163,1.238] & 1.130 [1.120,1.139]\\
\bottomrule
\end{tabular}
\end{table}

This subsequent sensitivity analysis reuses the original five model seeds as
outer units. Each seed geometrically aggregates eight nested cells. In order
21--25, the capacity/LEF seed effects are $(1.614,1.276,1.224,1.251,1.183)$,
$(1.308,1.160,1.151,1.189,1.143)$, and
$(1.130,1.147,1.121,1.115,1.137)$ for the three regimes. At $N_{\rm eff}=45.5$,
maximum target mass lies between 0.0369 and 0.0603. All 120 cells remain in the
report, including the one static LEF portfolio that slightly outperforms
conditional greedy in the lowest-support regime.

The calculation performs 855,960 grounded Poisson solves with maximum relative
residual $1.50\times10^{-13}$. Separately constructed transition matrices and
full spectral sums check all 480 final kernels, with maximum relative difference
$1.34\times10^{-12}$. Only the stationary eigenmode is removed. Six synthetic
correctness tests cover the spectrum, kernel construction, and greedy steps.
The calculation takes 58.1 CPU seconds with cached energies.

\section{External Ising Protocol and Individual Results}
\label{app:ising}

\subsection{Instances, supports, and selectors}

For eight spins $s_i\in\{-1,1\}$, use the energy
\begin{equation}
 E(s)=-\sum_{(i,j)\in\mathcal E_{\rm lattice}}J_{ij}s_i s_j-\sum_i H_i s_i.
\end{equation}
The physical interaction graph is the open $2\times4$ lattice with ten bonds.
Six ferromagnetic instances use PCG64 seeds 2026101600--2026101605; six mixed-sign
instances use 2026101606--2026101611. Coupling magnitudes are uniform integers
from 10 through 30 divided by 20. Mixed-sign couplings additionally receive
independent equiprobable signs. Fields are uniform integers from $-5$ through
5 divided by 20. The protocol fixes all parameters and hashes the source before
evaluating any formal instance.

Every instance uses both $\beta=0.3$ and $\beta=0.8$. All 256 configurations are
enumerated. The transition support is the single-spin hypercube, with degree
eight and 1,024 undirected edges. Of its 31,616 nonedges, the first 512 under
SHA-256 ordering of
\texttt{ising-natural-support-v1:ensemble:seed:u:v}
form the common candidate set. This set is energy-blind and shared across
temperatures. Canonical edge order breaks all score ties.

LEF and capacity choose their top four once using the original degrees.
Exact conditional greedy tests all remaining candidates under the complete
degree-MH kernel at each of four stages. Random uses 32 uniformly sampled
four-edge portfolios, without replacement within each portfolio, from seed
\texttt{instance seed + 1000000}; the same portfolios are reused across
temperatures. Its cell comparator is the geometric mean of their 32 IACTs.
All evaluated kernels use the final degrees and self-loop floor 0.05.

Energy is the selection and primary evaluation observable. Magnetization
$M(s)=\sum_i s_i$ is evaluated on those same selected supports. Table~\ref{tab:ising-cells}
contains every cell's absolute energy IACT and paired energy/magnetization
comparison. The external trial uses 12 independently generated coefficient
instances as outer units. Each ensemble/temperature interval exhaustively
resamples its six instance log effects ($6^6$ resamples). Temperatures and
random portfolios are nested conditions.

\begin{table}[htbp]
\centering\small
\caption{All external Ising cells. Absolute energy IACTs use the direct degree-MH kernels. Ratios compare capacity to LEF using the same energy-selected edges; above one favors LEF.}
\label{tab:ising-cells}
\begin{tabular}{llrrrrrrr}
\toprule
Family & Seed suffix & $\beta$ & $N_{\rm eff}$ & Base $\tau_E$ & Capacity $\tau_E$ & LEF $\tau_E$ & Ratio $E$ & Ratio $M$\\
\midrule
Ferro & 00 & 0.3 & 75.50 & 11.761 & 11.515 & 11.095 & 1.0379 & 0.9541\\
Ferro & 00 & 0.8 & 3.52 & 68.458 & 53.792 & 62.090 & 0.8663 & 0.7461\\
Ferro & 01 & 0.3 & 60.89 & 13.743 & 13.630 & 12.945 & 1.0529 & 0.8892\\
Ferro & 01 & 0.8 & 3.91 & 32.421 & 31.277 & 30.630 & 1.0211 & 1.0604\\
Ferro & 02 & 0.3 & 100.54 & 10.393 & 10.232 & 9.622 & 1.0633 & 0.9340\\
Ferro & 02 & 0.8 & 4.06 & 74.433 & 61.123 & 63.032 & 0.9697 & 0.9532\\
Ferro & 03 & 0.3 & 130.90 & 8.328 & 8.325 & 7.740 & 1.0755 & 0.9984\\
Ferro & 03 & 0.8 & 8.11 & 40.487 & 34.739 & 35.332 & 0.9832 & 0.7063\\
Ferro & 04 & 0.3 & 65.24 & 13.586 & 13.516 & 13.166 & 1.0266 & 1.0077\\
Ferro & 04 & 0.8 & 2.11 & 164.689 & 127.887 & 155.060 & 0.8248 & 0.7878\\
Ferro & 05 & 0.3 & 79.81 & 10.764 & 10.736 & 10.384 & 1.0338 & 0.9633\\
Ferro & 05 & 0.8 & 3.63 & 57.130 & 54.801 & 57.258 & 0.9571 & 0.8922\\
Mixed & 06 & 0.3 & 115.61 & 7.284 & 7.194 & 7.065 & 1.0182 & 0.9825\\
Mixed & 06 & 0.8 & 19.41 & 25.607 & 22.552 & 22.798 & 0.9892 & 1.0784\\
Mixed & 07 & 0.3 & 103.87 & 8.584 & 8.455 & 8.257 & 1.0239 & 0.9174\\
Mixed & 07 & 0.8 & 27.82 & 19.631 & 15.013 & 15.970 & 0.9401 & 0.5168\\
Mixed & 08 & 0.3 & 104.78 & 9.218 & 9.131 & 8.819 & 1.0353 & 0.9705\\
Mixed & 08 & 0.8 & 17.23 & 21.757 & 19.194 & 20.129 & 0.9535 & 0.8686\\
Mixed & 09 & 0.3 & 97.91 & 9.905 & 9.949 & 9.466 & 1.0511 & 0.9569\\
Mixed & 09 & 0.8 & 9.25 & 30.598 & 30.700 & 29.654 & 1.0353 & 0.8607\\
Mixed & 10 & 0.3 & 117.18 & 8.581 & 8.440 & 8.077 & 1.0449 & 0.9586\\
Mixed & 10 & 0.8 & 11.31 & 40.020 & 31.719 & 30.840 & 1.0285 & 1.0013\\
Mixed & 11 & 0.3 & 96.01 & 9.871 & 9.634 & 9.422 & 1.0225 & 1.0057\\
Mixed & 11 & 0.8 & 10.86 & 30.693 & 28.717 & 27.919 & 1.0286 & 1.0023\\
\bottomrule
\end{tabular}
\end{table}

\subsection{Validation and computation}

Ten tests check small analytical certificates, graph structure, local scoring,
and independent spectral/Poisson agreement. The 49,032 exact greedy evaluations
have maximum Poisson residual $4.51\times10^{-13}$. Independently constructed
transition matrices and full eigendecompositions verify 864 final kernels and
1,728 energy/magnetization measurements. Maximum spectral/Poisson relative
difference is $4.10\times10^{-13}$; maximum kernel residual is
$4.44\times10^{-16}$. Every saved certificate and source hash also verifies.
The metric stage takes 44.7 CPU seconds. The twelve energy enumerations require
3,072 analytic energy values, reused at both temperatures.

\section{Actual-Intervention Flow Accounting}
\label{app:flow-accounting}

For degree-MH, $c_{uv}=(1-\ell)\min(\pi_u/d_u,\pi_v/d_v)$ on every edge.
Adding edges increases endpoint degrees, so original conductances weakly
decrease. Define $N$ from the new-edge conductances and $R$ from original-edge
decreases, both as positive semidefinite Laplacians. Along the convex path
$A_t=(1-t)A_0+tA_1$, the chain remains reversible, feasible, and connected.
On the Euclidean mean-zero subspace,
\begin{equation}
 \frac{d}{dt}q^\top A_t^\dagger q
 =-g_t^\top(N-R)g_t,\qquad g_t=A_t^\dagger q.
\end{equation}
Integration yields Equation~\ref{eq:flow-decomposition}. The two nonnegative
terms quantify new-edge benefit and lost original flow along this specified
path; their difference equals the actual endpoint variance change.

All 80 LEF/capacity selector-cells use rational targets and exact old-loss
assembly, followed by 80-digit analytic integration and separate 100-digit
endpoint solves. The largest relative endpoint-closure discrepancy is
$1.2204\times10^{-72}$. In Table~\ref{tab:flow-parts}, each part is divided by
its cell's baseline asymptotic variance, averaged over eight cells per seed,
then equally over five seeds. This additive aggregation preserves the identity.

\begin{table}[ht]
\centering
\caption{Baseline-normalized additive flow accounting on the original 40 cells.}
\label{tab:flow-parts}
\begin{tabular}{lrrrr}
\toprule
Selector & New benefit & Old loss & Net gain & Base-improved cells\\
\midrule
LEF & 0.71434741 & 0.02117495 & 0.69317246 & 35/40\\
Capacity & 0.65029703 & 0.02259880 & 0.62769823 & 35/40\\
\bottomrule
\end{tabular}
\end{table}

The paired LEF advantage is $+0.06405038$ from new flow and $+0.00142385$ from
the difference in old-flow losses, totaling $+0.06547423$. Gross new-flow
preferences are 28 LEF, four capacity, and eight identical portfolios. Net
preferences are 27, five, and eight. The single reversal is
\texttt{seed25\_off\_graph02}: a new-flow advantage of $0.00001523166$ is offset
by an extra old-flow loss of $0.00819855982$, giving a net LEF disadvantage
$0.00818332816$ and capacity/LEF IACT ratio 0.93938179. All eight ties have exactly
the same canonical selected edge sets.

\section{Statistical Units and Evidence Roles}
\label{app:statistics}

The one-step proposal confirmation uses five fresh trained-model seeds. Its
matched-query comparison is a separate development-model experiment with
nested anchors. The four early finite components support exact mechanism
diagnostics. Molecular checkpoint transfer uses five seeds crossed with two
training modes and four fixed components (40 cells); new-component transfer
uses eight components crossed with the same ten checkpoints (80 cells).
Component selection follows the deterministic structural rule in
Appendix~\ref{app:component-spec}. The original five-seed base/LEF geometric
effects are reported with the original experiment's results; numerical
resampling precision leaves the outer sample size unchanged.

The original selector-attribution Random-4 comparator geometrically averages
64 fixed portfolios per cell. Its 30/40-favorable condition is a prespecified
heterogeneity safeguard, while inference aggregates by the five outer model
seeds. Later molecular weighting, intervention accounting, temperature, and
correlation analyses reuse their existing outer units. The Ising experiment
adds twelve coefficient-instance units. The separate measured-budget study
adds five new coefficient seeds, with three paired sizes, three starts,
and 32 nested chains per start (Appendix~\ref{app:finite-budget}).
Exact-rational inequalities establish
instancewise comparisons independently of sampling-based confidence intervals.
Release-score ablations reuse the twelve original Ising instances, with three
timing repeats and paired temperatures. The separately pre-specified factorial
confirmation adds twelve new Ising coefficient-instance units crossed with two
temperatures, two selectors and two release rules, and maintains three nested
construction repeats per selector. Its four primary tests operate on the twelve
paired-temperature outer log effects with Holm correction. The earlier
independent eight-instance confirmation evaluates the original combined
score against hash release; these cohorts are not pooled.

\section{Artifact Map}
\label{app:artifacts}

The ancillary files accompany the manuscript source.

\begin{itemize}
\item \textbf{Verification instructions.}
The ancillary \path{README.txt} lists the runnable checks.
\item \textbf{Exact analytic witness.} \path{analytic_witness.json}
contains the target, observable, supports, primal current and conductances,
and base dual variables for the $m=11$ construction in
Proposition~\ref{prop:structural-separation}.
\path{verify_witness.py} checks both witnesses and the strict IACT
separation using rational arithmetic.
\item \textbf{Prospective Ising table.}
\path{rebuild_ising_table.py} rebuilds Table~\ref{tab:ising-factorial}
from the supplied \path{ising_factorial_20pct.md}.
The script reproduces the table from reported aggregates. The original
per-condition records, including \path{paired_cells.csv}, and empirical
witness bundles belong to the underlying research runs.
\end{itemize}

\section{Proposal, Spectral, and Observable Context}
\label{app:context}
\label{sec:movement}
\label{sec:proposal}

\paragraph{Proposal-level motivation.}
A radius-three quadratic molecular proposal uses pair interactions to improve
accepted one-step movement over unary proposals on five fresh trained-model
seeds: ESJD difference $+0.16994$, with hierarchical interval
$[0.13692,0.20061]$. A separate development comparison gives simpler repeated
kernels more movement and energy reduction under the same 226-query ceiling.
The quadratic attempt costs 226 queries versus 34 per unary attempt. Reverse
support failures and extreme-tail surrogate selection supply additional
diagnostics (Appendices~\ref{app:quadratic} and~\ref{app:proposal-boundaries}).

\paragraph{Spectral gap versus observable variance.}
For nonstationary reversible eigenpairs $(\lambda_k,\psi_k)$, expand
$h=\sum_{k\ge2}a_k\psi_k$ in $L^2(\pi)$. Then
\begin{equation}
 \tau_f=\frac{\sum_{k\ge2}a_k^2(1+\lambda_k)/(1-\lambda_k)}
                    {\sum_{k\ge2}a_k^2}.
\end{equation}
Fastest-mixing reweighting improves the absolute gap by $2.609$--$6.657\times$
on four development components. In seed 2, the SLEM mode contributes 14.96\%
of energy IACT after reweighting, while slow rank 3 contributes 80.79\%, explaining
the $6.083\times$ gap versus $1.043\times$ energy-IACT contrast. The main
certificates instead compare the same observable under both support constraints.

\paragraph{Insertion and shortlist controls.}
Four exact marginal self-loop-to-edge insertions improve the development
components by $32.288\times$ geometrically. In the separate 16-edge, four-bridge
set problem, exhaustive evaluation of all 1,820 subsets leaves exact conditional
greedy within $1.0107\times$ geometrically of the optimum (maximum 1.0416).
This localizes the small remaining combinatorial headroom to that tested
shortlist once exact conditional values are supplied.

\paragraph{Controlled mechanism and finite traces.}
A fixed 16-state synthetic family varies mode mass, cross-mode conductance,
within-mode temperature, and observable alignment. At a uniform target, an
observable-aware cross edge retains $21.999\times$ IACT improvement; cross edges
beat within-mode controls in all 16 diffuse comparisons. This is an oracle
mechanism result. A molecular trajectory-driven discovery follow-up builds a
common 129-state universe before evaluating frozen graphs with 1,024-step
stationary-start traces. Its exact-operator effects remain favorable, while
the prespecified joint finite-trace criterion is missed.

\begin{figure}[ht]
\centering
\includegraphics[width=\textwidth]{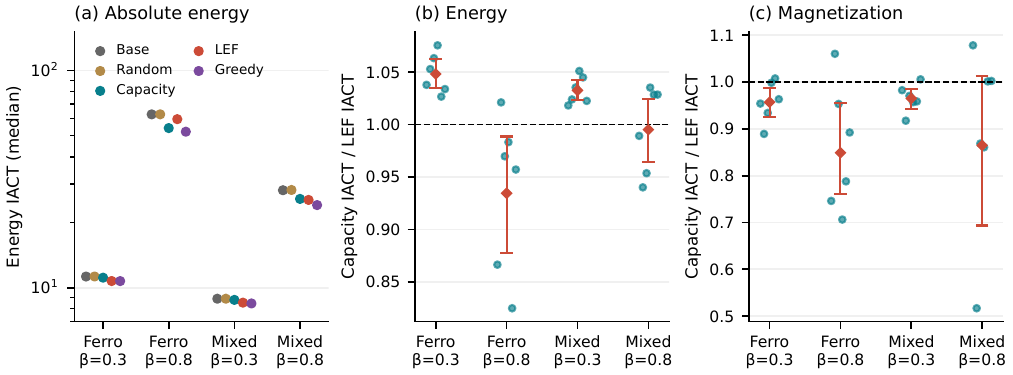}
\caption{Direct degree-MH evaluation on the external Ising cohort. Left: median
absolute energy IACT. Center and right: capacity/LEF ratios for energy and
magnetization using identical energy-selected edges. Points are the six
coefficient instances per group; diamonds and bars give geometric ratios and
outer-instance bootstrap intervals. Above one favors LEF. These are direct-kernel
results; Section~\ref{sec:common-weighting} separately compares optimized supports.}
\label{fig:ising}
\end{figure}

\section{Edge Information and Low-Dimensional Approximations}
\label{app:operator}

On the tested 16-edge shortlist with four insertions, the global optimum improves
independent LEF top-four by $3.749\times$ geometrically, while exact conditional
greedy is within $1.0107\times$ of the global optimum. Exact edge valuation thus
almost closes the global-search headroom in this finite problem. Rank-one
conductance insertion isolates the required Poisson contrast and resistance.

Using the exact transition operator, an eight-step Krylov approximation yields
a geometric IACT ratio 1.037 to exact marginal greedy. An empirical full operator
with eight Krylov steps and 256 transition samples per state per stage gives
ratios 1.108, 2.494, 129.805, and 3.330 on the four development components.
The sampled operators satisfy the stationary-flow and detailed-balance checks.
These results separate solving with available operator information from acquiring
that information through the tested estimator.

The projected marginal-score diagnostic further tests endpoint sensitivity. In
the difficult component, using the projected numerator with exact resistance
gives an IACT ratio 129.498 to the reference; using exact numerator with projected
resistance gives 1.293. Median candidate-edge coverage in the observable Krylov
subspace is 0.370. These diagnostics locate the dominant error in endpoint
contrasts for this representation, motivating direct attention to $g(u)-g(v)$.

\section{Dynamic Candidate-Refresh Stress Test}
\label{app:dynamic-refresh}

A separately prespecified experiment uses an exact-MH pilot to choose four
successive refresh roots. Each refresh contributes 32 energy-blind, exactly
rescored states. All selectors receive the same final 129 states, target, base
edges, 16-edge budget, and 129 neural-energy queries; final graphs are frozen
before running 64 stationary-start chains for 1,024 steps. The protocol combines
dynamic shared candidate discovery with a frozen common-universe evaluation.

Exact operator IACT favors LEF over random and Maximum-Hamming in 40/40 cells, by
537.730 and 554.493 times geometrically across outer seeds. Finite traces retain
favorable 5/5 seed aggregates and intervals for both controls. Maximum-Hamming
passes the full criterion (31/40 favorable cells), but random has 29/40 rather
than the required 30/40. The joint outcome is therefore directional. Capacity
records exact 1.641 times and 30/40, with finite 1.273 times, 4/5 seeds, and
29/40. These results provide
directional finite-chain evidence under chain-driven candidate refresh.

\section{Quadratic Proposal Construction and Matched-Query Evaluation}
\label{app:quadratic}

\subsection{A bounded quadratic proposal improves one-step movement}

At a current graph $x$, let $S$ denote a compatible set of categorical edits and
$E_x(S)=E_\theta(x\oplus S)$. The discrete energy admits a M\"obius expansion
\begin{equation}
 E_x(S)=\theta_\emptyset+
 \sum_{i\in S}\theta_i+
 \sum_{\{i,j\}\subseteq S}\theta_{ij}+R_{\ge3}(S).
\end{equation}
The quadratic surrogate retains terms through order two. Current, single, and
pair queries determine
\begin{equation}
 \theta_i=E_x(\{i\})-E_x(\emptyset),\qquad
 \theta_{ij}=E_x(\{i,j\})-E_x(\{i\})-E_x(\{j\})+E_x(\emptyset).
\end{equation}
For the primary radius-three kernel, four categorical coordinates produce a
369-state support. The model queries 113 current/single/pair states in each
direction; 256 triple-edit states are scored only by quadratic extrapolation.
Writing $\widehat E_x$ for this fitted surrogate and $\mathcal C(x)$ for the
finite compatibility-filtered support (including $x$), the proposal is explicitly
normalized as
\begin{equation}
 q(y\mid x)=\frac{\mathbf 1\{y\in\mathcal C(x)\}
 \exp[-\beta_{\rm prop}\widehat E_x(y)]}
 {\sum_{z\in\mathcal C(x)}\exp[-\beta_{\rm prop}\widehat E_x(z)]}.
 \label{eq:qubo-proposal}
\end{equation}
At the proposed endpoint, the block, compatibility filter, surrogate, and
normalizer are rebuilt; this supplies the exact reverse probability in the MH
ratio. Incompatible assignments have zero proposal mass.
Across five fresh trained-model seeds, QUBO minus unary accepted edge-edit ESJD is
$+0.16994$, with hierarchical 95\% interval $[+0.13692,+0.20061]$; all five
seeds are favorable. The 50-step valid-connected difference is $+0.0164$, with
interval $[+0.0076,+0.0252]$, as a secondary safeguard.

This is a genuine local interaction result: queried pair terms improve the
ranking of unqueried triple edits. The next comparison evaluates efficiency under
a matched neural-energy-query budget; exact bidirectional normalization requires
226 queries for QUBO versus 34 for unary.

\subsection{A separate development comparison at matched neural-energy queries}

On the development anchors, under a matched 226-query ceiling, one radius-three QUBO attempt is compared with
six sequential unary attempts, nine pair Gibbs attempts, and nine metropolized
pair Gibbs attempts from the same warmed anchors. The two modes use paired
checkpoints trained with QAP alignment disabled or enabled; QAP is ancillary to
the sampling argument and is detailed in Appendix~\ref{app:qap}.

\begin{table}[t]
\caption{Matched neural-energy-query comparison under a 226-query ceiling.}
\label{tab:equal-work}
\centering
\scriptsize
\setlength{\tabcolsep}{3.2pt}
\begin{tabular}{llrrrr}
\toprule
Mode & Method & Queries & Acc. ESJD & Net Ham.$^2$ & $\Delta E$ \\
\midrule
QAP-off & QUBO & 226 & 2.125 & 2.125 & $-0.819$ \\
QAP-off & unary $\times6$ & 204 & 7.438 & 14.906 & $-1.878$ \\
QAP-off & pair Gibbs $\times9$ & 225 & 5.156 & 14.375 & $-2.141$ \\
QAP-off & met. pair $\times9$ & 225 & 5.219 & 16.063 & $-2.169$ \\
QAP-on & QUBO & 226 & 2.781 & 2.781 & $-0.793$ \\
QAP-on & unary $\times6$ & 204 & 8.688 & 18.813 & $-2.335$ \\
QAP-on & pair Gibbs $\times9$ & 225 & 5.156 & 16.188 & $-2.778$ \\
QAP-on & met. pair $\times9$ & 225 & 6.500 & 21.344 & $-2.738$ \\
\bottomrule
\end{tabular}
\end{table}

All three repeated baselines have strictly higher net movement under the matched
query ceiling; one interaction-aware transition does not offset their repeated opportunities.

\subsection{Why does local proposal quality fail to transfer?}

Three losses recur. \textbf{Construction:} the dense attempt uses 226 energy
queries; a $p=6$ screen lowers this to about 176 but still loses to every strong
comparator. \textbf{Reciprocity:} endpoint rebuilding omits the reverse source in
9/96 QAP-off and 24/96 QAP-on attempts, forcing MH rejection. \textbf{Extreme
selection:} at 1,316,401 states, lower random triple-edit MAE coexists with worse
selected true energy and accepted ESJD. Each retains its own evidence contract
(Appendix~\ref{app:proposal-boundaries}).

\section{Extended Proposal Boundaries}
\label{app:proposal-boundaries}

\begin{figure}[ht]
\centering
\includegraphics[width=0.90\linewidth]{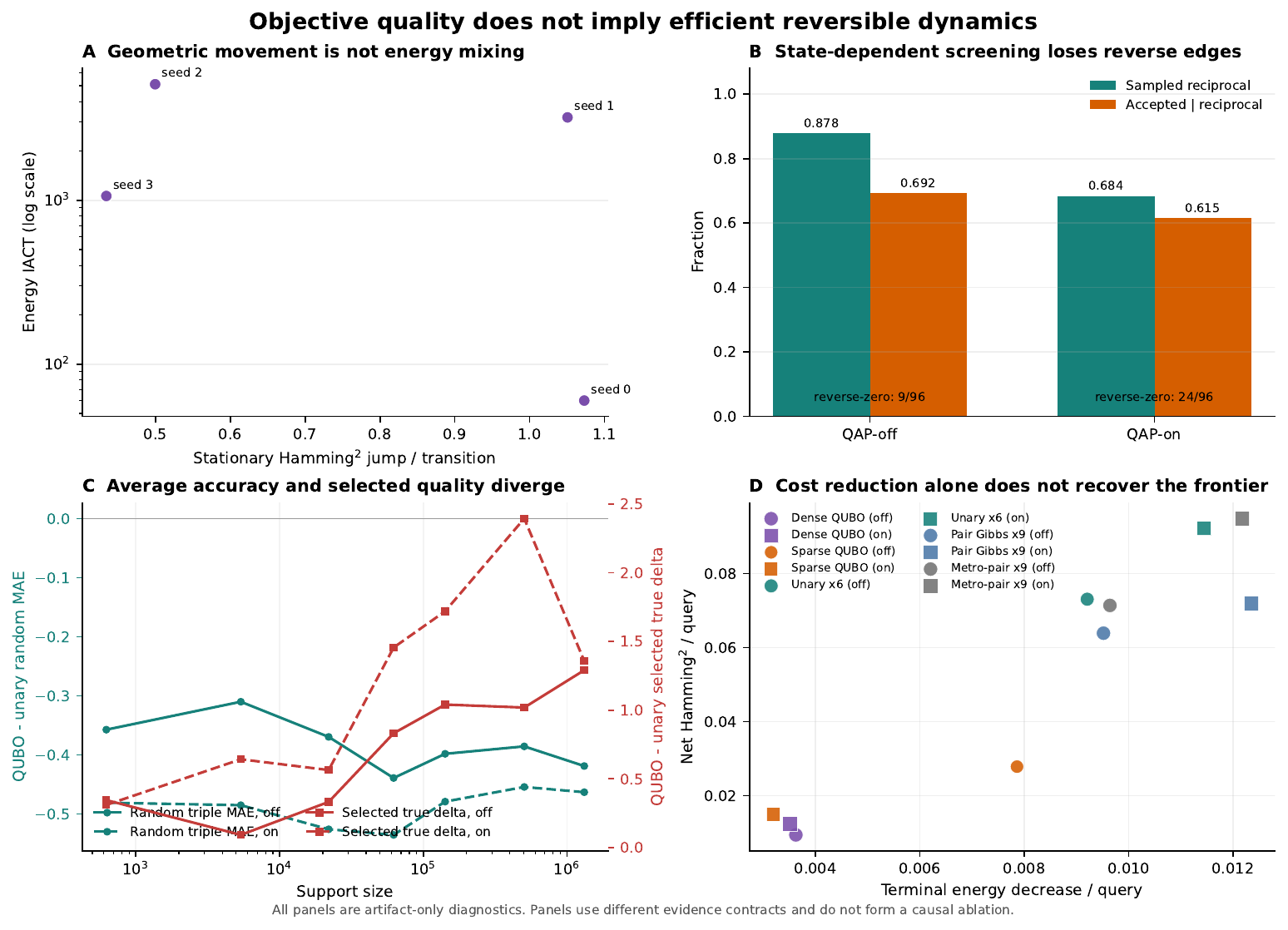}
\caption{A local interaction gain is filtered through neural-energy-query cost,
reciprocal support, extreme-tail selection, and target-relevant mixing. The four
panels report separate evidence contracts.}
\label{fig:boundaries}
\end{figure}

\subsection{Exact rescoring leaves the discovery gap}

At the 1.3-million-state support, selecting QUBO top-64 candidates and exactly
rescoring them still produces worse root ESJD than unary by $-2.9741$, interval
$[-5.8458,-0.3589]$, and lowers target-weighted valid-connected by $-0.1337$.
Positive stationary ESJD ($+0.2823$) applies only to the target restricted to
the frozen component; root-mobility and validity criteria remain below threshold.

The retained top-$B$ QUBO optimism gap is $+5.983/+8.099$ in the two modes; a
queried-true negative control is zero. Resulting H2 graphs have mean leaf fraction
0.969, mean cycle rank 1.625, and a 7.438-order collapse in stationary ESJD from
$\beta=1$ to $\beta=40$. These separate mechanism diagnostics characterize the
observed failure modes.

\subsection{Detailed reciprocity and same-graph reweighting}

For the screened sparse proposal, excluding self draws gives realized-edge
reciprocal fractions 0.878 and 0.684 in QAP-off and QAP-on, respectively. As in
the main text, these are sampled-edge diagnostics rather than full proposal-mass
characterizations.

\begin{table}[ht]
\caption{Same-graph fastest-mixing reweighting.}
\label{tab:fmmc-full}
\centering
\scriptsize
\setlength{\tabcolsep}{3pt}
\begin{tabular}{lrrrrrr}
\toprule
Cell & MH gap & FMMC gap & Gap ratio & MH IACT & FMMC IACT & IACT ratio \\
\midrule
seed 0 & 0.002458 & 0.011499 & 4.678 & 59.91 & 26.31 & 2.277 \\
seed 1 & 0.0000222 & 0.000147 & 6.657 & 3204.70 & 612.48 & 5.232 \\
seed 2 & 0.0000128 & 0.0000778 & 6.083 & 5105.65 & 4896.75 & 1.043 \\
seed 3 & 0.001153 & 0.003009 & 2.609 & 1061.90 & 434.99 & 2.441 \\
\bottomrule
\end{tabular}
\end{table}

\subsection{Detailed four-edge oracle}
\label{app:bridge-table}

\begin{table}[ht]
\caption{Four-edge fixed-state oracle. Ratios greater than one indicate
improvement.}
\centering
\small
\begin{tabular}{lrrrr}
\toprule
Selector & Mean gap & Gap range & Mean IACT & IACT range \\
\midrule
Random & 1.004 & 1.000--1.017 & 1.862 & 0.994--4.301 \\
Maximum Hamming & 1.000 & 1.000--1.000 & 1.539 & 1.000--3.152 \\
Target-mass capacity & 17.821 & 1.883--33.184 & 24.171 & 1.233--47.861 \\
Slow eigenspace & 17.782 & 1.728--33.184 & 24.408 & 1.612--48.266 \\
Energy Poisson & 17.687 & 1.350--33.184 & 24.428 & 2.668--46.869 \\
\bottomrule
\end{tabular}
\end{table}

\subsection{Closed interventions}

\begin{itemize}
\item A quadratic validity penalty selected on development models fails fresh
confirmation: valid-connected difference $-0.0076$, interval
$[-0.0164,+0.0012]$.
\item Longer-chain development violates prespecified endpoint safeguards; the
50-step endpoint is retained.
\item A synthetic CPU check verifies exact-kernel implementation, but the
prespecified count-only follow-up remains below its advancement criterion and
produces no scientific summary.
\item Learned single-edge and interacting bridge-set QUBOs remain below the
simple LEF selector and their hardware-advancement conditions.
\item Exact MILP improves a frozen additive global-LEF allocation objective by
10.76\% while worsening energy IACT and gap, showing that stronger proxy
optimization preserves its target-dynamics mismatch.
\item The initial held-out-checkpoint LEF invocation stops before scoring because
a hydrogen-including 19-node structural bank is incompatible with a QM9
\texttt{remove\_h=true}, 9-node checkpoint context. Zero scientific cells are
serialized; this is an infrastructure outcome rather than an LEF test.
\item A separate V2 binds compatible energy-blind QM9 components and passes its
prespecified held-out-checkpoint criterion with the frozen selector.
\item The initial post-V2 selector-attribution run stops before science because
of an input-header contract defect. Its V2 changes only that validation plus a
no-science real-input preflight. It attributes LEF over Random-4 and
Maximum-Hamming-4, while the endpoint-capacity increment misses the prespecified
criterion (27/40 favorable versus 30/40 required).
\end{itemize}

\section{Parallel Training-Side Objective--Dynamics Mismatch}
\label{app:qap}

QAP supplies a parallel observation: a more structured training objective can
improve its own loss while downstream dynamics follow a different response.

\subsection{QAP shortens the target and lowers training loss}

Across six paired QM9 trained-model seeds, QAP-on minus QAP-off terminal-window
flow-matching loss is $-0.055126$; all six pairs have the same sign (two-sided
exact sign test $p=0.03125$). A held-out 2,048-pair audit shows that
zero-predictor target displacement falls by 20.60\%, while observed tail loss
falls by 29.19\%. Hungarian alignment already reduces target displacement by
20.10\%. The loss result supports a target-geometry mechanism; the paired
generation endpoints below evaluate downstream behavior separately.

The frozen crossed generation endpoint fails its joint rule. Valid-only FCD is
favorable by $-1.1263$ with interval $[-1.6515,-0.5593]$, but valid-connected
changes by $+0.0734$ with interval $[-0.0419,+0.2360]$ and only 3/6 seeds
improve. FCD was a safeguard and remains secondary supportive evidence.

\subsection{Low-budget generation moves in the opposite direction}

Under a separately prespecified zero-warmup local sampler, QAP-on minus QAP-off
valid-connected AUC is $-0.15109$ with interval $[-0.25006,-0.05091]$; 2/6
seeds are favorable. Oriented valid-only FCD AUC is $-7.93332$ with interval
$[-10.59279,-4.86294]$; 0/6 seeds are favorable. Validity is noninferior, while
connected-given-valid AUC is $-0.22809$.

An offline artifact audit associates QAP-on with more persistent fragmentation
and slower recovery. Mean valid-molecule component count increases by 2.9816 in
AUC and the largest-component heavy-atom fraction decreases by 0.3326; both
severity directions hold for all six seeds. Together, these results show reliable
training-objective reduction alongside persistent-fragmentation signatures and
mixed generative behavior.

\section{Physical-CIM Stress Tests}
\label{app:cim}

Physical CIM supplies a parallel stress test of the movement--utility
distinction.

\subsection{Execution and candidate coverage}

Two independent 64-cell problem banks complete physical execution and achieve
64/64 feasible-candidate coverage. The fresh solver-quality comparison is worse
than its digital reference by $+0.1744$ true-energy delta, with interval
$[+0.0635,+0.3128]$ and 0/4 favorable trained-model means. The hardware evidence
records solver quality and candidate coverage.

A separate 2,000-read sample-mode branch yields 280--674 unique valid-connected
candidates per evaluated cell. Exact rescoring and symmetric frozen graphs
use these optimizer-endpoint candidates to evaluate nonlocal movement and exact-MH
target dynamics.

\subsection{Movement, lifecycle, and reservoir boundaries}

The cached-bank chain improves accepted Hamming-squared/work by
39.86--200.70 times relative to a weak local baseline, while terminal learned
energy is worse by $+0.2396$--$+0.9702$ in all four cells. Single-root
availability fails after departure: 88/128 chains leave and only 1/88 returns.

In a later common-target reservoir comparison, physical CIM has lower energy
IACT than the strongest digital pool in only 1/4 cells. Geometric best-digital
IACT divided by CIM IACT is 0.848, although LEF improves the native bank for every
generator--cell combination. The strongest target-relevant endpoints therefore
come from the digital pool despite real physical-candidate diversity.

An execution-only 656-spin block-diagonal prefetch packs 16 archived roots into
one physical task and passes its 16/16 diversity criterion. A subsequent
actual-trajectory atlas improves ESS/work over local-only by 1.544 times but
achieves only 0.640 times the performance of a stronger trajectory-only
reciprocal cache. Physical promotion was therefore stopped.

Together these experiments show target utility following reciprocal cache quality
more closely than candidate diversity or structured movement.

\end{document}